\documentclass[12]{umd-thesis}
\usepackage{graphicx,psfig,lingmacros,alltt,BJDmacros,ATEX,tocloft} 
\newcommand{\hide}[1]{}
\newcommand{\ul}[1]{\underline{#1}}

\setboolean{masters}{false}     % PhD degree

\title{GENERATION-HEAVY HYBRID MACHINE TRANSLATION} 
\abstractpagetitle{GENERATION-HEAVY HYBRID\\MACHINE TRANSLATION}

\author{Nizar Yahya Habash}
\department{Department of Computer Science}
\advisor{Professor Bonnie J. Dorr}
\chairtitle{Chairperson/Advisor}
\committee{Professor Amy Weinberg\\
  Professor Philip Resnik\\
  Professor William Gasarch\\
  Professor Norbert Hornstein}
\date{2003}

\dedication{\centering {\arabiclarge  
\C241\C68\C20\C234\A{5pt}\C241\C214\C20\A{5pt}\C239\C210\C22}} 

\acknowledgements{ 

I would like to thank my advisor Dr. Bonnie Dorr for always being there
for me, pushing me forward and reviewing my work.  I also would like
to thank Dr. Amy Weinberg for all the hard questions and objective
realism.  I am especially grateful to both Bonnie and Amy for being my
unofficial therapists through this process.

I thank Dr. Philip Resnik for teaching me how to project and how to
evaluate. I thank all my committee members, for their time and effort.
I thank Dr. Louiqa Raschid for her always helpful criticism.  I also
thank the staff of the Computer Science Department and University of
Maryland Institute for Advanced Computer Studies (UMIACS) for their
support and taking care of bureaucratic business.

I thank Mona Diab, David Zajic, Clara Cabezas, Adam
Lopez, Okan Kolak, Dina Dumner-Fushman, Grazia Russo-Lassner, and
Michael Nossal for helpful discussion and for making this a fun
experience.  I thank the lovely trinity Eve Jordan, Ethel B. and Mimi
G. for their instrumental love and support.  I also thank Carole and
Spencer Waesche for giving me a home.

I thank all organizations, institutes and individuals who supported me
financially through this journey.

Most importantly of all, I thank Mama and Baba for their infinitely
wonderful love, confidence, support, and genes.  This work is
dedicated to them.

} 

\abstract{The state of the art techniques in Machine Translation (MT)
require large amounts of symmetric resources from source and target
languages.  This is true regardless of whether the approach is
Transfer or Interlingua, Symbolic or Statistical.  Symmetry within
these approaches is necessary to ensure quality, robustness and
retargetability.  In reality, such symmetry, whether it is in terms of
structural transfer lexicons, interlingual dictionaries or parallel
corpora, is a major bottleneck in developing any MT system.

This dissertation presents an approach to MT that addresses the lack
of symmetry by exploiting symbolic and statistical target language
resources in source-poor/target-rich language pairs. This approach is
called Generation-Heavy Hybrid Machine Translation (GHMT). Expected
source language resources include a syntactic parser and a simple
one-to-many translation dictionary. No transfer rules or complex
interlingual representations are used. Rich target language symbolic
resources are used to overgenerate multiple structural variations from
a target-glossed syntactic dependency representation of source
language sentences.  Statistical target-language resource are then
used to select amongst the overgenerated translations.  The
source-target asymmetry of systems developed in this approach makes
them more easily retargetable to new source languages.

The contributions of this research include: (1) a new model for
machine translation that transcends the need for large amounts of
symmetric knowledge while maintaining a high degree of robustness,
quality, and retargetability; (2) a systematic framework for handling
translation divergences that uniformly accommodates a wide range of
seemingly different divergence types and their interactions; (3) a
hybrid (symbolic-statistical) generation approach that expands the
concept of symbolic overgeneration to include conflation and
head-swapping of structural variations; (4) the introduction and use
of structural n-grams on a large scale in natural language generation;
and (5) the creation of several resources that have been used by other
researchers including an extensible MT system for translating into
English and a large-scale categorial variation database for English.

An extensive evaluation suggests that GHMT is more robust and has
superior output quality, in terms of grammaticality and accuracy, relative to a
primarily statistical approach.}

\setboolean{hasfigures}{true}
\setboolean{hastables}{true}
\setboolean{hascopyright}{true}

\spacing{1.667}
\begin{document}

\makefrontmatter

%\tableofcontents

\chapter{Introduction}

\section{Motivation}
Machine translation (MT) systems can be classified in terms of their
complexity and their approach.  Complexity refers to the place of the
system on the MT pyramid: word-gist, transfer or interlingua (IL).
Approach refers to the place of the system on the hybridity continuum:
symbolic, hybrid or statistical.  These classifications are
theoretically equally valid choices for building an MT system.
Practically, however, other criteria are more influential in making
decisions on the kind of system to build.  There are four such
criteria: Quality, Cost, Retargetability and Robustness.

Quality refers to a system's translation quality.  It can
be further dissected into Accuracy (or Fidelity) and Fluency (or
Intelligibility).  Accuracy is a measure of how much of the
information in the source language (SL) is carried in translation to
the target language (TL).  Fluency is a measure of how TL-like is the
output of the MT system.  Fluency can further be divided into specific
aspects of the TL output: grammaticality and lexical choice of TL
sentences.  A prevalent MT problem relevant to translation quality is
Translation Divergence, cases where the underlying concept or ``gist''
of a sentence is distributed differently in two languages.  Examples
include {\it \textbf{making improvements in} one's life} versus {\it
\textbf{improving}} one's life, {\it \textbf{swimming across} the
lake} versus {\it \textbf{crossing} the lake \textbf{swimming}} or
{\it \textbf{passing} the night \textbf{laughing}} versus {\it
\textbf{laughing} the night \textbf{away}}.

The second criterion is the cost of building an MT system specifically
in terms of time it takes to build resources and use them.  Third,
retargetability refers to how reusable the different modules and
resources of an existing system are.  Retargetability and cost are
quite related.  The more retargetable a system, the more it is worth
building and the more distributed the cost can be over several future
systems.  Finally, robustness refers to how well a system does in
terms of quality when used on new kinds of input such as a new genre that
the system is not trained on or does not have resources for.

These criteria interact with each other and with the complexity and
approach of the MT system under consideration.  A gist-based system
allowing multi-word translation can be built to provide high quality
translation if enough phrases and clauses are included in a
translation dictionary.  The size of the dictionary and the cost of
building it would be prohibitively high (theoretically infinite).  A
transfer approach producing the same level of quality would cost less
to build although the complexity of the involved resources would be
higher.  This is the tradeoff between cost and complexity while
maintaining a constant level of quality.  Additionally, a gist-based
approach requires resources that are not retargetable for any other
kind of MT system or any other pair of languages.  A transfer approach
has a similar problem at the level of transfer-lexicon building.  But
SL parsers and TL generators, both necessary modules for transfer MT,
can be independent of the translation lexicon and thus are
retargetable.  An approach with higher complexity such as interlingua
requires more complicated lexicons to translate SL to IL and IL to TL.
But such components are highly retargetable since they are
language-pair independent.

The complexity of an MT system is independent of its approach.  The
symbolic approach has been used for all levels of
complexity. Currently, most primarily statistical systems are low in
complexity (multi-word gist-based with movement). Syntax-based
statistical MT \cite{Yamada:2001} is an exception.  There is currently
no IL-based statistical MT.  The attractive simplicity of statistical
approaches comes at the cost of development and annotation of training
data, parallel corpora that are not retargetable to other language
pairs.

The cost-complexity tradeoff while maintaining quality applies to
statistical approaches as well.  For a statistical system to be able
to handle a wide range of phenomena, it needs to train on such
phenomena.  Syntax-level information can allow a statistical model to
abstract linguistic knowledge with a smaller parallel corpus.  But
such a corpus needs to be syntactically annotated, which comes at a
higher cost.  Additionally, statistical systems have a negative effect
on robustness since the bottom-up type of information derived from a
parallel corpus is very related to the specific genre of the corpus.
Therefore, such information is less likely to apply well to other
input genre.

In the context of the work presented in this dissertation, it is
necessary to add another dimension to classifying MT systems: symmetry,
which refers to the amount of resources and modules used for analysis
and generation in an MT system.  The MT pyramid assumes such symmetry
to be inherent in any MT system.  At the word-gist level, a
translation dictionary is a perfectly symmetrical resource dedicated
and coupled with the two specific languages it translates between.  At
the transfer level, the output of SL parsing and the input of TL
generation are expected to be of the same level of syntactic
complexity.  Additionally, the transfer lexicon is a more
sophisticated symmetrical translation dictionary.  This symmetry also
carries to the interlingual level.  As for statistical approaches,
parallel corpora are, by definition, symmetrical resources.
 
The symmetry assumption is a major bottleneck in developing any
MT system in a world of asymmetrical language resources.  There are few
languages that have extensive resources and that have been well studied
linguistically. Resource-richness can be seen in the amount of
available digital text, the size of lexical semantic resources such as
word-nets, the availability of IL lexicons, and the existence of
annotated tree-banks.  The problem of symmetry is twofold: resources
need to be built for resource-poor languages while already existing
rich resources for rich languages are not usable.  The cost of
enforcing symmetry affects the different criteria discussed above
(with different degrees depending on the complexity and approach of the MT system): lower
retargetability, quality and robustness and higher cost.

The MT approach presented in this dissertation, Generation-Heavy
Hybrid Machine Translation (GHMT), is an asymmetrical hybrid approach
to MT that has a higher degree of retargetability and robustness
compared to other symmetrical approaches that either have a much
higher cost (symbolic approaches) or a much lower quality on new input
genre (statistical approaches).

\section{Generation Heavy Machine Translation}

GHMT addresses the lack of symmetry by exploiting symbolic and
statistical TL resources in source-poor/target-rich
language pairs.  Expected SL resources include a
syntactic parser and a simple one-to-many translation dictionary. No
transfer rules or complex interlingual representations are used. Rich
TL symbolic resources such as word lexical semantics, categorial
variations and subcategorization frames are used to overgenerate
multiple structural variations from a TL-glossed syntactic dependency
representation of SL sentences. This symbolic overgeneration, which
accounts for possible translation divergences, is constrained by
multiple statistical TL models including surface n-grams and
structural n-grams.  The source-target asymmetry of systems developed
in this approach makes them more easily retargetable (re-{\it
source}-able) to new source languages (provided a SL parser and
translation dictionary).
 
The basic intuition of the approach parallels the experience of most
language learners whose lack of symmetrical knowledge impairs their
ability to translate into their newly learned language but does not
hinder them as much when translating from the foreign language into
their native tongue (where they are assisted by rich resources).

\begin{figure}[htb]
\centerline{\includegraphics[width=4.5in]{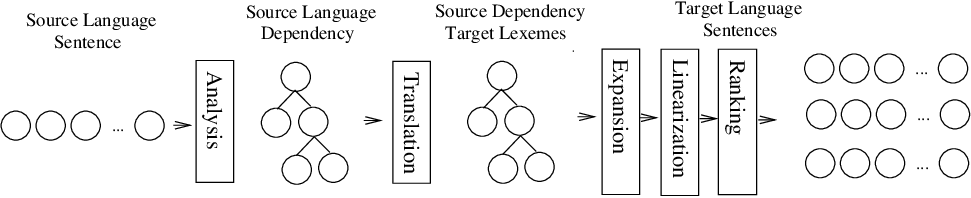}}
\caption{Generation-Heavy Machine Translation}
\label{simpleghmt}
\end{figure}

Figure~\ref{simpleghmt} presents an overview of this approach.  The
input is a SL sentence that is analyzed to produce a dependency tree,
a representation describing the syntactic relations among the words in
the sentence.  For example, in a Spanish-English GHMT, the Spanish
sentence {\it Maria puso la mantequilla en el pan (Mary put the butter
on the bread)} is analyzed as follows:

\begin{sentences}
\item \label{c0}\begin{verbatim}
(puso :subject Maria
      :object  (mantequilla :modifier la)
      :modifier (en :object (pan :modifier el))) 
\end{verbatim}
\end{sentences}

This dependency tree specifies that {\it Maria} is the subject of the
verb {\it puso} and that {\it mantequilla} is the object.  In the
translation step, each of the Spanish words in the dependency tree are
mapped into sets of English words:

\begin{sentences}
\item \label{c1}\begin{verbatim}
((lay locate place put render set stand) 
      :subject Maria
      :object  ((butter bilberry) :modifier the)
      :modifier ((on in into at) 
                    :object ((bread loaf) :modifier the))) 
\end{verbatim}
\end{sentences}

During the expansion step, different variants of \ref{c1} are
generated using lexical semantic information and other
English-specific heavy resources.  The following are only a few of these variants:

\begin{sentences}
\item \label{c2}\begin{verbatim}
(put  :subject Maria
      :object  ((butter bilberry) :modifier the)
      :modifier (on :object ((bread loaf) :modifier the))) 
(lay  :subject Maria
      :object  ((butter bilberry) :modifier the)
      :modifier (at :object ((bread loaf) :modifier the))) 
(butter
      :subject Maria
      :object ((bread loaf) :modifier the)) 
(bread
      :subject Maria
      :object  ((butter bilberry) :modifier the))
\end{verbatim}
\end{sentences}

The first two examples show little difference in structure from the
Spanish structure in~\ref{c1}, but the last two are much different.
In the linearization step, the dependency trees in~\ref{c2} are
converted into a word lattice compressing multiple possible sentences:

\begin{sentences}
\item \label{c3}\begin{verbatim}
(OR (SEQ Maria (OR puts put) the (OR butter bilberry) 
         (OR on into) (OR bread loaf))
    (SEQ Maria (OR lays laid) the (OR butter bilberry) 
         (OR at into) (OR bread loaf))
    (SEQ Maria (OR butters buttered) the (OR bread loaf))
    (SEQ Maria (OR breads breaded) the (OR butter bilberry)))
\end{verbatim}
\end{sentences}

These different sequences are then ranked using a statistical language
model.  The overgenerated variants score higher than direct word
translations, e.g., the top-ranked output in this example is {\it
Maria buttered the bread}.

\begin{sentences}
\item \label{c4}\begin{verbatim}
Maria buttered the bread         -47.0841
Maria butters the bread          -47.2994
Maria breaded the butter         -48.7334
Maria breads the butter          -48.835
Maria buttered the loaf          -51.3784
Maria butters the loaf           -51.5937
Maria put the butter on bread    -54.128
\end{verbatim}
\end{sentences}

\section{Matador}

A large-scale end-to-end GHMT system for Spanish-English translation,
named Matador, is created as part of this thesis.  The
implementational details of Matador are presented later.  An
evaluation of Matador's translation quality is conducted using the IBM
Bleu metric \cite{Papineni:2001} and comparing against three
systems---simple gisting, primarily statistical (IBM Model 4) and purely
symbolic (Systran)---over three corpora (UN, FBIS and Bible).  The
evaluation shows that although Matador scores lower than IBM Model 4
on the corpus where all language models were trained (UN), Matador has
a higher degree of robustness and scores higher when tested on text
with new genre (Bible). Additionally, Matador's output quality, in
terms of grammaticality and accuracy, is superior to IBM Model
4. Systran and gisting are the best and worst respectively for all
corpora.  A second evaluation verifies the advantages of using
structural n-grams, which increase Matador's speed by 60\% with no
negative effect on text quality.  Error analysis of the system's
components suggests that the use of structural expansion is promising
but more research concerning the use of both statistical and symbolic
constraints is necessary to limit overgeneration.

\section{Contributions}

The contributions of this research include:
\begin{itemize}
  
\item A new model for machine translation that transcends the need for
large amounts of symmetric knowledge while maintaining a high degree of
robustness, quality, and retargetability.
\item A systematic framework for handling translation divergences that
uniformly accommodates a wide range of seemingly different divergence
types and their interactions.
\item A hybrid (symbolic-statistical) generation approach that expands
the concept of symbolic overgeneration to include conflation and
head-swapping of structural variations.
\item The introduction and use of structural n-grams on a large scale
in natural language generation.
\item The creation of several reusable resources including an
extendible GHMT system for translating into English (Exerge --
Expansive Rich Generation for English),
% a language-independent linearization engine (Oxygen),
and a large-scale categorial variation
database for English.  These resources are used by other researchers
working on projects such as ChinMT (LCS-based Chinese-English MT)
 \cite{Habash:2003b}, Headline Generation \cite{Zajic:2002}, and DUSTER
(Divergence Unraveling for Statistical Translation) \cite{Dorr:2002a}.
\item The implementation of Matador, an end-to-end GHMT system for
Spanish-English translation.
\item An evaluation of Matador's output quality in comparison to
several other approaches including fully symbolic (Transfer-based)
and fully statistical systems.
\end{itemize}

\section{Outline of this Dissertation}
The next chapter, Chapter~\ref{related-work-chapter}, presents a
literature review of related work in three areas: Natural Language
Generation, Lexical Semantics and Evaluation.  This is followed by a
discussion of the GHMT approach in
Chapter~\ref{ghmt-chapter}. Chapter~\ref{matador-chapter}
and~\ref{exerge-chapter} present implementational details and issues
of Matador and Exerge.  Chapter~\ref{eval-chapter} presents an
evaluation of Matador followed by a discussion of results.  Finally,
Chapter~\ref{last-chapter} presents contributions and future work.

%%%%%%%%%%%%%%%%%%%%%%%%%%%%%%%%%%%%%%%%%%%%%%%%%%%%%%%%%%%%%%%%%%%%%%
%%%%%%%%%%%%%%%%%%%%%%%%%%%%%%%%%%%%%%%%%%%%%%%%%%%%%%%%%%%%%%%%%%%%%%
%%%%%%%%%%%%%%%%%%%%%%%%%%%%%%%%%%%%%%%%%%%%%%%%%%%%%%%%%%%%%%%%%%%%%%

\chapter{Related Work}\label{related-work-chapter}

This chapter discusses related work and concepts to this dissertation.
There are three sections: (1) Natural Language Generation (NLG) in the
context of Machine Translation (MT); (2) Lexical Semantics; and (3)
Evaluation.  The Generation-Heavy Hybrid MT (GHMT) approach presented
in subsequent chapters is a statistical-symbolic hybrid machine
translation model that is generation-heavy.  This model expands on
work done in hybrid NLG systems such as Nitrogen
\cite{Langkilde:1998b}, Halogen \cite{Langkilde:2000,Langkilde:2002}
and Fergus \cite{Bangalore:2000a,Bangalore:2000b}.  It also defines a
systematically uniform solution to the translation divergence problem
without using interlinguas or transfer rules
\cite{Dorr:1993,Nasr:1997}.  GHMT uses lexical semantic knowledge
extensively such as thematic role hierarchies \cite{Levin:1996},
limited lexical functions \cite{Melcuk:1988}, verb classes
\cite{Levin:1993b,Dorr:2001a} and some information from Lexical
Conceptual Structure (LCS) representations
\cite{Dorr:2001a,Dorr:2001b}.  A major component of this research is
an extensive evaluation of translation quality that compares several
models with varying levels of complexity. The evaluation metric used
is a modified n-gram precision system from IBM called Bleu
\cite{Papineni:2001}.

\section{Natural Language Generation} 

Natural Language Generation (NLG) is the process of constructing
natural language output from {\it non-linguistic} input
 \cite{Jurafsky:SALP-20}.  These inputs can be sophisticated structures
of conceptual representations ({\it meaning}) or simple numeric
contents from some database that are useless to a human unless {\it
synthesized} into natural language form.  In its broadest definition,
NLG is process-independent of other applications that are usually
associated with it: machine translation, summarization, etc.
Moreover, general NLG is concerned not only with the generation of
single sentences, but also with the generation of proper discourse
(the relationship between sentences). However, for the purpose of this
work, NLG will refer only to sentence-level generation in the MT context.

\subsection{Components of Natural Language Generation}
NLG has been traditionally divided into three successive tasks
 \cite{Reiter:1994}:
\begin{itemize}
\item Content Determination: Map the initial input into a semantic
form.  Determine which components of the input should be realized and
which should be left for the human consumer to infer \cite{KBMT-6}.

\item Sentence Planning: Map conceptual structures into linguistic
structures.  This includes: text structuring, lexical selection,
syntactic selection, generating referring expressions, choosing
grammatical relations and grouping information into clauses and
sentences \cite{KBMT-6,Reiter:1994}.

\item Surface Generation (Realization): Map abstract linguistic
specification (syntax and function words) into surface form that
communicates this information.
\end{itemize}

Generation in MT does not require a determination
stage since the analysis stage of an MT system serves as the content
determiner for the generation side \cite{KBMT-6}.  This work will
focus on two aspects of NLG: Lexical Selection and Realization.
Lexical selection is concerned with the choice of open class words to
use to generate a concept or a source-language word (e.g. {\it eat},
{\it devour} or {\it ingest}; {\it kick} or {\it give a kick}).
Realization includes many components such as mapping thematic
structure into syntactic roles, controlling syntactic paraphrasing and
alternations, preventing overgeneration, providing defaults, propagating
agreement features, selecting closed class words, providing linear
precedence constraints among syntactic constituents, inflecting open
class words, and linearizing the syntactic tree into a string of
inflected words \cite{Elhadad:1996}.

\subsection*{Approaches to Natural Language Generation}

The different systems for NLG and MT can be divided into three classes
based on how much emphasis in the system building process was put on
hand-crafted rules and solutions versus statistically collected
information: symbolic approaches, statistical approaches and hybrid
approaches.

\subsection{Symbolic Approaches}
Symbolic solutions to NLG depend on manually creating a system with
rules reflecting an understanding of the relationship between input
and output. The most basic symbolic approach to NLG, and by far the
most common, is Template Filling.  Templates can be the best choice
for very domain-specific interfaces such as ATM machines or junk mail
labels; however, they are extremely limited for large-scale
domain-independent (or even domain-specific) applications.\footnote
{An excellent discussion of the advantages of Natural Language
Generation technology over template filling is discussed in
 \cite{Reiter:1995}.}

As for the more realistic systems concerned with larger domains and
robust behavior, two major approaches to symbolic NLG are Systemic
Grammar and Functional Unification Grammar \cite{Jurafsky:2000}.

Systemic Grammar is a part of Systemic Functional Linguistics, a
theory of language as a means of communicating semantics,
interpersonal meaning, and contextual effect \cite{Halliday:1985,Melcuk:1988}.
Within this approach, a generation grammar is constructed as a system
network of realization statements (grammar rules) that specify
mappings from input features to syntactic positions and surface forms.
Two widely used examples of this approach are the Penman generation
system \cite{Penman:1989} and RealPro \cite{Lavoie:1997}.

Functional Unification Grammar (FUG) uses unification of input
representations, called Functional Descriptions (FD), and a 
%Functional
Unification Grammar as its main process for generation decision
making \cite{Elhadad:1995,Elhadad:1992}. The original functional
unification formalism was put forward by Kay \cite{Kay:1979}.

As far as both approaches go, there is no consensus on the format of
the input.  It ranges from propositional content (functional
specification) to grammatical/lexical specifications.  The complexity
of selection and linearization depends on the type of input expected.
Both approaches require multiple passes through the grammar to make
decisions at multiple levels.  Both use defaults and functional
categorizations to handle input underspecifity.

%-----------------------------------------------------------------------
%-----------------------------------------------------------------------
%-----------------------------------------------------------------------

\subsection{Statistical Approaches}

Statistical natural language processing is concerned with the creation
of computer programs that can perform language-processing tasks by
virtue of information gathered from (typically large) corpora. Usually
this information is in the form of statistics, but it may be distilled
into other forms, such as dictionary entries, or various kinds of
rules \cite{Charniak:2000}.  In the context of statistical machine
translation and statistical natural language generation, the object is
to create a statistical translation model, a mathematical model in
which the process of input to output translation is statistically
modeled.  Two examples of statistical approaches are Candide (MT) and
IBM Air Travel Reports (NLG).

%---------------------------------------------------------------------
\subsubsection{Candide}
Candide is a system for French-English Machine Translation developed
at IBM \cite{Brown:1990,Brown:1993,Berger:1994}. Candide uses a
probability model of the translation process developed using methods
from information theory and statistics.  Its main tools are (1) the
noisy channel model of communication, (2) a parametric probability
translation model, (3) a parametric probability language model and (4)
an assortment of numerical algorithms for training such
models.\footnote{For a detailed discussion of the mathematical models
and the training issues of collecting them, see \cite{Brown:1993}. But
for a great introductory manual to the mathematics of statistical MT,
see \cite{Knight:1999b}.}  The basic idea is that the channel converts
a sequence of words in one language (English) into another
(French). The channel operations are movements, duplications, and
translations applied at each word independently.  Movement is
conditioned only on word classes and positions in the string.
Duplication and translation are conditioned only on word identity.
Parameter probabilities for the different operations are collected
from sentence aligned corpora and are used to calculate the most
probable English sentence $\hat{e}$ given a French sentence $f$:
\[ \hat{e} = \argmax_{e} P(e|f) = \argmax_{e} P(f|e)\cdot P(e) \]
This formula maximizes the probability from two models: $P(f|e)$ is a
translation model and $P(e)$ is a language model.

The translation model $P(f|e)$ returns the probability of the French
input given some {\it divined} English output.  In the translation
process, this model is used in the creation of a large set of possible
English outputs that are then constrained by the language model.  The
following are the different models explored in Candide.  They increase
in complexity as they introduce new parameters and restrictions to
make the training process as realistic and correct as possible.  For
example, while Model 1 focuses on word translation only, later models
add parameters for handling multiple word translations (fertilities)
and categorical information (Model 4 Class-based Alignment).  These
models are trained in succession using the (expectation-maximization)
EM algorithm \cite{Dempster:1977} with the results of one model being
passed onto the next as initiating values.

The language model $P(e)$, in principle, returns the probability of
occurrence of a string $e$. The language model is important in
constraining the output to only valid English sequences.  The language
model probability can be calculated as the product of the
probabilities of each word in $e$ given its history:

\[ P(e) = P(e_1 ... e_l) = P(e_1)\cdot P(e_2|e_1)\cdot P(e_3|e_{1}e_{2}) ... \]

For practical reasons, the history is usually limited to 1 or 2
previous words (respectively, bigrams and trigrams).  A logarithmic
sum of the probabilities is used to maintain numerical accuracy by
avoiding numerical underflow.  So, for a sentence, $S$, that is a
sequence of words,$w_i$, bigram and trigram probability is calculated
as follows:

 \[ bigrams: \log P(S) = \sum_{i} \log P(w_i|w_{i-1}) \]
 \[ trigrams: \log P(S) = \sum_{i} \log P(w_i|w_{i-1},w_{i-2}) \] 

Candide uses a smoothed trigram model that combines unigrams, bigrams,
trigrams and a basic uniform probability of the inverse of the number
of tokens in the language.  The training of the translation language
model was done using over two million English-French sentence pairs
from the Canadian Hansards (parliament proceedings).

Translation in Candide does not take place between raw French and
English texts.  During training, text is transformed into an
intermediate normalized form to suppress French-English lexical
variations and reduce vocabulary size. This was accomplished through
preprocessing operations such as text segmentation, name and number
detection, morphological normalization and word order normalization
through series of word rewriting and reordering rules.

The translation (Decoding) process is referred to as
Analysis-Transfer-Synthesis \cite{Berger:1994}.  First, analysis
converts French text into intermediate French (same as in training
earlier).  Transfer decoding happens in two steps: (1) a candidate
set of sentences is created using a coarse version of the translation
and language models; then, (2) the system expands this set and uses the
more sophisticated models to score it.  This two-step approach
simplifies the search to some extent.  The output of this step is
Intermediate English that is converted into English output using an
English synthesis step.  The ``normalization'' done in Candide's
Analysis-Transfer-Synthesis uses a lot of symbolic knowledge of
properties of the two involved languages.  However, no sophisticated
syntactic or semantic knowledge is used.

Candide was revolutionary in that it scored very comparatively to
Systran in the ARPA evaluation and, sometimes, even better than
Systran \cite{Berger:1994}.

%------------------------------------------------------------------
\subsubsection{IBM Air Travel Reports}

This project was produced at IBM as part of the generation component
of their automated air travel agent.  This work is interesting because
it was the first to automatically learn mappings from semantics to
surface realization by searching for parameters and learning them
automatically without using a hand crafted grammar
 \cite{Ratnaparkhi:2000}.  What made this possible is a limited domain
with simple semantic representation (simple set of {\it attribute-value}
pairs) and an available semantically annotated corpus.

Three models were implemented with varying degrees of complexity: NLG1
used template frequency; NLG2 used sequential n-gram model; and NLG3
used dependency n-gram model.  All these models ignore attribute
values and train only on template words and attribute names.  The
results of a comparative evaluation of these models showed that the
sequential n-gram model fared better than the baseline model but the
dependency n-gram model improved accuracy the most.

This system is at best a proof of concept on the use of statistical
techniques for Semantic-to-Surface mapping. It works well for
domains with similar complexity to air travel.  However, the
techniques used in it are useless for large-scale domain-independent
NLG applications especially with the lack of large scale semantically
annotated corpora.
%------------------------------------------------------------------
\subsubsection{Structural Language Models}\label{struct-models}
Three types of statistical language models are distinguished
\cite{Daume:2002}: (1) surface n-gram language models, (2)
non-lexicalized syntax (structural) models and (3) lexicalized syntax
models. The first type has been used as part of various hybrid
implementations of NLG and MT tasks
\cite{Knight:1995,Brown:1995,Langkilde:1998a,Bangalore:2000c,Habash:2003b}.
There has been a lot of research using structural language models in
parsing \cite{Collins:1997a,Charniak:2000,Simaan:2000}.  In
comparison, only one large-scale system built with NLG in mind uses a
structural language model \cite{Bangalore:2000c}.  The dependency
n-gram model mentioned in the previous section uses templates and
focuses on travel reports only \cite{Ratnaparkhi:2000}.  However, a
recent study using the Charniak parser \cite{Charniak:2000} as a
lexicalized syntax model for generation purposes demonstrated the
usefulness of these models in a variety of NLG tasks
\cite{Daume:2002}.

%-----------------------------------------------------------
\subsection{Hybrid Approaches}

Hybrid NLG systems combine symbolic and
statistical techniques to maximize the advantages and minimize the
disadvantages of these different approaches.  Most
statistically-oriented systems sacrifice depth or accuracy to breadth
of coverage.  They are powerful but do not provide deep analyses and
may make occasional mistakes that are hard to fix.  Statistical
techniques work as well as they do because one can get a great deal of
leverage from relatively weak (and easily collected) pieces of
information.  Also, at the shallower linguistic depths, computers seem
better at collecting linguistic phenomena than people are at thinking
them up \cite{Charniak:2000}.  However, linguistically blind
statistical operations such as permutations and fertility are also
problematic in that they can create search spaces that are large
and yet exclude correct answers. 
%I am not sure about that any more... GHMT does worse at expanding too much
To statistically train for these more complex phenomena requires very
large (the more the better) corpora annotated with semantic, syntactic
or bilingual information.

Symbolic systems are good for small domains.  Large scale systems,
however, need large amounts of knowledge -- lexical, grammatical and
conceptual. Most available knowledge bases are filled with gaps
that compromise their robustness in handling input with missing pieces
of knowledge \cite{Knight:1995}.  Therefore, restrictive defaults or
large sets of rules are needed.  These defaults and rules must be
created manually and are restricted by domain.  Also defaults can
often be wrong and disfluent (e.g. definiteness defaulting to {\it the} and 
tense to {\it past}).

Hybrid systems try limiting the space of ambiguity resulting from
linguistically blind statistical hypotheses using linguistically
motivated rules.  But they depend on statistical knowledge to rank and
make the final choice.  One simple example of the advantage of using
hybrid techniques is the generation of an article for an English noun
when there is some ambiguity about definiteness.  So, the system is
faced with making a choice among four options: ({\it the}, {\it a},
{\it an} and $\phi$).

A purely statistical solution would be to overgenerate all
possibilities and depend on a bigram model for picking up the best
option.  Bigrams are excellent at selecting between 'a' and 'an' since
the important context window for this choice is always a bigram.
However, bigrams are not good at picking definite versus indefinite
articles.  Moreover, they are most likely to pick $\phi$ because
shorter sentences are more probable! That can be an extremely bad
choice.

A purely symbolic system can distinguish between 'a' and 'an' with
special modules that will have to encode knowledge about the
phonological representation of the following noun or adjective.
Deciding on definite versus indefinite articles can be done, but it
requires a lot of rules.  Most symbolic systems use defaults that may
not always be appropriate in such cases.  However, a simple rule using
number (singular/plural) and conflated definiteness (e.g. proper
nouns) can easily decide when to pick $\phi$ versus an article.

A hybrid approach takes advantage of both: for example, using a rule
to decide between $\phi$ and the articles.  Then, it generates all
three articles in the second case and leaves the final word to
the statistical model.

%------------------------------------------------------------------
%------------------------------------------------------------------
%------------------------------------------------------------------
\subsubsection{The Nitrogen Generation System}

Nitrogen was the first hybrid NLG system
 \cite{Langkilde:1998a,Langkilde:1998b,Langkilde:1998c}.  It was
developed at the Information Science Institute at the University of
Southern California as part of JAPANGLOSS
 \cite{Knight:1994b,Knight:1995a}, a Japanese-English newspaper machine
translation system.

\begin{figure}[htb]
\centerline{\includegraphics[]{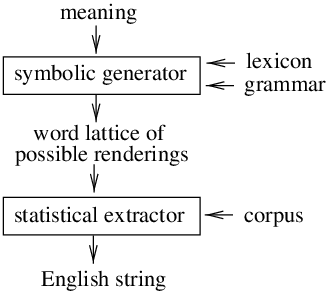}}
\caption{The Nitrogen Generation System}
\label{nitrogen}
\end{figure}

As shown in Figure(\ref{nitrogen}), the system breaks the generation
process into two steps: overgeneration and statistical extraction.
The first step is implemented symbolically using a simple generation
grammar. The second step is implemented using a statistical
uni/bigram model of the English language.

The input to Nitrogen is an Abstract Meaning Representation (AMR).  An
AMR is a labeled directed graph using the syntax of the PENMAN
Sentence Plan Language (SPL) \cite{Penman:1989}.  The nodes of an AMR
are concepts from the Sensus ontology
 \cite{Knight:1994a}.\footnote{This is a knowledge base of 70 thousand
nodes derived from several sources such as WordNet, Longman dictionary
and Penman upper model \cite{Knight:1994a}.}  AMR labels are a mix of
semantic and syntactic relations such as :agent, :patient and :source,
and also :subject, :object and :mod.  The following is an example AMR
to represent the sentence ``the man will eat chicken'':

\begin{sentences}
\item
\begin{verbatim}
(n1 / |eat,take in|
  :AGENT (n2 / |man|)
  :PATIENT (n3 / |poulet|)
  :TIME FUTURE)
\end{verbatim}
\end{sentences}

The slash mark specifies an instance of something (the man, the eating,
and the chicken).

The Nitrogen symbolic generator uses morphological and syntactic
knowledge to transform the input AMR into a word lattice that
efficiently compresses the space of possible paths.\footnote{A typical
newspaper sentence of 30 words or such can have on the order of
millions to trillions of paths associated with it.} Morphological
knowledge is implemented using a morphology derivation grammar that
handles both derivations and inflections.  Nitrogen overgenerates and
depends on the statistical extractor to discard bad cases.  For
example, one morphology rule creates a plural ending {\it -xes\/} and
{\it -xen\/} for all nouns ending with {\it -x\/}.  This generates
{\it boxes\/} and {\it oxen\/} but also {\it *boxen\/} and {\it
*oxes\/}.  Syntactic knowledge in Nitrogen is implemented using a
grammar database that contains two types of transformation rules:
linearization and recasting.  Linearization rules transform an AMR
into a word sequence or several ambiguous sequences. Recasting rules
transform an AMR into another AMR by redefining the original relations
or AMR structure.  The Nitrogen grammar is organized around the
semantic input rather than the syntax of English as opposed to
symbolic systems such as FUF/SURGE or RealPro.

The resulting word lattice is passed on to a statistical language
model of English to rank and extract the most fluent paths.  The
language model estimates unigram and bigram probabilities from a large
collection of 46 million words of the Wall Street Journal.

Nitrogen's greatest strengths are its simplicity, coverage and
robustness.  The simplicity of the system can be seen in its simple
grammar and morphology that allow for an incremental grammar building
approach (mindless overgeneration constrained only by statistical
extraction and the grammar is refined only if needed).  The n-gram
language models required for Nitrogen's operation are also very simple
to collect once corpora are available.  Because it is lexically
driven (albeit at the uni/bigram level), Nitrogen has a great
coverage of lexical constraints not explicitly represented in lexicon
for selection of open-class words.  Nitrogen's defaulting to word
choice that is more common is a robust technique to make better more
fluent last choice than symbolically defined defaults.

Nitrogen's simplicity is however, its biggest weakness: (1)
interactions between simple rules in Nitrogen's grammar need to be
controlled to get correct reasonable output; (2) the simplicity of the
grammar results in an explosive overgeneration requiring inefficient
use of time and space (3) dependencies between non-contiguous words
cannot be captured by bigrams alone and the simplicity of the grammar
does not compensate for that.  For example, the lack of subject-verb
agreement allow for such bigram-fluent sequences as ``I only hires men
who is good pilots''; (4) there is a preference for shorter sentences
(violating the balance of conciseness versus clarity) which results in
dropping of function words such as ``the'' and in preferring shorter
paths in general; and finally (5) Nitrogen statistical extraction
component has a search cutoff parameter that limits the size of the
search space which can lead to non-optimal choices.

One other issue that Nitrogen shares with all systems dealing with
semantic representation is that its input definition might be too
complex or too specific for retargetability.

\subsubsection{Oxygen Linearization}

The Nitrogen grammar formalism is unification based and it provides a
small number of tools to recast and linearize AMRs.  There are several
limitations to the use of this formalism.  For example, the grammar is
interpreted which results in inefficient time/space use.  Another
limitation is that the tools provided are rather simple
transformations which causes the linearization grammars to be long and
complex. The linearization engine, Oxygen addresses these issues
\cite{Habash:2000}.  Oxygen is an efficient language-independent
linearization engine.  Linearization grammars for Oxygen are written
using Oxyl, a powerful linearization grammar description language that
has the power of a programming language with the focus on natural
language linearization.  Oxyl grammars are compiled into programs that
run independently.

The power of Oxyl is accomplished by providing recasting mechanisms
for the most common needs of a linearization grammar and also by
allowing embedding of code in a standard programming language (Lisp).
The Oxyl linearization grammars are also simple, clear, concise and
easily extendible.  The simplicity of Oxyl grammars is apparent when
one considers issues of redundancy: the handling of ambiguities at
every phrase rule is hidden from the linearization grammar designer
and is treated only in the compiler and support library. For a
detailed presentation of Oxyl's syntax, see \cite{Habash:2001b}.

\subsubsection{Halogen: Forest-Based Statistical Sentence Generation}

A later version of Nitrogen, Halogen, addresses the issue of
time-space inefficiency by compressing the search space into {\it
forests} \cite{Langkilde:2000,Langkilde:2002}.  In this approach, the
input, the AMR semantic representation, is mapped into
alternative phrases that are represented as packed sets of syntactic
constituency trees or forests.  Similar subtrees are linked
symbolically instead of being repeated as in the lattice approach used
in Nitrogen and described in \cite{Knight:1995}.  These forests are
then statistically ranked using a bottom-up dynamic programming
algorithm and the statistics are bigram or trigram based.

The advantages of forests include (1) more compactness than lattices;
(2) representation of hierarchical information relevant to syntactic
representation; and (3) more efficient statistical ranking with
respect to space and time use.  Although Halogen is more efficient, it
still is unable to guarantee picking the optimal solution in longer
sentences.

%-----------------------------------------------------------------------
\subsubsection{Unification Based Glossing}

Unification Based Glossing is a hybrid MT approach that combines
syntax-based unification style interpretation with statistical
processing \cite{Hatzivass:1995}.  Whereas Nitrogen expects
source-language text to be parsed and composed into semantic
representation before it can be generated; glossing computes a target
language word lattice directly from the source-language parse tree.

The symbolic component of this approach, the {\it glosser}, takes as
input a constituency tree of the source language.  The glosser
features transfer-like rules that map source words and syntactic
structures directly into target lattices. The resulting word lattice
is passed through a uni/bi/trigrams statistical extractor to select
the most fluent paths.

%------------------------------------------------------------------------------
\subsubsection{FERGUS}
FERGUS (Flexible Empiricist/Rationalist Generation Using Syntax)
extends the use of n-gram models with a tree-based statistical model
and a traditional tree-based syntactic grammar
 \cite{Bangalore:2000a,Bangalore:2000b,Bangalore:2000c}.  Modeling
syntax in FERGUS is done using an existing wide-coverage grammar of
English, the XTAG grammar developed at University of Pennsylvania
 \cite{XTAG:1999}.  XTAG is a Tree Adjoining Grammar (TAG)
 \cite{Joshi:1987} that has been extended to include lexicalization and
unification-based feature structures \cite{XTAG:1999}.  In TAG, the
elementary structures are phrase tree structures that are composed
using two operations: (1) substitution (which appends one tree at the
frontier of another) and (2) adjunction (which inserts a tree into the
middle of another).  Morphosyntactic constraints such as subject-verb
agreement and subcategorization can be specified within the tree
structure associated with the lexical items. There is no distinction
between the Grammar and Lexicon in lexicalized TAGs.

FERGUS is composed of three modules: the Tree Chooser, the Unraveler,
and the Linear Precedence (LP) Chooser.  The input to FERGUS is an
unordered dependency tree labeled only with lexemes but no syntactic
annotation.  First the Tree Chooser uses a stochastic tree model to
choose TAG trees for the nodes in the input tree. Then the Unraveler
uses the XTAG grammar to produce a lattice of all possible
linearizations that are compatible with the semi-specified input tree
and XTAG. Finally, the LP Chooser selects the most likely
traversal of this lattice given the language model.  The tree model
used by the Tree Chooser and the trigram model used by the Linear
Precedence Chooser are both based on one million words from the Wall
Street Journal corpus.

FERGUS is different from Nitrogen in many ways: First of all, the
input to FERGUS is much shallower and closer to the target-language
surface form than Nitrogen's semantic representation.  Although both
systems are hybrids, they have a different balance of statistical and
symbolic components with FERGUS employing more power both
statistically and symbolically: FERGUS constrains its lattice of
possible paths by symbolically handling agreement and other morphology
issues using the XTAG component. Nitrogen, on the other hand, uses an
explosively overgenerating simple grammar.  Statistically, FERGUS
uses both a stochastic tree model and a trigram language model whereas
Nitrogen uses bigrams only.

%------------------------------------------------------------------------

\subsubsection{Syntax Based Statistical Translation}\label{sbst}

The IBM statistical MT models \cite{Brown:1990,Berger:1994} did not
statistically model structural or syntactic aspects of language or the
translation process.  This might not be important for languages that
are structurally close (e.g. English and French) but it is necessary
for structurally divergent language pairs such as English and
Japanese.

The first attempt to a syntax-based statistical translation model was
achieved by \cite{Yamada:2001}.  In this approach, the noisy channel
model is used.  But instead of converting sequences of words, it
accepts constituency parse trees as input.  The channel performs the
following operations on each node in the tree: (1) reordering of child
nodes, (2) insertion of extra words at each node, and (3) translation
of leaf words.  Reordering helps translating languages with different
word order; while insertion allows for adding post-nominal case marker
particles (for Japanese and Korean).  These operations are very
reminiscent of the IBM noisy channel operations of movement,
duplication and translation. Just like Candide, some preprocessing
of the input is needed (in addition to syntactic parsing): since
relative order is important for the channel operations,
sub-constituents are flattened by combining projections of phrase
heads to make the relative order more accessible: For example, (NP (VB
NP)) is flattened into (NP VB NP).

The system uses three types of parameter tables to collect frequencies
for reordering, insertion and translation operations.  Reordering
probabilities of the children of a node are assumed to be dependent on
the sequence of child node labels.  Insertion probabilities combine
position (right, left, nowhere) and word insertion likelihood and are
dependent on parent and node labels. Translation probabilities specify
the likelihood of a word translating into another.  The final
probability of a tree equals the product of the total reordering,
insertion and translation probabilities each of which is computed as
the product of all sub-tree probabilities.

Parameter estimation was done on a corpus of 2,000 pairs of English
and Japanese sentences using the EM algorithm \cite{Dempster:1977}.  An
evaluation on the task of generating Viterbi alignments compared
this syntax-based model with IBM model 5.  The results show that the
syntax-based model fared better than model 5 and had more perfect
sentences in general.

Although simple and ambitious, syntax-based statistical MT has a major
bottleneck: the availability of lots of correctly analyzed parallel
data for training.

%--------------------------------------------------------------------------
%--------------------------------------------------------------------------

\section{Lexical Semantics}

Lexical Semantics is the linguistic study of the relations among words
and their meanings as well as the internal structure of individual
words \cite{Jurafsky:2000}.  This includes the study of argument
structure, thematic roles, selectional restrictions, semantic
relations, ontologies, lexical conceptual representations, event
structure and qualia structure \cite{Saintdizier:1994}. This review
will address four sub-topics of lexical semantics that are relevant to
the this research: Lexical Relations, Predicate Argument
Structure, Syntax-Semantics Linking, and the use of lexical semantics
to handle translation divergences.

\subsection{Lexical Relations}

Lexical relations describe relative relationships amongst
words. According to \cite{Cruse:1986}, lexical relations are either
hierarchical taxonomic relations (such as hypernymy, hyponymy and
entailments) or non-hierarchical congruence relations (such as
identity, overlap synonymy and antonymy). Lexical relations are
reviewed here to highlight their relationship to the categorial
expansion and word class components in GHMT.  Next,
three distinct approaches to lexical relations are discussed: WordNet, lexical
functions and verb classes.

\subsubsection{WordNet}
WordNet is the most well-developed and widely used lexical database of
English \cite{Fellbaum:1998}.  In WordNet, both types of lexical
relations are specified among words with the same part of speech
(verbs, nouns, adjectives and adverbs).  The most used feature of
WordNet is its synsets (synonym sets), where synonyms are grouped
together.  They are useful in representing meaning (instead of
hard-to-define {\it primitives}) and in expanding the generative space
of a given word.  WordNet has been used by many researchers to build
or extend knowledge bases such as the SENSUS knowledge base
 \cite{Knight:1994a} and the Lexical Conceptual Structure database
 \cite{Green:2001a}.  Both were used for Machine Translation.  In
 \cite{Bangalore:2000c}, WordNet synsets were used to {\it fake}
meaning ambiguity as part of system evaluation.  In the scope of this
work, one criticism of WordNet is its lack of cross-categorial links,
such as verb-noun or noun-adjective relations.

\subsubsection{Lexical Functions}
Mel'\v{c}uk's Meaning-Text-Theory \cite{Melcuk:1988} defines a lexical
combinatorial zone that specifies semantically related lexemes through
Lexical Functions (LF), simply defined as a correspondence between a
{\it key} lexical item and a set of lexical items.  There are two
types: paradigmatic and syntagmatic \cite{Ramos:1994}.  Paradigmatic
LFs associate a lexical item with related lexical items.  The {\it
relation} can be semantic or syntactic. Semantic LFs include
Synonym(calling) = {\it vocation}, Antonym(small) = {\it big}, and
Generic(fruit) = {\it apple}.  Syntactic LFs include
Derived-Noun(expand)= {\it expansion} and Adjective(female) = {\it
feminine}.  Syntagmatic LFs specify collocations with a lexeme given a
specified relationship.  For example, there is a LF that returns the
light verb that takes the LF's {\it key} as its object:
Light-Verb(attention) = {\it pay}.  Other LFs specify certain semantic
associations such as Intensify-Qualifier(escape) = {\it narrow} and
Degradation(milk) = {\it sour}.

Lexical Functions have been used in Machine Translation and
Generation.  For example, \cite{Ramos:1994} used LFs to determine
coreference in text generation.  For example, in {\it John was
painting his walls. Suddenly (a,*the) fly got stuck on (the,*a)
brush}, there is an LF that relates {\it brush} (but not {\it fly}) to
{\it paint} and thus help determine that one is definite but not the
other.  The use of LFs in \cite{Ramos:1994} is a classic example of
symbolic techniques for natural language generation but it is beyond
the scope of this work (even to consider it in a hybrid context).

\subsubsection{Verb Classes}
Linguists have long noted the expression and interpretation of a verb's
arguments is largely determined by its meaning. So, verb behavior can
be used to understand pertinent aspects of verb meaning
 \cite{Levin:1993b}.  For example, the Locative Alternation ({\it X V Y
on Z} and {\it X V Z with Y}) appears with the verb {\it spray} but
not {\it cover} or {\it put} although all three are semantically
related:
\begin{sentences}
\item
\begin{tabular}{lll}
 X spray Y on Z   &  X put Y on Z   & *X cover Y on Z \\
 X spray Z with Y & *X put Z with Y &  X cover Z with Y\\
\end{tabular}
\end{sentences}

There are two specific components of the meaning that are both present
only in {\it spray}; whereas the other two verbs have only one of the two
 \cite{Levin:1996}.  These components are: (1) causing stuff to go
on a surface and (2) affecting a surface by applying stuff to it.
Verbs that have both components are considered part of the same verb
class as {\it spray} (e.g. brush, drizzle, splash).  They are expected
to allow both syntactic realizations.  This is useful in analysis too.
For example, {\it it drizzled all morning} has a syntactic realization
that is not shared with spray verbs which means, correctly, its
semantics is different.  In this case {\it drizzle} is part of the
Weather verb class such as {\it rain} and {\it snow}.

%-------------------------------------------------------------------
\subsection{Predicate Argument Structure}

Every predicate is associated with an argument structure that
specifies the number of arguments it requires \cite{Grimshaw:1990}.
Arguments are those elements which necessarily must be involved in the
action or state described by the predicate \cite{Saintdizier:1994}.
Adjuncts are other elements that add to the meaning but are not
necessarily involved.  Arguments headed by a preposition are called
{\it obliques}.  For example, in {\it X give Y to Z on A for B},
$give$ is a predicate, $X$, $Y$ and $Z$ are all arguments, $Z$ is an
oblique and $A$ and $B$ are adjuncts.

There are many levels of complexity to the representation of predicate
argument structure.  The simplest form is the specification of the arity
of a predicate without any semantic or syntactic information about the
arguments \cite{Saintdizier:1994}.  For example, give:3, think:2,
sleep:1. Next in complexity is the specification of the different
relations between a predicate and its arguments.  There are two
approaches that describe such relations: logical and thematic
 \cite{Saintdizier:1994}.

\subsubsection{Logical Approach: Subcategorization Frames}
 The logical approach orders arguments with a rigid underlying
meaning.  This order, also called a {\it subcategorization frame},
relates arguments across different syntactic variations and
alternations.  For example, {\it clear} has the subcategorization frame
{\it (arg1 arg2)}: {\it arg1 cleared arg2}, {\it arg2 clears easily},
{\it arg2 was cleared by arg1}, etc.

The problem with this approach is that it makes no generalizations
about the behavior of arguments across verbs.  {\it Arg1} can only be
understood in the context of its verb head, e.g. the {\it clearer},
{\it giver}, etc.  This information is very important for
cross-lingual applications where two verbs that are translations of
each other would define the meaning of their {\it args} differently,
e.g. {\it gustar} (please) and {\it like}.  Another example involving
obliques and different alternations is the English {\it John(1) filled
the glass(2) with water(3)} but not {\it *John(1) filled water(2?) in
the glass(3?)}  both of which are allowed in Korean.

Although this representation is limited, its simplicity makes it
theory and system-independent.  This is the reason why the PropBank
project uses this approach \cite{Palmer:2001}.

\subsubsection{Thematic Approach: Thematic Roles}
In the thematic approach, arguments are assigned thematic labels (also
called thematic roles, semantic roles or theta roles) that abstract
away from the specifics of deep semantic roles by generalizing over
similar roles across classes of verbs \cite{Jurafsky:2000}.  This is
by far the most common approach in the field to represent
predicate-argument structure.  There are a lot of variations with
little agreement even on terminology: e.g. case
frames \cite{Fillmore:1968}, theta-grids \cite{Stowell:1981}, and
thematic relations \cite{Jackendoff:1972}.  However, all thematic role
lists are based on the same set of ideas: (1) they are semantically
unanalyzable, (2) they are defined independently of the verb
(predicate), and (3) they are small in size \cite{Levin:1996}.  Unlike
subcategorization frames, thematic roles have no significant order.
They are intended to describe the relationships between arguments and
predicates. For example, {\it give} has three thematic roles
associated with its three arguments: agent (giver), theme (given) and
goal (givee).  Since theta roles are abstractions, their actual
meaning is hard to pin down.  For example, Agent is doer or causer,
Patient is undergoer or affected, Theme is changing participant or
undergoer and Goal is target or final state.

There are several problems with thematic role lists \cite{Levin:1996}.
First, there lacks a reliable diagnostic for determining the role and
real meaning of an argument.  Syntactic positions (subject, object,
and prepositions) are not helpful: For example, a subject can be an
agent {\it John killed Paul}, a theme {\it Paul died}, a goal {\it
Paul received a bullet in the chest}. Prepositions can limit the
possibility of the thematic roles to be assigned but are not
unambiguous.  For example, the object of {\it to} is never a source,
but the object of {\it with} can be an instrument {\it John killed
Paul with a gun}, a theme {\it John loaded the gun with bullets} or a
companion {\it John left with Paul}.  Second, determining the proper
granularity level for the definition of a role is hard.  Cruse
decomposes {\it agent} into four types (volitive, effective,
initiative and agentive) to distinguish between the agents of {\it
encourage, avoid, facilitate} and {\it confirm} \cite{Cruse:1973},
whereas Dowty suggests defining roles by means of clusters of features
rather than criteria \cite{Dowty:1991}.  He defines a continuum of
opposing features between what he calls proto-agent and proto-patient.
Properties contributing to proto-agency include volition, sentience,
causation and independent existence.  As for proto-patiency, they
include change of state, causal affectedness, relative movement and
dependent existence.  Third, plain lists of thematic roles lack
internal organization.  For example, source, goal, route and direction
theta roles are related to each other in that they all involve a path:
e.g. respectively {\it come to, come from, come through and come
toward}.  However no organization exists in the theta roles themselves
that shows such a relationship.

Faced with these problems, some researchers concluded that thematic
roles should be avoided altogether \cite{Ravin:1990}.  However, other
researchers attempted to deal with some of these issues by creating
more complex representations of predicate-argument structures: Event
Structures.

\subsubsection{Event Structure}
Event structures attempt to explain/define thematic roles of arguments
by enriching predicates with a complex internal structure. There are
three kinds of event structure schemata: the Localist, Aspectual and
Causal \cite{Levin:1996}. All of them share distinction between
semantic structure and semantic content in event structure
decomposition \cite{Grimshaw:1993}.  For example, Run (a race) and jog
share the same structure but different content, whereas Run (a race)
and Run (a store) share the same manner content while differing in
structure.  Predicate event structures are composed using two basic
blocks of primitive predicates and constants.  Primitive predicates
are used to define the semantic structure of the predicate and
constants are used to define its semantic content.

\paragraph{Localist Approach}
The Localist approach claims that events involving motion and location
in space are central to the construction of all events
\cite{Jackendoff:1972,Jackendoff:1976,Jackendoff:1983,Jackendoff:1990}.
Jackendoff's Lexical Conceptual Structure (LCS) uses motion/location
primitives such as {\it GO, BE, STAY, TO, TOWARD, FROM} and also {\it
CAUSE} and {\it LET}.  Jackendoff also defines different fields of
action to differentiate between verbs that would share similar
localist structure such as identificational, possessional,
circumstantial, etc.  For example, {\it John went into the room} and
{\it John changed into a frog} both share a similar structure of [GO
John [TO John [AT room/frog]]], but the first is in the locational
field while the second is in the identificational field.  The LCS
approach unifies substructures of different verbs to give a consistent
account of what brings about a thematic role.  For example, {\it the
rock fell} and {\it john dropped the rock} both share a substructure
that describes the rock's movement: [LET john \ul{[GO rock [TOWARD
down]]}].  Dorr's implementation of LCS to handle machine translation
divergences is discussed further in Section~\ref{divergences}.

\paragraph{Aspectual Approach}
In the aspectual approach, events are classified into 4 classes
according to their lexical aspect or {\it aktionsart}: states,
activities, achievement, and accomplishment.  This approach can make
predictions about the behavior of different verbs and can help
constructing better semantic interpretations for ambiguous
cases \cite{Levin:1996}.  For example,
accomplishments but not activities can take an /in/ time adverbial:
{\it *Sam worked in an hour} and {\it Sam wrote the report in an
hour}.

\paragraph{Causal Approach}
The Causal approach decomposes events into simple non branching causal
chains: {\it Sam broke the window} = (1) Sam acted on window (2)
window changed state (3) window is in the result state {\it broken}.
This approach can capture subtle distinction between {\it X load Y with
Z} and {\it X load Z on Y} \cite{Croft:1991}.

\paragraph{}
The rest of this dissertation will focus on the localist approach
using Dorr's implementation of Jackendoff's Lexical Conceptual
Structures only.  This is for the practical reason that LCSes have
been shown to be useful in solving Machine Translation and generation
problems (see Section~\ref{divergences}) and they also have a lot of
resources available in the LCS formalism.  Currently, there are no
translation or generation implementations using the other approaches.

%---------------------------------------------------------------
\subsection{Linking Semantics and Syntax}

Linking semantics and syntax is a complex problem where any solid
solution must consistently explain a wide variety of syntax-semantic
interaction phenomena such as alternations and divergences
(monolingual and cross-lingual).  First, alternations are variant ways
in which the arguments of a predicate are realized on the surface.  A
linking algorithm should be able to predict this many-to-many mapping:

\begin{sentences}
\begin{small}
\item John$_{agent}$ opened the\_door$_{theme}$ with a\_key$_{instr}$\\
%the\_door$_{theme}$  opened a\_key$_{instr}$\\
the\_door$_{theme}$ opened\\
a\_key$_{instr}$ opened the\_door$_{theme}$\end{small}
\end{sentences}

In the examples above, the theme maps into subject or object positions,
and agent and instrument both can map into a subject position.  Note
that systematic transformations on the syntactic level such as
passivization and wh-questions are not included in the discussion of
alternations.

Second, divergences are variant syntactic realizations of concepts
with related semantics.  Monolingually, a semantics-syntax linking
theory is preferred to be consistent over semantically related
predicates.  For example, Psych verbs alternate the positions of their
arguments {\it John fears Death} and {\it Death frightens John}.
Another example is verbs related on a deeper level through an exchange
relationship such as {\it buy} and {\it sell}.  Cross-lingual
(translation) divergences will be discussed in detail in the next
section.

The amount of research on the semantics-syntax interface is sizable yet
with little agreement.  The following are the main dimensions of
variations of different approaches in the field.

\begin{itemize}
\item {\bf Representation:} The first dimension of variation of
linking approaches is the semantic and syntactic representations being
linked to/from.  These representations vary in form and depth.  For
example, syntactic representations can be dependencies or
constituencies (form) and they can be shallow or deep with traces or
without traces (depth).  Here is a list of different semantic
representations ordered from deep to shallow and complex to
simplistic: decompositional semantics \cite{Katz:1972,Ravin:1990},
lexical conceptual structure
 \cite{Jackendoff:1972,Jackendoff:1976,Jackendoff:1983,Jackendoff:1990,Dorr:1993},
abstract meaning representation
 \cite{Langkilde:1998a,Langkilde:1998b}, logical form, semantic roles
list, and lexico-syntactic structure \cite{Hale:1993}.

%------------------------------------------------------

\item {\bf Thematic Relevance:} In most of the approaches that link
semantics to syntax, out of all semantic information
available, only thematic roles are considered relevant to the linking process.
Such approaches adhere to two principles: Principle of Syntactic
Relevance ({\it only theta roles out of all semantics are interesting
for syntax}) and Correlation Principle ({\it theta roles of a
predicate determine syntactic configurations}) \cite{Ravin:1990}.  A
more specific statement of these two principles is Baker's
Uniformity of Theta Assignment Hypothesis(UTAH): {\it Identical
thematic relationships between items are represented by identical
structural relationships between those items a the level of
d-structure} \cite{Baker:1988}.  Two examples of thematic-based
linking are Fillmore's Case theory \cite{Fillmore:1968,Fillmore:1977}
and Chomsky's Government Binding Theory \cite{Chomsky:1981}.

Alternatively, Ravin claims that the principles of Syntactic Relevance
and Correlation are too severe and fail to explain a lot of syntactic
phenomena.  She states that all theories using thematic roles exhibit
a variety of syntactic problems that are addressed by implicitly or
explicitly depending on decompositional analysis of meaning that ends
up rendering thematic roles as mere labels of superfluous
value \cite{Ravin:1990}.  Her proposed approach is a non-thematic model
based on Katz's theory of semantic decomposition \cite{Katz:1972}. The
approach uses a two level representation model, one of syntax and one
of semantics.

%------------------------------------------------------
\item {\bf Mapping Scheme:} The type and complexity of a mapping
scheme depend heavily on the starting semantic/lexico-syntactic
representation. Mapping schemes are classified on two dimensions:
modularity (direct versus mediated) and preserving constraint type
(prominence versus equivalence-class) \cite{Levin:1996} .

Direct mapping approaches are simple reassignments of thematic or
structural information.  For example, agents are mapped into subjects
and themes into objects.  Mediated mapping approaches can be broken
into two classes: thematic-hierarchy-based and
intermediate-role-based.  A thematic hierarchy is a ranking of
possible semantic roles chosen because it figures in the statement of
linguistic generalizations \cite{Levin:1996}. 
 
\hide{[put example from journal paper here]}

Intermediate role mapping approaches map finer semantic roles into
coarser ones that are easier to map to syntax.  For example, in role
and reference grammar the more semantic roles are mapped using a
thematic hierarchy into two macro roles: Actor and Undergoer which map
directly into subject and object \cite{Vanvalin:1993}.  Dowty's
feature-based proto-roles are another example of intermediate-role
mapping.  Proto-agent and proto-them map directly to subject and
object positions.

Prominence preserving mappings assume a hierarchy of prominence in
syntax and semantics that must be preserved.  For example, agents are
more prominent than themes semantically and therefore are linked to
subjects which are more prominent than objects on the syntactic level.
Prominence of arguments can be explained in predicate structural terms in
of a relativized UTAH (Identical relationships are represented by
identical relative hierarchical relations between items at
D-structure) \cite{Larson:1990}.  The simplest example of this approach
is Thematic Hierarchies.  Equivalence class preserving mappings assume
that each distinct class of arguments is associated with a distinct
syntactic realization.  Equivalence classes can be structural
configurations (e.g. leftmost child of VP is subject) or semantic
features (e.g. agent is subject).  The difference between the
prominence and equivalence constraints is what they consider
important: relation of argument to verb individually (equivalence
class) or the whole hierarchical structure (prominence).

%------------------------------------------------------

\item {\bf Mapping Direction} Linking from syntax to semantics is more
complicated than linking from semantics to syntax since the former
maps from a shallower representation to deeper one.  The two processes
are asymmetrical.

\end{itemize}

There is a tradeoff among the granularity of the representation, the
complexity of the mapping scheme, and the coverage of linguistic
phenomena.  The deeper a representation is the more phenomena it can
unify and explain, but at the cost of a more complex mapping.  For
example, deeper semantic representations that obviate the need for
thematic roles have a finer granularity that is powerful but at a
much higher complexity cost.  Similarly, shallower representations
cannot be consistent in explaining the diversity of linguistic phenomena,
but the mapping schemes are very simple.

\subsection{Lexical Conceptual Structure} \label{lcs-section}
Lexical Conceptual Structure (LCS) is a compositional abstraction with
language-independent properties that transcend structural
idiosyncrasies.  LCS has been formulated by Dorr
 \cite{Dorr:1993,Dorr:1994} based on work by Jackendoff
 \cite{Jackendoff:1983,Jackendoff:1990,Jackendoff:1996}.  This
representation has been used as the interlingua of several projects
such as UNITRAN \cite{Dorr:1993b} and MILT \cite{Dorr:1997e}.

Formally, an LCS is a directed graph with a root.  Each node is
associated with certain information, including a {\em type}, a {\em
primitive} and a {\em field}.  The type of an LCS node is one of {\em
Event, State, Path, Manner, Property} or {\em Thing}.  There are two
general classes of primitives: {\em closed class} (also called {\em
structural primitives}, e.g., \verb|cause|, \verb|go|, \verb|be|,
\verb|to|) and {\em open class} primitives (also called {\em
constants}, e.g., {\tt john+, reduce+ed, jog+ingly}).  Suffixes such
as {\tt +, +ed, +ingly} are markers of open class primitives,
signaling also the type of the primitive (thing, property, event,
etc.). The structural primitive {\tt go},
denoting generalized movement, is distinguished from the constant {\tt go+ingly}: the
first appears in many lexical entries but the second appears only in
specific lexical entries such as the one for the English verb ``go''.
Examples of fields include {\tt Locational}, {\tt Possessional}, and
{\tt Identificational}.  Structurally, an LCS node has zero or more
LCS children.  There are three ways a child node relates to its
parent: as a subject (maximally one), as an argument, or as a
modifier.

An LCS captures the semantics of a lexical item through a combination
of semantic structure (specified by the shape of the graph and its
structural primitives and fields) and semantic content (specified
through constants).  The semantic structure of a verb is something the
verb shares with a {\it semantic verb class\/} whereas the content is
specific to the verb itself.  For example, all the verbs in the
semantic class of ``Run'' verbs (e.g., run, jog, walk, zigzag, jump,
roll, etc.) have the same semantic structure but vary in their
semantic content.  The {\it LCS Verb Database\/} (LVD)
 \cite{Dorr:2001a} extends the classification in English Verb Classes
and Alternations (EVCA) \cite{Levin:1993} by refining the class
divisions and defining the underlying meaning components of each class
in the LCS representation.  LVD also provides a relation between
Levin's classes and both thematic role information and hand-tagged
WordNet synset numbers.

Consider the sentence {\it John jogged to school\/}.  This can be
fully represented (except for features such as tense, telicity, etc.)
as shown in \ref{jog-clcs}, glossed as  `John moved ({\it location}) to
the school in a jogging manner':

\begin{sentences}
\item
\begin{center}
\begin{verbatim}
(event go loc 
   (thing john+) 
   (path to loc 
      (thing john+)
      (position at loc (thing john+) (thing school+)))
   (manner jog+ingly))
\end{verbatim}
\end{center}
\label{jog-clcs}
\end{sentences}

The lexicon entry for one sense of the English verb `jog' and the
preposition `to' are shown in Figure~\ref{lex1}.

\spacing{1}
\begin{figure}[htb]
\begin{tabular}{p{4.5in}}
\begin{alltt}
\small{ 
(DEFINE-WORD
  :DEF_WORD "jog"
  :CLASS "51.3.2.a.ii"
  :THETA_ROLES "_th,src(),goal()"
  :WN_SENSE (01315785 01297547)
  :LANGUAGE ENGLISH
  :LCS
   (event go loc (* thing 2) 
     ((* path from 3) loc (thing 2) 
        (position at loc (thing 2) (thing 4)))
     ((* path to 5) loc (thing 2) 
        (position at loc (thing 2) (thing 6)))
     (manner jog+ingly 26))
  :VAR_SPEC ((3 :optional) (5 :optional)))

(DEFINE-WORD
  :DEF_WORD "to"
  :LANGUAGE ENGLISH
  :LCS (path to loc 
         (thing 2) 
         (position in loc (thing 2) (* thing 6))))}
\end{alltt}\\
\end{tabular}
\caption{Lexicon Entries for {\it jog} and {\it to}}
\label{lex1}
\end{figure}
\spacing{1.667}

These entries include several pieces of information such as the root
form of the word, introduced by the field \verb-:DEF_WORD-, and the
word's semantic verb class, introduced by the field \verb-:CLASS-.

The field \verb-:THETA_ROLES- specifies the set of thematic roles
appearing in the RLCS entry. Theta roles preceded by an underscore
(\verb-_-) are obligatory; whereas roles proceeded by a comma
(\verb-,-) are optional.  Parentheses indicate that the corresponding
phrases must necessarily be headed by a preposition.  Sometimes the
specific preposition is provided inside the parentheses.  The roles
are ordered in a canonical order that reflects their relative surface
order: first available role is subject; second is object; etc.

The field \verb-:WN_SENSE- links the entry to 
its corresponding WordNet synset.  The Lexicon entries use
WordNet 1.6 senses \cite{Fellbaum:1998,Miller:1991}. 

The most important field in the lexicon entry is \verb-:LCS-, which
introduces the {\it Root LCS\/} (RLCS), the uninstantiated LCS
corresponding to the underlying meaning of the word entry in the
lexicon.  The top node in the ``jog'' RLCS in Figure~\ref{lex1} has
the structural primitive \verb|go| in the locational field.  Its
subject is marked with a ``*'' indicating that the node must be filled
recursively with other lexical entries during semantic composition.
The restriction on this particular LCS node is that the filler must be
of type {\tt thing}.  The number `2' in that node specifies the
thematic role: in this case, {\it theme}.  The second and third child
nodes are in argument positions filled with the primitives {\tt FROM}
and {\tt TO}.  The numbers `3' and `5' stand for {\it source} particle
and {\it goal} particle respectively.  The numbers `4' and `6' stand
for {\it source} and {\it goal}.  Figure~\ref{theta-list} contains a
list of variable numbers with their associated thematic roles.  The
second argument in the ``jog'' RLCS is the substructure {\tt (to loc
...)} that unifies with the RLCS for the preposition ``to''.  This
secondary RLCS itself has a star-marked argument that must be
instantiated with a {\tt thing} such as ``school''.

The variable specifications (indicated here as  \verb-:VAR_SPEC-) assign 
the arguments headed by {\tt FROM} and {\tt TO} an {\tt :optional} status.
Other possible variable specifications that appear in our lexicon 
include {\tt :obligatory}, {\tt :promote}, {\tt :demote}, 
{\tt :EXT} {\it (external)}, {\tt :INT} {\it (internal)} and {\tt :conflated} 
(see \cite{Dorr:1993b} for more details).

\spacing{1}
\begin{figure}[htb]
\caption{Inventory of Thematic Roles in LCS}
\label{theta-list}
\begin{scriptsize}
\begin{center}
\begin{tabular}{|p{0.2in}|p{1.3in}|p{2.4in}|}
\hline
\textbf{\#} & \textbf{Thematic Role} &  \textbf{Definition} \\
\hline
  0 &    & no thematic role assigned \\
  1 & AG & agent \\
  2 & TH ,EXP ,INFO & theme or experiencer or information\\
  3 & SRC() & source preposition\\
  4 & SRC & source \\
  5 & GOAL(), PRED()& goal or pred preposition \\
  6 & GOAL & goal\\
  7 & PERC()& perceived item particle\\
  8 & PERC & perceived item\\
  9 & PRED & identificational predicate \\
 10 & LOC() & locational particle \\
 11 & LOC & locational predicate\\
 12 & POSS & possessional predicate \\
 13 & TIME()& temporal particle preceding time \\
 14 & TIME& time for TEMP field \\
 15 & MOD-POSS()& possessional particle \\
 16 & MOD-POSS& possessed item modifier \\
 17 & BEN()& beneficiary particle \\
 18 & BEN& beneficiary modifier \\
 19 & INSTR()& instrumental particle \\
 20 & INSTR& instrument modifier \\
 21 & PURP()& purpose particle \\
 22 & PURP& purpose modifier or reason \\
 23 & MOD-LOC()& location particle \\
 24 & MOD-LOC& location modifier \\
 25 & MANNER()& manner \\
 26 &  & reserved for conflated manner  \\
 27 & PROP& event or state \\
 28 & MOD-PROP& event or state modifier\\
 29 & MOD-PRED()& identificational particle \\
 30 & MOD-PRED& property modifier \\
 31 & MOD-TIME& time modifier \\
\hline
\end{tabular}
\end{center}
\end{scriptsize}
\end{figure}
\spacing{1.667}

The current English lexicon contains 10K RLCS entries such as those in
Figure~\ref{lex1}.  These entries correspond to different senses of
over 4K verbs.  Figure~\ref{run-rlcs} compares three of the nine RLCS
entries for the verb ``run''.  These entries are classified by verb
class. Verb classes are used as templates to generate the RLCS entries
of verbs in the class. For example, the lexical entry for ``bake'' in
class 26.3 would be identical to the top RLCS entry shown in
Figure~\ref{run-rlcs}, except that node 9 would instead contain the
primitive {\tt bake+ed} rather than {\tt run+ed}.

\spacing{1}
\begin{figure}[ht]
\caption{Different RLCS Entries for ``run''}
\label{run-rlcs}
\begin{center}
\begin{tabular}{|p{4.3in}|}
\hline
26.3 Verbs of Preparing
\begin{alltt}\scriptsize{(event cause (* thing 1)
  (event go ident (* thing 2)
    (path toward ident (thing 2) 
      (position at ident (thing 2) (property run+ed 9))))
  ((* for 17) poss (*head*) (* thing 18)))}
\end{alltt}
Example: {\em John ran the store for Mary.}

Other verbs:\tiny{ bake boil cook fix grill iron mix prepare roast
 roll wash ...}  
\\ \hline
%47.7.a Meander Verbs (from to)
%\begin{alltt}\scriptsize{(event go_ext loc (* thing 2)
%  ((* path from 3) loc (thing 2) (position at loc (thing 2) (thing 4)))
%  ((* path to 5) loc (thing 2) (position at loc (thing 2) (thing 6))) 
%  (manner run+ingly 26))}
%\end{alltt}
%Example: {\em The river runs from the lake to the sea. }
%
%Other verbs:\tiny{crawl drop go meander plunge run  
% sweep turn twist wander ...  }
%\\
%\hline
47.5.1.b Swarm Verbs (Locational) 
\begin{alltt}\scriptsize{(event act loc (* thing 2)
   ((* position [at] 10) loc (thing 2) (thing 11))
   (manner run+ingly 26))}
\end{alltt}
Example: {\em The dogs run in the forest.}

Other verbs:\tiny{ bustle crawl creep  swarm swim teem ...}
\\
\hline
51.3.2.a.i Run Verbs - (Locational,Theme only) 
\begin{alltt}\scriptsize{(event go loc (* thing 2)
  ((* path from 3) loc (thing 2) (position [at] loc (thing 2) (thing 4)))
  ((* path to 5) loc (thing 2) (position [at] loc (thing 2) (thing 6)))
  (manner run+ingly 26))}
\end{alltt}
Example: {\em The horse ran into the field from the barn.}

Other verbs:\tiny{climb crawl  fly jog  jump leap  race  
 swim  walk ...}
\\
\hline
\end{tabular}\end{center}
\end{figure}
\spacing{1.667}

As described in \cite{Dorr:1993}, the meaning of complex phrases is 
captured through a {\it composed LCS\/} (CLCS).
A CLCS is constructed (or {\it composed\/}) from
several RLCS entries corresponding to individual words.  The composition
process starts with a parsed tree of the input sentence and maps
syntactic leaf nodes into RLCS entries whose argument positions are filled
with other RLCS entries.  For example, the two RLCS entries presented earlier
can compose together with the constants for ``John'' and 
``school'' to give the CLCS for the sentence: {\it John jogged to
school}, shown in \ref{jog-clcs}.  The star-marked node {\tt (* path
from 3)} is optional, and is left unfilled in this case. The same RLCS
could also be used to compose different CLCS representations (in combination with
other RLCS entries) to produce sentences like {\it John jogged from home}
or {\it John jogged from home to school}.

\subsubsection{Lexogen}
The LCS-based generation system, Lexogen \cite{Habash:2003b}, is
implemented to decompose and generate linguistic output from CLCS.
Lexogen is a hybrid system that uses symbolic resources such as a LCS
lexicon and an Oxygen grammar together with a statistical language
model (Nitrogen's Extractor).  Lexogen also uses two interlingual
representations: LCS and Nitrogen's AMR.  The two representations
serve different but complementary roles in the translation/generation
process. The deeper lexical-semantic expressiveness of LCS is
essential for language independent lexical selection that transcends
translation divergences \cite{Dorr:1993b}.  The shallower yet mixed
semantic-syntactic nature of AMRs makes it easier to use directly for
target-language realization.

\begin{figure}[htb]
\begin{center}
\begin{tabular}{p{4.5in}}
\centerline{\includegraphics[width=3.3in]{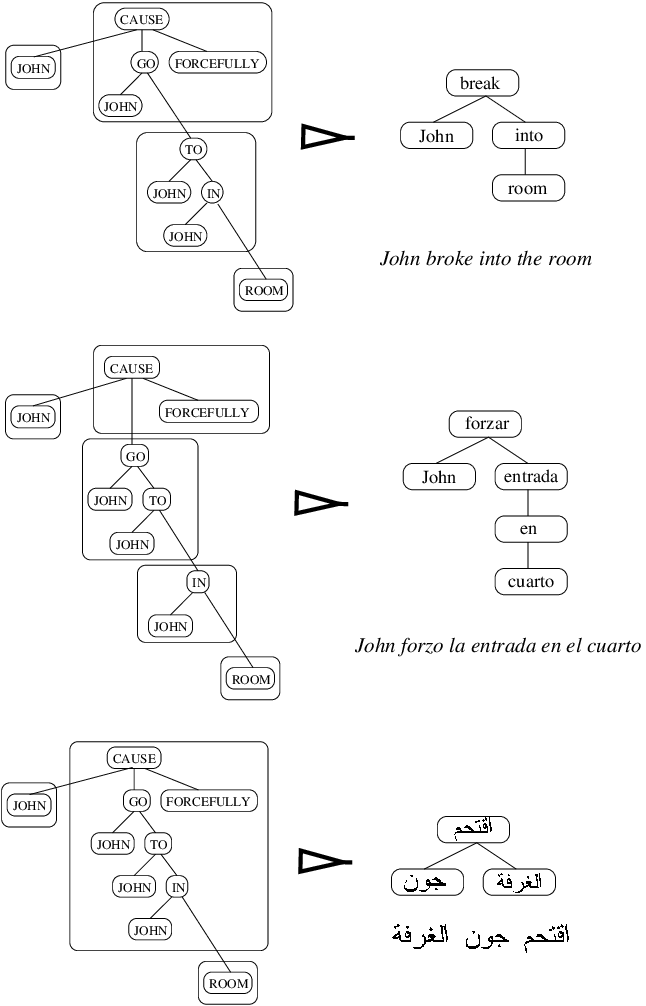}}
\end{tabular}
\caption{Different CLCS Decompositions into English, Spanish and Arabic} 
\label{clcs-decompose}
\end{center}
\end{figure}

A CLCS can be decomposed on the generation side in different ways
depending on the RLCS entries from the target language.
Figure~\ref{clcs-decompose} uses a compressed graphic representation
of LCS to visually compare three different decompositions in three
languages of a single CLCS.  The CLCS generated can be paraphrased as
{\it John caused himself to go to the inside of a room in a forceful
manner}.  

After decomposition, Lexogen uses a Universal Thematic Hierarchy (UTH)
to resolve the unordered nature of siblings under an AMR node and map
between AMR relations and their surface positions \cite{Habash:2001}. In the case of
thematic role ordering, the situation is more complicated by the lack
of one-to-one mapping between a particular thematic role and an
argument position.  For example, a theme can be the subject in some
cases and it can be the object in others or even an oblique. Observe
{\it cookie} in \ref{theme-map}.

\begin{sentences}
\item \label{theme-map}
\begin{subsentences}
\item John ate {\it a cookie} (object)
\item {\it the cookie} contains chocolate (subject)
\item she nibbled {\it at a cookie} (oblique)
\end{subsentences}
\end{sentences}

To solve this problem, a thematic hierarchy is used to  determine the argument position
of a thematic role based on its co-occurrence with other thematic roles.  
Several researchers have proposed different versions of thematic
hierarchies (see \cite{Jackendoff:1972,Carrier:1985,Bresnan:1989,Kiparsky:1985,Larson:1988,Giorgi:1984,Wilkins:1988,Nishigauchi:1984,Alsina:1993,Baker:1989,Grimshaw:1998}).\footnote{For
an excellent overview and a comparison of different thematic
hierarchies see \cite{Levin:1996}.} The hierarchy proposed in \cite{Dorr:1998a} differs from these in that it
separates (non-adjunct) arguments from obliques (i.e. adjunct arguments) and provides a more complete list of
thematic roles (31 roles overall) than those of previous approaches
(maximum of 8 roles).  See Figure \ref{theta-list} for a complete listing 
for the thematic roles used. The following is final thematic hierarchy 
for arguments.
\begin{sentences}
\item \label{final-hierarchy}
\begin{verbatim}
special case -- ag src th 
ext > ag > instr > th > perc > \*
\end{verbatim}
\end{sentences}

In the case of the occurrence of theme alone, it is mapped to first
argument position.  If a theme and an agent occur, the agent is mapped
to first argument position and the theme is mapped to second argument
position.  When an agent, a theme and a source co-occur, The order in
the hierarchy is violated as in {\it John$_{ag}$ charged Paul$_{src}$
\$40$_{th}$.} The term \verb-ext- is used to handle verbs that violate
the thematic hierarchy.  \verb-ext- refers to an externally
marked thematic role such as the perceived {\it John} in {\it
John$_{perc}$ pleases Mary$_{th}$} versus {\it Mary$_{th}$ likes
John$_{perc}$}.  This information is provided in the RLCS lexicon
entry using the special marker :EXT.  The use of the thematic
hierarchy eliminates the need to specify the thematic role to surface
position mapping in every verb lexicon entry.

As for the ordering of obliques, an ad hoc order was established:
 
\begin{sentences}
\item
\begin{verbatim}
particle > mod-prop() > perc() > th() > 
purp() > mod-loc() > mod-pred() > 
src() > goal() > mod-poss() > ben() 
\end{verbatim}
\end{sentences}

Note that the order of obliques is not a strict hierarchy but rather a
possible topological sort. A more detailed discussion is available 
in \cite{Dorr:1998a}.

Within LCS-based MT, the lexicon plays a central role
through providing symmetrical coverage of the mapping between target
and source languages {\it and} LCS.  The richness of LCS lexicon as
described earlier is very hard to create, debug and maintain
 \cite{Dorr:1992e,Dorr:1993}.

\subsection{Translation Divergences}\label{divergences}

%A translation divergence occurs when the underlying concept or ``gist''
%of a sentence is distributed over different words for different
%languages.  For example, the notion of floating across a river is
%expressed as {\it float across a river\/\/} in English and {\it cross a
%river floating\/} ({\it atraves\'{o\/} el r\'{\i}o flotando\/}) in
%Spanish \cite{Dorr:1993}.  

An investigation done by \cite{Dorr:2002a} found that divergences
occurred in approximately 1 out of every 3 sentences in a sample size
of 19K sentences from the TREC El Norte Newspaper Corpus. This
analysis was done on the TREC Spanish Data\footnote{LDC catalog no
LDC2000T51, ISBN 1-58563-177-9, 2000.}  using automatic detection
techniques followed by human confirmation.  Each
divergence type is described before turning to alternative approaches to handling
these in MT.

\subsubsection{Translation Divergence Types}

While there are many ways to classify divergences \cite{Dorr:1992a}, they
are presented here in terms of five specific divergence {\it types\/}
that can take place alone or in combination with other types of
translation divergences.  Table~\ref{transdiv2} presents these
divergence archetypes with Spanish-English examples.
The last column in Table~\ref{transdiv2} displays a percentage of
occurrences of the specific divergence type, taken from the first 48
unique instances of Spanish-English divergences from the TREC El Norte
corpus.  Note that there is often overlap among the divergence types
with the categorial divergence occurring almost every time there is any
other type of divergence. An extreme example of divergence type
co-occurrence is {\it Maria tiene gustos de pol\'{\i}ticos diferentes},
which can be translated as {\it different politicians please Maria}.
There are four divergence types in this pair: categorial
($gusto_{noun}$ to $please_{verb}$), conflational ($tener$ $gusto$ to
$please$), thematic ($Maria$ and $politicians$ switch syntactic roles)
and structural ($politican$ is an oblique in Spanish but an argument
in English).  This highlights the need for a systematic approach to
handling divergences that addresses all their different types and the
interactions amongst them rather than addressing specific cases one at
a time.

\spacing{1}
\begin{table*}
\begin{small}
\begin{center}
\begin{tabular}{|p{1.1in}|p{1.5in}|p{1.2in}|p{0.7in}|}
\hline
Divergence& Spanish  &English & Occurrence \\ \hline

\hline
Categorial &
X {\it tener hambre\/}

(X {\it have hunger\/})
&
X {\it be hungry\/}
& 
98\%
\\

\hline
Conflational &
X {\it dar pu\~{n\/}aladas a} Z

(X {\it give stabs to\/} Z)
&
X {\it stab\/} Z
&
83\%
\\
\hline

Structural &
X {\it entrar en\/} Y

(X {\it enter in\/} Y)
&
X {\it enter\/} Y
&
35\%
 \\
\hline

Head Swapping 
&
X {\it cruzar\/} Y {\it nadando\/}

(X {\it cross\/} Y {\it swimming\/})
&
X {\it swim across\/} Y 
&
8\%
\\
\hline

Thematic &
X {\it gustar a\/} Y 

(X {\it please to\/} Y)
&
Y {\it like\/} X
&
6\% \\
\hline
\end{tabular}
\end {center}
\caption{Translation Divergence Types}
\label{transdiv2}
\end{small}
\end{table*}
\spacing{1.667}

\paragraph{Categorial Divergence}
Categorial divergence involves a translation that uses different
parts of speech. This is by far the most common divergence type.
In the Spanish-English example
below, the light verb and noun phrase are translated as another light
verb and an adjectival form of the noun.
\begin{sentences}
\begin{small}
\item tener celos (to have jealousy)  $\Leftrightarrow$ to be jealous \\
 tener plena conciencia (have full awareness) $\Leftrightarrow$ to be fully aware 
\end{small}
\end{sentences}

\paragraph{Conflation}
Conflation involves the translation of two words using a single word that combines
the meaning of the two.  In Spanish-English translation, this divergence type has two forms:
light verb conflation and manner conflation.
Light verb conflation involves a single verb in one
language being translated using a combination of a semantically
``light'' verb, i.e., it carries little or no specific meaning in its
own right, and some other meaning unit (perhaps a noun) to convey the
appropriate meaning.  English light verbs include {\it give\/}, {\it
make\/}, {\it do\/}, {\it take\/}, and {\it have\/}. 

\begin{sentences}
\begin{small}
\item
dar una patada (give a kick)  $\Leftrightarrow$ to kick \\
poner fin (put end) $\Leftrightarrow$ to end  \\
tomar nota (take note) $\Leftrightarrow$  to note 
\end{small}
\end{sentences}

Manner conflation involves translating of a single manner verb (e.g.,
{\it float\/}) as a light verb of motion and a manner-indicating
content word that is typically a progressive manner verb in Spanish.

\begin{sentences}
\begin{small}
\item
to float $\Leftrightarrow$ ir flotando (go (via) floating)\\
to pass $\Leftrightarrow$ ir pasando (go passing)
\end{small}
\end{sentences}

\paragraph{Structural Divergence}

A structural divergence involves the realization of incorporated
arguments such as subject and object as obliques (i.e. headed by a
preposition in a PP) or vice versa.

\begin{sentences}\begin{small}
\item
 entrar en la casa (enter in the house) $\Leftrightarrow$ to enter the house \\
pedir un referendum (ask-for a referendum) $\Leftrightarrow$  ask for a referendum  
\end{small}\end{sentences}

\paragraph{Head Swapping}
\label{head-swapping-section}
This divergence involves the demotion of the head verb and the
promotion of one of its modifiers to head position.  In other words, a
permutation of semantically equivalent words is necessary to go from
one language to the other.  In Spanish, this divergence is typical in
the translation of an English motion verb and a preposition as a
directed motion verb and a progressive verb. 

\begin{sentences}
\begin{small}
\item
entrar corriendo (enter running) $\Leftrightarrow$ to run in \\
andar volando (go-about flying) $\Leftrightarrow$ to fly about 
\end{small}
\end{sentences}

\paragraph{Thematic Divergence}
A thematic divergence occurs when the verb's arguments switch thematic
roles from one language to another.  The Spanish verbs {\it gustar\/}
and {\it doler\/} are examples of this case. 

\begin{sentences}\begin{small}
\item
Me gustan uvas (to-me please grapes) $\Leftrightarrow$ I like grapes  \\
me duele la cabeza (to-me hurt the head) $\Leftrightarrow$  I have a headache 
\end{small}
\end{sentences}

\paragraph{Lexical Semantics and Translation Divergences}\label{lexsemtradiv}
Since translation divergences require a combination of lexical and
structural manipulations, they are traditionally handled minimally
through the use of transfer rules \cite{Han:2000,Lavoie:2000}.  A pure
transfer approach is a brute force attempt to manually encode all
translation divergences in a transfer lexicon \cite{Dorr:1999}.  Very
large parsed and aligned bilingual corpora have also been used to
automatically extract transfer rules \cite{Lavoie:2001,Watanabe:2001}.
This approach depends on the availability of such resources, which are
very scarce.

Alternatively, more linguistically-sophisticated techniques that use
lexical semantic knowledge to detect and handle divergences have been
developed. One approach uses Jackendoff's Lexical Semantic Structure
(LCS) \cite{Jackendoff:1983,Jackendoff:1990} as an interlingua
 \cite{Dorr:1993}.  LCS is a compositional abstraction with
language-independent properties that transcend structural
idiosyncrasies by providing a granularity of representation much finer
than syntactic representation.
% LCS has been used in several projects
%such as UNITRAN \cite{Dorr:1993b} and ChinMT \cite{Traum:2000}.  
As an example, the Spanish sentence {\it Juan cruza el r\'{\i}o
nadando\/} can be ``composed'' as the following LCS using a Spanish
LCS lexicon as part of the interlingual analysis:
\begin{sentences}
\item \begin{footnotesize}\begin{alltt}
[event CAUSE JOHN
   [event GO JOHN [path ACROSS JOHN [position AT JOHN RIVER]]]
   [manner SWIM+INGLY]]      
\end{alltt}\end{footnotesize}
\label{lcs-example}
\end{sentences}
In the generation phase, this same LCS is ``decomposed'' using English
LCS lexicon entries to yield {\it John swam across the river\/}.

Another approach enriches lexico-structural transfer at Mel'\v{c}uk's
Deep Syntactic Structure (DSyntS) level \cite{Melcuk:1988} with
cross-linguistic lexical semantic features \cite{Nasr:1997}.  Transfer
lexicon rules are written to capture generalizations across
the language pair instead of addressing specific paired instances.  As
an example, the following transfer rule can be used to handle the head
swapping divergence discussed in the last example.
\begin{sentences}
\item \begin{footnotesize}\begin{verbatim}
@TRANS_CORR
   @EN V1 [cat:verb manner:M]
       (ATTR Y [cat:prep path:P event:go] (II N))
   @SP V2 [cat:verb path:P event:go]
       (II N ATTR Z [manner:M])
\end{verbatim}\end{footnotesize}
\end{sentences}
Here, a transfer correspondence is established between the different
components of two DSyntS templates.  Note how the manner variable M
and the path variable P switch dominance.

A major limitation of these interlingual and transfer approaches
(whether using lexical semantics or modeling corpora) is that they require
a large amount of explicit symmetric knowledge for both SL and TL.

%------------------------------------------------------
%------------------------------------------------------
%------------------------------------------------------
%------------------------------------------------------

\section{Evaluation}
The evaluation of machine translation and natural language generation
systems is more of an art than a science. Evaluation of generation
systems is difficult, because the ultimate criteria is translation
quality, which can, itself, be difficult to judge. Moreover, it can be
hard to attribute specific deficits to the analysis phase, the lexical
resources, or the generation system proper.  A wide range of metrics
and techniques have been developed over the last fifty years to assess
`how good' a system is.  Evaluation schemas vary in their focus from
addressing the system's interface to system scalability, faithfulness,
space/time complexity, etc.  Another dimension of variation is human
versus automatic evaluation.  Fully automatic evaluation, a task that
is AI-complete {\it (i.e., encompassing all components of any system
that would be deemed ``intelligent'')}, is the ultimate goal in the
field.\footnote{For an excellent survey of machine translation
evaluation metrics and techniques, see \cite{Hovy:2000}.}

Evaluation is complex, labor-intensive, time-consuming and less
interesting than doing research.  But it is a very important
irreplaceable component of any healthy research practice
 \cite{Cole:1997}.  In the area of machine translation, it has been
pointed out that as much has been written about MT evaluation as about
MT itself \cite{Wilks:1994}.  The following three sections discuss
general evaluation types, MT evaluation metrics, and translation
quality evaluation metrics.

\subsection{Evaluation Types}

Three kinds of evaluations are defined based on appropriateness for
different goals \cite{Cole:1997}:

\begin{itemize}
\item Adequacy Evaluation (aka Evaluation proper): The goal of this
kind of evaluation is to determine the fitness of a system for a
purpose, i.e., does it do the job required?, how well?, at what cost?
Adequacy evaluations are focused on the users and their needs.

\item Diagnostic Evaluation (aka Diagnosis proper): The goal of this
kind of evaluation is to examine the system's performance profile
with respect to a space of possible inputs.  To that extent,
diagnostic evaluations require the creation of a {\it test suite} that
enumerates all the elementary linguistic phenomena addressed in the
input domain.  Once the test suite is constructed, comparisons of two
generations of the same system can be done automatically.  This type
of Diagnostic evaluation is called Regression testing. Diagnostic
evaluations are typically done by/for system developers.

\item Performance Evaluation (aka Assessment, Progress Evaluation,
Summaritive evaluation or Technology Evaluation): The goal of this
kind of evaluation is to measure the systems performance in one or
more specific areas. Performance evaluations are typically done for
developers and Research and Development program managers.

\end{itemize}

A fourth type of evaluation, Formative Evaluation, is a hybrid of
adequacy and performance evaluations. In formative evaluations, system
design is guided by the needs of the users.  During system
development, user trials of prototypes can help guide pinpoint
directions for research and development.  Other evaluation type
distinctions are Black-box/Glass-box and intrinsic/extrinsic
evaluations.  Black-box evaluations are interested in the behavior of
the system as a whole, while Glass-box evaluations are component-wise
evaluations. Within a Glass-box evaluation, one can distinguish
between intrinsic and extrinsic evaluations: intrinsic evaluations
focus on how a particular component is behaving in its own terms,
while extrinsic evaluations focus on how that component contributes to
the performance of the whole system.

To evaluate a system, it is important to distinguish between three
interdependent levels of specificity \cite{Cole:1997}:
\begin{itemize}
\item Criterion: what is the evaluation interested in? Speed,
error rate, precision, recall, fluency, accuracy,  etc.

\item Measure: which specific property of system performance should be reported
to get the chosen criterion?  For example, for speed report {\it
processes per second}, for precision report {\it the number of
retrieved relevant data to all retrieved data}, for recall report {\it
the number of retrieved relevant data to all relevant data}, etc.

\item Method: how to determine the appropriate value for a given
measure and a given system.  For example, post-analytic measurement of
the system's behavior over some benchmark task.
\end{itemize}

\subsection{MT Evaluation Metrics}
There are three categories of MT evaluation metrics: system-based,
text-based and cost-based \cite{Church:1991}.

\subsubsection{System-based Evaluation Metrics}

System-based metrics count internal resources: size of lexicon, number
of grammar rules, etc.  These metrics are easy to measure although
they are not comparable across systems.  Their value is
questionable since they are not necessarily related to utility.

In \cite{Elhadad:1996}, a list of qualitative criteria that fit within
this class of metrics was provided to compare different realization
systems. It included comprehensive coverage, prevention of
overgeneration, encapsulation of syntactic knowledge, expecting
regular input, allowing partially canned input, compact input,
extensibility, efficiency, robustness, user-friendliness, and
versatility.

\hide{
\begin{itemize}
\item 
Comprehensive Coverage: a large set of valid syntactic forms and
their alternations should be generated from a single input.
\item Prevention of overgeneration: only grammatical outputs should be
generated.
\item Encapsulation of Syntactic Knowledge: As little syntax as
possible should be required for the creation of input.
\item Regular input: input structure should be well-defined and regular
and thus easily producible.
\item Partially canned input: input where canned phrases coexist with
individual words should be accepted \cite{Reiter:1995}.
\item Compact Input: appropriate defaults should be provided for all
syntactic features so that input is as compact as possible.
\item Extensibility: new syntactic constructors should be easily added
and verified.
\item Efficiency: generation from input should be as fast as possible
\item Robustness: ill-formed or incomplete inputs should result in
grammatical fragments
\item user-friendliness: developers and users should be able to use
the system with little knowledge of theory or implementation details
\item Versatility: the realization component should be reusable in a
wide variety of system architectures and application domains.
\end{itemize}
}
\subsubsection{Text-based Evaluation Metrics}

Text-based metrics can be divided into sentence-based and
comprehensibility-based.  Sentence-based metrics examine the quality
of single sentences out of context.  These metrics include Accuracy,
Fluency, Coherence, etc.  Typically, subjects evaluating sentences are
given a description of the metric with examples and are asked to rate
the sentences on an x-point scale.  These scales range from 3-point to
100-point. Comprehensibility metrics measure the comprehension or
informativeness of a complete text composed of several sentences.  The
subjects are typically given questionnaires related to the processed
text.  Text-based metrics are much more related to utility than
system-based metrics, but they are also much harder to measure.  There
are some automatic text-based evaluation metrics that measure the
amount of post-editing needed for a sentence given a gold standard.
These are variations on edit-distance, i.e., the number of
deletions, additions or modifications measured by words or keystrokes
per page or sentence. These techniques are not necessarily related 
to utility; however, it was recently shown that the smarter tree-based edit
distance might actually correlate better to human judgment
 \cite{Bangalore:2000c}.

\subsubsection{Cost-based Evaluation Metrics}
Cost-based metrics evaluate a system on how much money/time
it saves/costs per unit of text, say a page.  These are secondary
metrics since they depend on other metrics to evaluate how much
post/pre-processing is necessary for a commercially functional system.

\subsection{Text Quality Evaluation Metrics}
A good measure of translation performance can be defined as {\it the
closer MT is to a professional human translation, the better it
is} \cite{Papineni:2001}.  But how can this {\it closeness} be
measured? The following two sections explore qualitative metrics and
automated metrics on different dimensions of this problem.

\subsubsection{Qualitative Text Quality Evaluation Metrics}

Qualitative evaluations of MT/NLG generally present a set of human
subjects with a sample of the output of the evaluated system and a
scaled metric.  The subjects are asked to assign scores on the metric
to the different outputs.  Qualitative evaluations vary on several
dimensions:
\spacing{1}
\begin{table*}
\begin{center}
\caption{Nagao's Fidelity Metric}
\label{nagao-fidelity}
%\begin{scriptsize}
\begin{tabular}{|p{0.4in}|p{4.6in}|}
\hline
1 & The entire content of the original text have been translated into
an easily readable form.\\
\hline
2& The entire content of the original text have been faithfully
reproduced and is easily understood in the translation, but minor
post-editing is required.\\
\hline
3 & The translation is generally faithful to the original, but some
reorganization of the word-order is required.\\
\hline
4 & The translation is generally faithful to the original, but there
are errors in the relationships between phrases, tenses,
singular/plural, position of adverbs, etc.  post-editing requires
structural transformations.\\
\hline
5 &  The structure and content of the original work are not well
preserved in translation, with potions of the original text
incorrectly translated or referents of words and phrases incorrect.\\
\hline
6 &  The structure and content of the original are poorly translated.
Phrases and clauses are missing but valid sentence structure is
achieved.\\
\hline
7 &  The translation reflects neither the structure nor the content of
 the original text.  Valid sentence structure is not achieved because
of missing subjects, predicates, etc.\\
\hline
\end{tabular}
%\end{scriptsize}
\end {center}
\end{table*}
\spacing{1.667}
\spacing{1}
\begin{table*}
\begin{center}
\caption{Nagao's Ease of Understanding Metric}
\label{nagao-understanding}
%\begin{scriptsize}
\begin{tabular}{|p{0.4in}|p{4.6in}|}
\hline 
1 &The sentence has a clear meaning, with no possibility of
misunderstanding. Appropriate grammar, terminology, and sentence
structure are employed, and post-editing is not necessary. \\
\hline
2 &The meaning of the sentence is clear, but there are slight problems
in the sentence structure, grammar, or terminology. Post-editing would
easily be accomplished by human hand. \\
\hline
3 &The overall meaning can be grasped, but there are small ambiguities
due to grammar or terminology.  Post-editing would best be done with
reference to the original work. \\
\hline
4 & The quality of the translation is poor, and there are many
problems of grammar and terminology.  Some understanding of the
sentence can be inferred after a lengthy study of the translation, but
it would be better to undertake the translation anew by human
translator than to post-edit the machine translation. \\
\hline
5& No understanding of the contents of the translated sentence is
possible.  Human translation is required.\\
\hline
\end{tabular}
%\end{scriptsize}
\end {center}
\end{table*}
\spacing{1.667}

\begin{itemize}

\item Verbose or Abbreviated Metrics.  Some metrics specify to their
subjects what is exactly meant by each score on the scale.  For
example, see Nagao's measures for Fidelity and Ease of Understanding
in Tables \ref{nagao-fidelity} and \ref{nagao-understanding}
respectively \cite{Nagao:1989}.  Others would specify two or three
points on the scale and let the subject's imagination fill the gaps.
Sometimes, examples are provided instead of actual criteria.  Clearly,
there is a tradeoff: fewer criteria can lead to inconsistent
scoring by different subjects and more criteria can lead to general
skewing of scores towards specific readings of these criteria.

\item General or Specific Metrics.  Some metrics focus on certain
aspects of the data: translation accuracy without regard to fluency
or fluency without regard to accuracy.  General metrics would
combine these and other criteria considered important in a single
metric.  Again, the tradeoff here is on the clarity of the results of
the evaluation: General metrics, although realistic about the
interaction between different features of the evaluated data do not
contribute to an understanding of what exactly was lacking in the
data.

\item Independent or Contextual Output. Independent qualitative evaluations
ask subjects to assign a score to a sentence given a metric.
Contextual qualitative evaluations ask for the same task but situate
the sentence in a context of other sentences.  Contextual evaluations
are more realistic in recreating the situation where a sentence is
read; however, they are too demanding of the evaluating subjects.
Another variation on contextual evaluations is to present the subjects
with several outputs from different systems simultaneously.

\item Scale. The scale of a qualitative evaluation varies from 3 (for
example: okay=1, not sure=0.5, wrong=0 \cite{Yamada:2001}) to 100
 \cite{Hovy:2000}.
%some comparison added here .. HCIL prefer 7 ? 

\item Subjects. The number and type of subjects varies too in
different evaluations.  The number of human subjects varies from 2 in
 \cite{Ratnaparkhi:2000} to 80 in \cite{Habash:2003b}.  Some evaluators
constrain the evaluation to native speakers of the target language
only \cite{Habash:2003b}; while others compare results from monolingual
and bilingual speakers \cite{Papineni:2001}.

\end{itemize}

\subsubsection{Automatic Text Quality Evaluation Metrics}
Automatic text quality evaluation metrics allow us to evaluate without
human intervention, objectively and automatically.  A common goal of
such metrics is to show that they correlate to human judgment. There
are two types of these metrics: Sentence Error Rate (or String Error
Rate) and Word Error Rate \cite{Evermann:1999}. Both are measured
relative to some gold standard reference.  Sentence error rate checks
whether the whole generated sentence matches the reference sentence
fully or not.  It is a coarse metric since no partial credit for
partially correct answers is given.  Word error rate is much more
commonly used.  There are many ways to calculate this metric which
tries to capture how different a generated string is from the reference
string.  Two common metrics are string edit distance and n-gram
precision variation.

\paragraph{String Edit Distance Metrics}
There are three automatic evaluation metric classes: string-based
metrics, tree-based metrics and combined metrics
\cite{Bangalore:2000b}.

The string-based metric class includes the following two metrics
suggested in MT literature \cite{Alshawi:1998}: Simple Accuracy and
Generation Accuracy.  Simple Accuracy is calculated by dividing the
number of deletion (D), insertion (I) and substitution (S) errors
(between the target-language strings in the test corpus and the output
of the generation system) by the number of tokens in the target string
(R):

\[ Simple Accuracy = (1 - (I+D+S)/R) \]

Generation Accuracy avoids penalizing misplaced tokens twice as
deletion and insertion errors. Instead these are considered movement
errors (M)

\[ generation Accuracy = (1 - (M+I'+D'+S)/R) \]

For example,
\begin{verbatim}
Gold Standard:     the man likes to eat        chicken .
Generated Output:      man          eats likes chicken .
                    D   .    D   D   S     I     .     .
                    D'  .    M   D'  S           .     .
\end{verbatim}

Simple accuracy for the above example is (1 - 5/7) = 2/7.  Generation
accuracy for the same example is (1 - 4/7) = 3/7.

The tree-based metric class extends string-based metrics by
distinguishing between token moves that violate projectivity (creating
discontinuous constituents) and those which do not.  For example, {\it
man chicken like eat} violates the projectivity of the components of
the dependency tree {\it (like man (eat chicken))} whereas {\it man
like eat chicken} does not.  Tree-based accuracy (simple tree accuracy
or generation tree accuracy) is similar to string-based accuracy
except that each treelet (parent and children nodes ) is checked
separately.  A string of the treelet in the dependency tree is
constructed according to the order it appears in the target string and
another string is constructed according to the order it appears in the
generated string.  The number of insertions, deletions, substitutions
and movements is calculated just like string-based metric.  When done,
the values are summed and used in the same formulas described earlier.

The class of combined metrics addresses the problem that the intrinsic
metrics described earlier cannot replace human evaluation.  However, a
combination of these metrics was shown to correlate significantly
with human judgments of quality and understandability and can be used
for evaluation during development \cite{Bangalore:2000b}.  A large set
of combinations of metrics and values were experimented with to create
a regression that correlates with human judgments qualitative scores.
The result was two new metrics:

\begin{small}
  \begin{itemize}
  \item Understandability Accuracy = \\
    \verb|     |(1.3147*simple tree accuracy - 0.1039*substitutions -0.4458)/0.8689 
    
  \item Quality Accuracy = \\
  \verb|     |(1.0192*simple tree accuracy -0.0869*substitutions -0.3553)/0.6639
  \end{itemize}
\end{small}

Questions about generalizability still stand: generalizability
to other systems, different domains, paragraphs (aggregated
sentences), etc.

%-----------------------------------------------------------------
\paragraph{Bleu: N-gram Precision Variation}

N-gram precision variation compares the generated string and the
reference string by taking the ratio of n-gram sequences in the
generated string that appear in the reference string to the total
number of n-gram sequences in the generated string
\cite{Papineni:2001}.  The most recent automatic evaluation system
using n-gram precision is Bleu from IBM.

Bleu (BiLingual Evaluation Understudy) is a method for MT evaluation
that is quick, inexpensive and language independent and most
importantly correlates highly with human evaluation
\cite{Papineni:2001}.  Bleu's major contribution is the use of
multiple references to score translation candidates (test sentences).  Since there are
many ``perfect'' translations of a single source sentence with
variations in lexical choice and word order, a good translation can
match multiple components in different references.  If a sentence
matched a reference perfectly, it gets a perfect score.  This solution
avoids the problem of the gold standard being too specific to 
certain domains or certain linguistic phenomena.  This feature also
makes Bleu more robust and consistent compared to the techniques
presented earlier.

Bleu uses a modified n-gram precision variation of the word error rate
metric.  The modified n-gram precision, $Precision_n$, equals the
ratio of the sum of n-gram sequences appearing in both candidate and
reference sentences to the total number of n-gram sequences in
candidate sentences.  This is modified so that a reference word is
clipped (exhausted) after a matching has occurred so that multiple
candidate words cannot match a single reference word:

\[Precision_n = { {\sum_{C\in Candidates} \sum_{ngram \in C} Clipped(ngram)}
                  \over {\sum_{C'\in Candidates} \sum_{ngram'\in C'} 1}} \]

The denominator in this equation counts all the n-grams in all
candidate sentences.  The numerator only counts clipped n-grams.
These are n-grams that appear in any reference sentence associated with
the candidate sentence under consideration. Once a match is detected,
it is not allowed to happen again---the n-gram is said to be
clipped.  The function $Clipped(ngram)$ returns 1 if the $ngram$ is
clipped or 0 if it is not.

The unigrams account for adequacy and the longer n-grams account for
fluency.  Bleu uses unigrams, bigrams, trigrams and 4-grams.  The
different n-gram precision scores are combined into a single precision
score using the geometric mean:

\[Bleu Precision = (\prod_{i=1}^{n} x_i )^{1 \over n} \]

To account for recall, a {\it brevity penalty} is calculated over the
whole test set as a decaying exponential in {\it r/c}, where $r$ is
the effective reference corpus length and $c$ is the candidate corpus
length.  The effective reference corpus length is calculated as the
sum of all the reference sentences that are closest in length to the
candidate sentences they are compared to.  The candidate corpus length
is simply the sum of the length of all candidate corpus sentences.

\[Brevity Penalty = \left\{ \matrix{
1 & c > r \cr
e^{(1-r/c)} &  c \le r \cr} \right\}  \]

The final Bleu score is calculated as the multiplication of $Bleu Precision$ with $Brevity Penalty$:

\[Bleu Score = Bleu Precision * Brevity Penalty \]

The following is an example of how the Bleu score is calculated for a
single test sentence given three gold standard references. The Bleu
score in this example is calculated using only unigrams and bigrams.

\begin{sentences}
\item \label{bleutest}TEST: colorless green ideas sleep furiously
\item \label{bleuref}REFERENCES:\\
all dull jade ideas sleep irately \\
drab emerald concepts sleep furiously \\
colorless immature thoughts nap angrily
\end{sentences}

The unigram precision equals the ratio of words in \ref{bleutest}
appearing in any sentence in \ref{bleuref} over all the words in
\ref{bleutest}. In this example, matching unigrams are {\it colorless,
ideas, sleep,} and {\it furiously}. Therefore, unigram precision
equals $4 \over 5$.  The bigram precision score is calculated
similarly by taking the ratio of matched bigrams to all bigrams in
\ref{bleutest}.  Matched bigrams are {\it ideas sleep} and {\it sleep
furiously}.  So, the bigram precision is $2 \over 4$.  The combined
$Bleu Precision$ (using only unigrams and bigrams) equals $\sqrt{{4
\over 5}\cdot{2 \over 4}}$, or 0.632.

To test the correlation between Bleu scores and human scores, a human
evaluation was conducted using the output of five Chinese-English
translation systems. The metric used for human scoring is the Language
Skill Level Descriptions for Writing used by the U.S. Government's
Interagency Language Roundtable \cite{Child:1993}.  The results show
that Bleu tracked human judgment almost perfectly (0.99 correlation)
and had enough sensitivity to distinguish small differences between
systems.

Although Bleu is attractive to use because of its simplicity and high
correlation with human judgment, it has two shortcomings.  First,
Bleu doesn't distinguish between mismatches resulting from
inflectional variations or synonyms and those resulting from wrong
choice of words. For example, {\it sleep} and {\it slept} or {\it
green} and {\it jade} share more meaning than {\it paper} and {\it
role} (both are possible translations of Spanish {\it papel}).  Yet,
all these mismatches are penalized equally.  Secondly, Bleu
prefers the availability of multiple references (although not
required).  Creating test sets with multiple reference is a time- and
cost-consuming effort.  There are few such sets for Chinese-English
\cite{multichin} and Arabic-English \cite{multiarab}.  These two
issues are inter-related since the availability of more multiple
references allows for good possible variations to be matched.
In the evaluation chapter, Chapter~\ref{eval-chapter}, some solutions
are presented to address the first issue above given that the
conducted evaluation uses only one reference per translation.

\paragraph{}
This chapter discussed research and concepts in NLG, lexical semantics
and evaluation, all of which are relevant to the Generation-Heavy
Hybrid MT (GHMT) approach presented in subsequent chapters.  The GHMT
model expands on the concept of constrained overgeneration introduced
in the hybrid NLG system Nitrogen \cite{Langkilde:1998b} by working
from syntactic dependencies instead of semantic representations and by
injecting TL-specific lexical semantic knowledge.  This knowledge
includes thematic role hierarchies \cite{Levin:1996}, limited lexical
functions \cite{Melcuk:1988}, verb classes
\cite{Levin:1993b,Dorr:2001a} and Lexical Conceptual Structure (LCS)
representations \cite{Dorr:2001a,Dorr:2001b}.  A major component of
this research is an extensive evaluation of translation quality that
compares several models with varying levels of complexity using the
Bleu evaluation metric \cite{Papineni:2001}.

%--------------------------------------------------------------------------
%--------------------------------------------------------------------------
%--------------------------------------------------------------------------
%--------------------------------------------------------------------------
%--------------------------------------------------------------------------
%--------------------------------------------------------------------------
\chapter{Generation-Heavy Hybrid Machine Translation}\label{ghmt-chapter}

The different approaches to Machine Translation (MT) whether symbolic,
statistical or hybrids to varying degrees share an assumption of
symmetry between analysis and generation in the algorithms, the
resources or both.  Symbolic approaches utilizing transfer or
interlingua require symmetrical amounts of resources and algorithms.
Statistical approaches demand large parallel corpora to achieve any
good results. The research presented here does not question the
theoretical validity of symmetrical approaches, but proposes an
alternative approach for translation between language pairs with
different densities of resources. Here, the intent is to translate from
a low density language (with minimally a parser) to a high density
language (with ample resources of all kinds) using a simple
translation lexicon that does not address complex structural variations
between the two languages.

This chapter presents an approach to MT that addresses the lack of
symmetry by exploiting symbolic and statistical target-language (TL)
resources in source-poor/target-rich language pairs. This approach is
called Generation-Heavy Hybrid Machine Translation (GHMT). Expected
source-language (SL) resources include a syntactic parser and a simple
one-to-many translation dictionary. No transfer rules or complex
interlingual representations are used. Rich TL symbolic resources such
as word lexical semantics, categorial variations and subcategorization
frames are used to overgenerate multiple structural variations from a
TL-glossed syntactic dependency representation of SL sentences. This
symbolic overgeneration, which accounts for possible translation
divergences, is constrained by multiple statistical TL models
including surface n-grams and structural n-grams.  The source-target
asymmetry of systems developed in this approach makes them more easily
retargetable (re-{\it source}-able) to new SLs (provided a parser and
translation lexicon).

%The basic intuition of the approach parallels the experience of most
%language learners whose lack of symmetrical knowledge impairs their
%ability to translate into their newly learned language but does not
%hinder them as much when translating from the foreign language into
%their native tongue (where they are assisted by rich resources).

The next section motivates this {\it hybrid} approach to translation
divergences and relates it to previous research. Then, an overview of
GHMT is followed by detailed discussion of its different components.
 
\section{A Hybrid View of Translation Divergences}

LCS-based MT provides a framework for handling both monolingual
alternations and translation divergences using the LCS representation
as an Interlingua and constraining generation options by setting
different parameters for each language.  A hybrid view of divergences
would consider them {\it uncommon alternations}: variations that are
symbolically compositionally determinable but whose probability of
occurrence is idiosyncratic and thus can only be statistically
modeled.  For example, the Spanish {\it tener hambre (have hunger)}
and the English {\it to be hungry} are translation divergences even
though {\it to have hunger} is correct English. The difference between
{\it to be hungry} and {\it to have hunger} is that the former is more
common than the latter. While lexical semantic knowledge can help link
the two as possible variants, only a statistical model can determine
which to pick as the most likely translation.  Different statistical
models within the same language and other genre-specific constraints
can participate in selecting the best choice in translation.  For
example, a model of English poetry with restrictions on rhyme would be
expected to produce different choices from a Wall Street Journal
model.

The approach presented here is closely related to work on Hybrid
Natural Language Generation
 \cite{Knight:1995,Langkilde:1998a,Langkilde:2000} and it draws heavily
on resources from research in Lexical Conceptual Structure (LCS) based
MT \cite{Dorr:1993,Dorr:2001a,Dorr:2001b,Habash:2003b}.
The idea is to combine symbolic and statistical knowledge in
generation through a two step process: (1) Symbolic Overgeneration
followed by (2) Statistical Extraction (or Ranking).  The hybrid approach has been
used primarily for lexical choice (including morphology and tense
selection) \cite{Langkilde:1998c,Bangalore:2000c} and for
linearization from semantic representation \cite{Langkilde:1998a} or
from shallow unlabeled dependencies \cite{Bangalore:2000a}.

GHMT extends the hybrid approach to handle translation divergences
without the use of a deeper interlingual representation or transfer
rules.  This is accomplished by extending the symbolic overgeneration
component to include structural and categorial expansion of the SL
lexico-structural representation using lexical semantic knowledge from
the TL (derived from LCS databases).  The overgeneration is
constrained by linguistically motivated rules that utilize TL lexical
semantics and subcategorization frames and are independent of SL
preferences.  The overgenerated lexico-structural combinations are
then ranked by the statistical ranking component.\footnote{Another non-interlingual
non-transfer approach is Shake-and-Bake MT \cite{Beaven:1992b} which
overgenerates TL sentences and symbolically constraints the output by
"parsing it."  It differs from GHMT in two ways: (1) Shake-and-Bake is
a purely symbolic approach and (2) it requires symmetric resources for
SL and TL.}

\section{Overview of GHMT}
\begin{figure*}[htb]
\centerline{\includegraphics[]{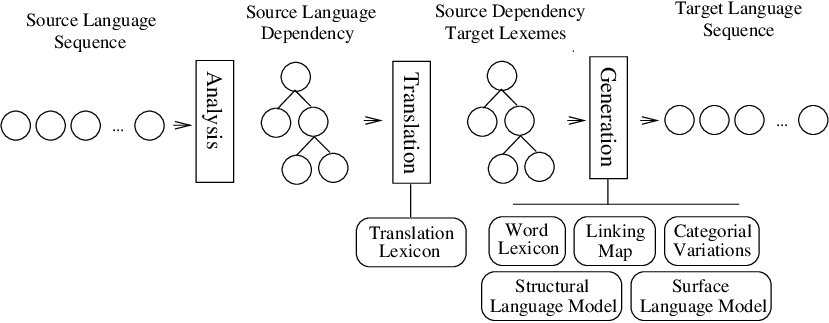}}
\caption{Generation-Heavy Machine Translation}
\label{ghmt}
\end{figure*}

Figure~\ref{ghmt} presents an overview of the complete MT system.
The three phases of Analysis, Translation and Generation are very
similar to other paradigms of MT: Analysis-Transfer-Generation or
Analysis-Interlingua-Generation \cite{Dorr:1999}.  However, these
phases are not symmetric.  Analysis relies only on the SL sentence
parsing and is independent of the TL. The output of Analysis is a deep
syntactic dependency that normalizes over syntactic phenomena such as
passivization and morphological expressions of tense, number, etc.
Translation converts the SL lexemes into sets of TL lexemes.  The
dependency structure of the SL is maintained.  The last phase,
Generation, is where most of the work is done to manipulate the input
lexically and structurally and produce TL sequences.

\subsection{Symbolic Components}
GHMT utilizes three major TL resources.  First, the
word-class lexicon defines verbs and prepositions in terms of their
subcategorization frames and lexical conceptual primitives.  A single
verb or preposition can have multiple entries for each of its senses.
For example, among other entries, $run_1$, as in ($John_{agent}$
$ran_{cause-go_{identificational}}$ $store_{theme}$), is distinguished
from $run_2$, as in ($John_{theme}$ $ran_{go_{locational}}$).  Second,
the categorial-variation lexicon relates words to their categorial
variants.  For example, $hunger_V$, $hunger_N$ and $hungry_{AJ}$ are
clustered together. So are $cross_V$ and $across_P$; and $stab_V$ and
$stab_N$. Finally, the syntactic-thematic linking map relates
syntactic relations (such as subject and object) and prepositions to
the thematic roles they can assign.  For example, while a subject can
take on just about any thematic role, an indirect object is typically
a {\it goal, source} or {\it benefactor}.  Prepositions can be more
specific. For example, {\it toward} typically marks a {\it location}
or a {\it goal}, but never a {\it source}.

\subsection{Statistical Components}

In GHMT, a surface n-gram model is used together with a {\it
structural n-gram model}, which is a new type of language model not
accounted for in the three-way distinction presented earlier
 \cite{Daume:2002} (Section~\ref{struct-models}).  A structural n-gram
language model characterizes the relationship between words in a dependency
representation of a sentence without taking into account the overall
structure at the phrase level.  Such a model is not useful for parsing
purposes since the goal of parsing is to determine an overall
structure.  However, in generation, where the structure is already
determined (albeit ambiguously), this model is very useful for making
lexical selection choices even though it is blind to the structural
relation (e.g. subject versus object). In addition, it is less complex
and less sparse than a lexicalized syntax language model.

The following sections describe the different steps and their purpose.
Implementational details will be described in
chapter~\ref{matador-chapter}.

\section{Analysis}\label{system-analysis}

The analysis component of a GHMT system is a SL
deep syntactic dependency parser.  Syntactic dependencies transcend
SL syntactic idiosyncrasies such as SL
surface word order without delving into semantic analysis and its
challenges.  Dependencies are, therefore, deeper representations (and
thus more language-independent) than constituencies yet they are
shallower than semantic representations.  The variety of syntactic
dependency used normalizes over syntactic phenomena such as
passivization, morphological variations including de-agglutination
(Turkish) or segmentation (Chinese), and case/order-driven resolution
of syntactic relations (subject-hood, object-hood,etc.).  These are
purely SL issues that should be dealt with before the
translation step.  Morphological analysis is also a simplifying factor
for the level of complexity of the translation lexicon since
inflectional variants of a single word are numerous in some languages
(e.g. Arabic or Turkish).

Verbs in the SL that violate the Universal Thematic
Hierarchy (such as Psych verbs) need to be thematically normalized (or
in other words, de-externalized: the subject and object are
reversed). Analysis is the only logical place for handling thematic
divergence since it is dependent on idiosyncratic SL
features.  Normalization allows such verbs to be translated correctly
into TL verbs that follow or violate the Universal Thematic
Hierarchy.

\subsection{Why Parse at All?}
In statistical MT models that view the SL sentence as a sequences of
words with no constituent dependence, relations among these words are
limited to a contextual window that varies from a few words to the
whole sentence.  With the smaller contextual window, the search space
is too restricted, which excludes valid outputs; and with the larger
full-sentence context, the space is so large that it becomes hard to
search \cite{Knight:1999}.
% and is likely to contain more `good target
%outputs' but not for the SL input.

Efficiency and correctness of operations such as lexical selection
and linearization are affected by the size of the space of
possibilities.  For example, statistical techniques ignore the fact
that linguistically dependent structures move together in translation:
the adjectives modifying a noun are semantically dependent on the noun
and that fact is reflected in the dependency relationship between the
adjectives and their head noun.  A brute-force statistical approach to
linearization overgenerates all possible permutations of the TL
sentence and orders them using n-gram models which can only capture
local context likelihood.  

With dependency parsing, the number of permutations is exponentially
smaller because it is restricted to a parent and its children (but not
the grand-children).  Moreover, the process of lexical selection can
be dependent on long-distance dependencies which are hard to capture
in a flat sentence structure.  A syntactic parse is necessary for
reducing the space of possibilities without compromising correctness
(compare with taking all possible permutations into consideration).

Parsing ambiguity is a problem of course, but it does not compare to
the complexity of ambiguity in flat sentences.  The latter is actually
an upper bound on ambiguity in syntactic parsing: the worst parse
possible has all the words in a sentences as sister nodes.  In other words,
a parse (shallow or deep) always does better compared to no
parsing.  The parsing process helps reduce the space of possibilities
through defining smaller structural dependencies and also by resolving
categorial ambiguities such as part of speech.

Although parsing is a relatively source-heavy module, new developments
in the field of statistical parsing allow the creation of parsers
using a parallel corpus of a poor language and a rich language (see
section~\ref{realpoor} for more details).

%-----------------------------------------------------------------

\section{Translation}\label{system-translation}

In this step, the SL syntactic dependency tokens are translated to TL
tokens.  Since this step does not include any modification of the
dependency structure, it can be thought of a simply ``lexical
transfer''.  This lexical transfer is done using a translation lexicon
(Tralex) or bilingual dictionary. This is the only language-pair
dependent component of the system.  The following is the format of the
Tralex:
\begin{sentences}
\item {\small
(word$_SL$ pos$_{SL}$) $\Rightarrow$ ((word$_{TL}^1$ pos$_{TL}^1$) (word$_{TL}^2$ pos$_{TL}^2$) .. (word$_{TL}^n$ pos$_{TL}^n$))}
\end{sentences}
where n represents the number of pairs of TL words and their part of
speech mapped to a SL word and its part of speech.  For example,
\begin{sentences}
\item \begin{small}
\verb|Spanish-English: (entrar v)  ((enter v) (engage v))|\\
\verb|Spanish-English: (gustar v)  ((like v) (please v))|\\
\verb|Spanish-English: (pedir v) ((ask_for v))|\\
\verb|German-English:  (gern adv)  ((like v))|\\
\verb|Arabic-English:  (|{\Y \C216\C215\C167\C80}\verb| v) ((wear_turban v))|\\
\end{small}
\end{sentences}

The mapping can be across part of speech categories (e.g. $gern_{adv}$
and $like_{v}$). Consistency between the Tralex and the tokens in the
syntactic dependency input is important.  For example, nouns in
languages with heavy case marking or concatenative postpositions can
be decomposed into separate morphemes in their syntactic dependency;
however each morpheme needs to be included in the Tralex.  In
principle, it is not important what form of the SL word is used so
long it is consistent with the output of the parser.  The form of the
TL word, however, must be an un-inflated form, or lexeme, since later
steps will need to use lexical semantic information accessible by the
word's lexeme.

The Tralex can also include {\it inflated} translations
(i.e. multi-word tokens such as {\it wear-turban}) on the TL side.
%Multi-word tokens will be expanded into structural interpretations of
%the relationship between the words.  
Multi-word tokens can be used on the SL side if they are valid tokens
in the SL parser output.  Another feature of the output of this step
is that the nodes of the dependency are now ambiguous sets of words.
This is very important for efficient processing since multiplying out
different input trees will lead to an exponential explosion of
possibilities.

%-----------------------------------------------------------
\section{Generation}

\begin{figure*}[htb]
\centerline{\includegraphics[width=4.5in]{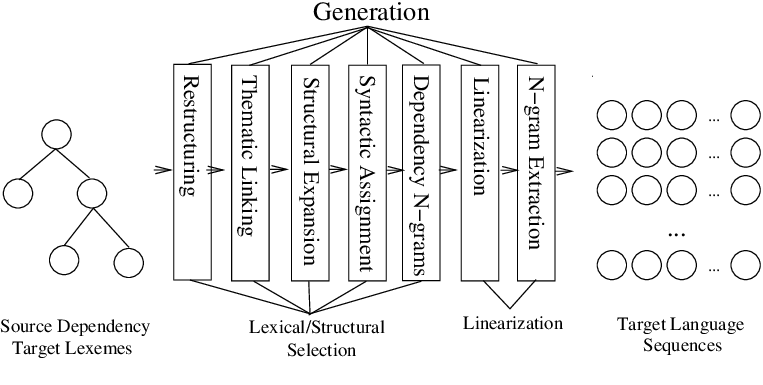}}
\caption{Generation}
\label{newsys}
\end{figure*}

The generation component
% [[ Exerge ]] 
contains seven steps (Figure~\ref{newsys}).  The first five are
responsible for lexical and structural selection and the last two are
responsible for linearization.  Initially, the SL syntactic
dependency, now with TL lexemes, is normalized and restructured into a
syntactico-thematic dependency format.  The thematic roles are then
decided in the thematic linking step.  This is followed by structural
expansion which explores conflated and inflated variations of the
thematic dependency.  The fourth step maps the thematic dependency to
a TL syntactic dependency. The first four steps are all symbolically
driven. The fifth step is the first statistical component where
dependency bigram statistics are used to reduce the ambiguity in the
overgenerated variations.  After the linearization generates a lattice
of word sequences, these are ranked using a statistical surface n-gram
model.

\subsection{Restructuring}

The goal of this step is to handle the cases when a SL word is
translated into a word with a different POS that violates the
wellformedness of the input.  For example, if a SL verb
dominating another verb is translated into an adverb or auxiliary,
then the dependency will be restructured such that the dominated verb
is promoted and the auxiliary or adverb is demoted.  In
Spanish-English translation, this step is where verbs like {\it deber
(should/owe)}, {\it poder (can/be able)} and {\it soler
(usually/tend)} are handled.  
%In other words, this is where simple
%head-swapping is handled.

%----------------------------------------------------------------
\subsection{Thematic Linking}\label{thematic-linking}
The generation algorithm uses thematic dependencies as an internal
representation.  A thematic dependency is shallower than a typical
interlingua but deeper and more language-independent than a
syntactic-dependency.  It is important to note that the SL analysis
comes before the mapping into the thematic representation. The
thematic linking step is done in the generation process using the TL
resources only.  Therefore, it is a {\it loose} linking algorithm that
is constrained by the thematic grids of the predicates in the TL
(verbs and prepositions).  The output of this phase is a thematic
dependency in which the edges linking children to parents are marked
with thematic roles (and modifiers) together with syntactic relations.
 
The goals of this phase are many:

\begin{itemize}
\item Reduction of ambiguity: since each verb can have multiple verb
class memberships (some of which have multiple thematic grids), this
step reduces the verb/verb-class/grid possibilities.
%to only those that
%rank highest according to the criteria described earlier.
\item Normalization: this step normalizes over structural variation
and thus approaches a solution to the structural and thematic
divergences on a thematic level.\footnote{This also applies to
expanding the possible set of alternations eventually.} 
\item Accurate thematic assignment: this is essential for handling
structural variations.
\end{itemize}

The basic idea here is to use prepositions as syntactic relations, or
case markers, that constrain the option of thematic roles that can be
assigned to the prepositions' objects. Some limitation of
ambiguity is possible here using the number and nature (obligatory,
optional) of the thematic roles determined by the verb thematic grid
together with additional constraints which will be discussed in
further detail in section~\ref{thetalink-algo}. Prepositions used as
syntactic relations are kept in the representation because some
contain semantic information, such as direction of motion, that is
used in the next step of structural expansion.

This step looks like analysis but it is fully driven by resources and
constraints from the TL.  That is why it is a central step in this
generation-heavy approach.

\subsection{Structural Expansion}

This step overgenerates alternative structural configurations of the
input's thematic dependency.  There are three concepts relevant to
this step.

\begin{enumerate}
\item \textbf{Basic Operations}\\ There are two basic types of
operations applied here: Conflation and Inflation. Each type has two
forms: Node-Node and Node-Relation. {\it Nodes} refer to words in a
thematic dependency and {\it Relations} refer to the prepositions
labeling the edges that connect these words.

\begin{itemize}
\item Node-Node Conflation occurs when the semantic content of two nodes conflates into one
node as in {\it give stab} becoming {\it stab}.
\item Node-Relation Conflation occurs when the semantic content of a
node and the label marking one of its relations to a child node conflate as in
{\it go across X} becoming {\it cross X}.
\item Node-Node Inflation occurs when the semantic content of a node
is split over two nodes as in {\it stab} becoming {\it give stab}.
\item Node-Relation Inflation occurs when the semantic content of a
node is split between itself and the label marking one of its
relations to a child node as in {\it cross X} becoming {\it go across X}.
\end{itemize}

Whereas Node-Node operations change the structure of a thematic
dependency, Node-Relation operations only modify the content of its
nodes and edges.  However, this change in content implies a structural
change at the syntactic level.  This change is not realized until the
next step of Syntactic Assignment.

Lexical-semantic information from the verb class lexicon, such as
theta grids and lexical conceptual primitives, is used to determine
the ``conflatability'' and ``inflatability'' of combinations of nodes
and relations in the thematic dependencies.

\item \textbf{Semantic Granularity}\\
This is the level of semantic refineness of the
underlying lexical semantics model.  For example, when using LCS as
the underlying model, conflations such as ({\it put+butter $\rightarrow$  butter})
are possible but not ({\it drink+fast $\rightarrow$ guzzle}). The former is
de/composable in LCS but the latter is not.  In principle, the GHMT
approach is independent of the lexical semantic representation. Other
implementations of GHMT using different lexical semantic models will
have different degrees of granularity and thus expressibility and
structural expansion-ability.  Idiomatic expressions are, by definition,
non-composable and thus cannot be addressed by a lexical semantics
approach of any granularity.

\item \textbf{Degree of Expansion}\\ The basic operations can be
applied compositionally to create more complex structural variations.
The {\it degree of expansion} is measured by the number of operations
applied.  It is restricted by the system's semantic granularity which
determines the maximum number of basic operations that can take place.
When using LCS, only first and second degree expansion is possible.
The following is an example of a second degree expansion:

\begin{enumerate}
\item Node-Relation Inflation: {\it (cross river swimming)} $\rightarrow$ {\it (go (across river) swimming)}
\item Node-Node Conflation: {\it (go (across river) swimming)} $\rightarrow$ {\it (swim (across river))}
\end{enumerate}

\end{enumerate}

%----------------------------------------------------------------
\subsection{Syntactic Assignment}
In this step, the thematic dependency is turned into a full TL
syntactic dependency.  This includes (1) assigning TL syntactic
positions to thematic roles, (2) enforcing various feature
restrictions of the TL head verb such as argument category, and
(3) generating different alternations associated with a single class
to allow for a widening range of expression that is specific to the
TL.  This step is needed to complete the handling of the
structural divergence and the categorial divergence.
%----------------------------------------------------------------
\subsection{Statistical Dependency Pruning}

The input to this step is an ambiguous syntactic dependency and the
output is non-ambiguous.  This step prunes ambiguous nodes using
dependency bigram statistics.  The purpose of this step is to
constrain the overgeneration of the previous steps using a language
model that is based on structural relations between lexemes. This is
different from the last step (Statistical Ranking) in three ways: (1)
it is structural not word-order-based, (2) it is based on lexemes not
final surface forms, and (3) its effect is only seen on lexical
selection whereas the n-gram statistical ranking determines both
lexical selection and linearization.

%----------------------------------------------------------------
\subsection{Linearization}
In this step a rule-based linearization grammar is used to create a
word lattice that encodes the different possible realizations of the
sentence. This step includes morphology level overgeneration as well
as word-order overgeneration.  The following example shows a very
simple lattice: 

\begin{sentences}
\item
\begin{small}\label{exlin1}\begin{verbatim}
(SEQ
 (OR (SEQ (WRD "the" 1) (WRD "bread" 1) (WRD "should" 1) 
          (WRD "be" 1) (OR (WRD "buttered" 1) (WRD "butterred" 1)))
     (SEQ (OR (WRD " " 1) (WRD "the" 1) (WRD "a" 1) (WRD "an" 1))
          (WRD "bilberry" 1) (WRD "should" 1) (WRD "be" 1)
          (WRD "breaded" 1))))
\end{verbatim}\end{small}
\end{sentences}

This lattice defines six possible sequences: two for the first
sentence with different verb conjugations and four for the second
sentence with different possible determiners.

%----------------------------------------------------------------
\subsection{Statistical Ranking}
The final step, extracting a preferred sentence from the word lattice
of possibilities is done using a surface n-gram language model.
The following is the output of statistical ranking of the lattice
shown in \ref{exlin1} using a bigram language model of English.  Th
model used is Halogen's default model built using a large collection
of English text including Brown corpus and Wall Street Journal.
 
\begin{sentences}
\item
\begin{small}\label{exlin2}\begin{verbatim}
the bread should be buttered    -269.286
the bread should be butterred   -272.747
bilberry should be breaded      -276.609
the bilberry should be breaded  -278.796
a bilberry should be breaded    -281.425
an bilberry should be breaded   -284.118
\end{verbatim}\end{small}
\end{sentences}
%----------------------------------------------------------------

\section{An Operational View of Divergences}

The constrained overgeneration approach, as applied in GHMT for
handling translation divergences, presents a different view from the
previous lexicon-based parameter-driven divergence classification
first defined by Dorr \cite{Dorr:1993}.  The operational view
implemented in this dissertation reduces translation divergence
types/classes to four simple compositional operations:
\begin{itemize}
\item Categorial Variation: This is a change of the POS of a word. The
resource associated with this operation is the categorial variation
database. The modules associated with this operation are structural
expansion and syntactic assignment.  In terms of graph (dependency
tree) modification, it only applies to node labels.  This operation is
the most common of all operations since it is applied in conjunction
with other operations most of the time.  For example, a conflation
always includes a categorial variation.

\item Case Variation: This is a change to the relation between two
words.  The resources associated with it are the thematic-syntactic
linking map and the word-class lexicon.  The modules associated with
case variation are input normalization, restructuring, thematic
linking, node-relation conflation/inflation and syntactic
assignment. The case variation operation combines the thematic
divergence and the structural divergence since both can be seen as a
change in assignment of a syntactic case to a thematic role.  In terms
of graph modification, it only applies to edge labels.

\item Conflation: This is a change where two words are combined into
one.  The resource associated with this operation is the
word-class lexicon.  The module associated with this operation is
structural expansion.  Conflation always occurs with categorial
variation. In terms of graph modification, conflation is the joining
of two nodes into one.

\item Inflation: This is a change where one word is split into two.
The resource associated with this operation is the word-class lexicon.
The module associated with this operation is structural expansion. In
terms of graph modification, inflation is the splitting of one node
into two.
\end{itemize}

These divergence operations are applied after word translation and are
fully driven by target-language resources.  A lexical-semantic
framework is necessary to define how some of these operations are
conditioned.  In the implementation presented in this dissertation,
the lexical-semantic frame work is LCS.  But this is not an inherent
restriction on these operations, only an implementational choice.  The
granularity of the LCS representation sets a limit on the number of
operations that can be applied.  For example, no more than two
inflation operations can be applied on a verb since the LCS
representation doesn't go beyond two levels of decomposition: {\it X
butter Y} can be represented in LCS as [CAUSE X [GO BUTTER [TO [AT
Y]]]] which maximally allows inflations such as {\it X made the butter
go on Y}.

\spacing{1}
\begin{table}
\begin{center}
%\begin{tabular}{|c|c|c|c|}
\begin{tabular}{|p{1.1in}|p{2.4in}|p{1.2in}|}
%\begin{tabular}{|c{1.0in}|c{1.0in}|c{1.0in}|c{1.0in}|}
\hline 
\textbf{Divergence Type} & \textbf{LCS-based MT} & \textbf{Divergence Operations}\\
\hline
Categorial & SL and TL lexical entries with shared primitives & categorial variation\\
\hline
Conflational & SL and TL lexical entries shared substructures & conflation\\
\hline
Inflational & SL and TL lexical entries shared substructures & Inflation\\
\hline
Structural & SL and TL lexical entries with argument-position  markings &  case variation \\
\hline
Head-swapping & TL lexical entries marked for predicate reversal w.r.t  SL predicates  & inflation followed by conflation\\
\hline
Thematic & TL lexical entries marked for argument reversal w.r.t. SL arguments & case variation\\
\hline
\end{tabular}
\end{center}
\caption{LCS-based MT vs Divergence Operations}
\label{lcsmt-divops}
\end{table}
\spacing{1.667} 
The different divergence types presented in Section~\ref{divergences}
can be redefined in terms of divergence operations.  While the list of
divergences in Section~\ref{divergences} does not include inflation as
a separate divergence, it may be implied as a reverse of conflation.
Table~\ref{lcsmt-divops} illustrates how different divergence types
are handled in LCS-based MT and compares them to necessary divergence
operations.

In addition to explaining different divergences in a simple manner,
the divergence operations predict divergences not previously
classified.  These divergences result from applying multiple
operations in new ways.  For example, the following multiple
head-swap case is predicted even within the LCS framework:
\begin{sentences}
\item John spent December across Europe\\   
John {\it went in December} across Europe (Inflation, Categorial, Case)\\ 
John {\it crossed Europe} in December     (Conflation, Categorial, Case)
\end{sentences}
Using a different lexical-semantic framework with finer granularity
such as the large-scale CYC knowledge base
\cite{Lenat:1990,Lenat:1991} is likely to predict even more complex
cases.

The divergence operations are reminiscent of the noisy channel
operations applied in Yamada's syntax-based statistical MT (SSMT)
\cite{Yamada:2001}: child-reordering, node-insertion and
node-translation (see discussion in Section~\ref{sbst}).  Divergence
operations are different from Yamada's operations in many ways: (1)
they apply to dependencies not constituents; (2) they operate on a
deeper level that is independent of surface order (e.g. no
child-reordering); and (3) they are generation-driven not
translation-driven.  The fact that the GHMT approach and the SSMT
approach are on opposite end of the hybridity continuum (GHMT more
symbolic than statistical and SSMT more statistical than symbolic) is
not an inherent requirement of the implementation of the four
divergence operations.  It is possible to define these operations over
a noisy channel model that transforms TL dependencies into SL
dependencies.  An annotated parallel corpus would be required for
training in this approach, although the resulting model would fall
outside of the generation-heavy spirit.

\paragraph{}
This chapter described the Generation-Heavy Hybrid Machine Translation
approach in a language independent manner.  The next chapter focuses
on implementation issues of an instantiation of GHMT for
Spanish-English translation, Matador.

%-------------------------------------------------------------------------
%-------------------------------------------------------------------------
%-------------------------------------------------------------------------
%[[GHMT, Exerge embodies the GHMT approach, Matador and Nitrogen/Halogen, oxygen?]]
%-------------------------------------------------------------------------
%-------------------------------------------------------------------------
\chapter{Matador: A Spanish-English GHMT System} \label{matador-chapter}

\begin{figure}[h]
\begin{center}
\begin{tabular}{p{4.0in}}
\centerline{\includegraphics[width=4.0in]{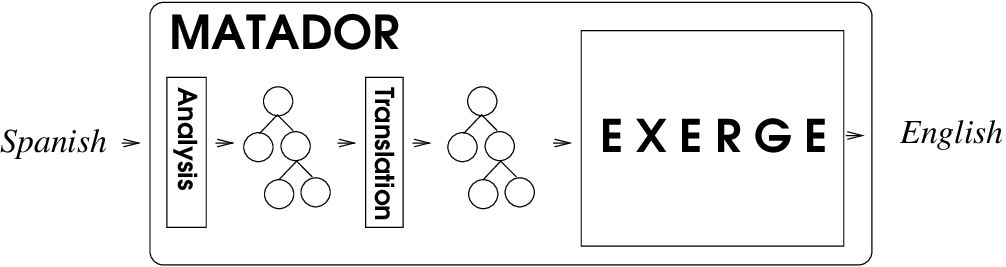}}
\end{tabular}
\caption{Matador and Exerge}
\label{matador-exerge}
\end{center}
\end{figure}

This chapter presents the implementational details of the analysis and
translation components of Matador, a Spanish-English GHMT system.  The
implementational details of the English-heavy generation component of
Matador, named Exerge---Expansive Rich Generation for English, are
presented in Chapter~\ref{exerge-chapter}.  Exerge is a GHMT reusable
component for translating from other languages into
English. Figure~\ref{matador-exerge} illustrates the relationship
between Matador and Exerge within the GHMT framework.

Matador uses some off-the-shelf components, namely the Connexor Spanish
parser for Analysis and the Halogen Forest Ranker for surface N-gram
ranking (in Exerge).  All other components were created as part of this
research.
%[[oxygen?]]
The primary programming language is Lisp for Exerge and Perl for the
translation and parser post-processing modules.

This chapter contains four sections.  First, the overall document
structure and formats used for input and output between all the
modules in Matador are described in
section~\ref{matador-overall-section}.  Section~\ref{matador-analysis}
describes the Spanish analysis component.
Section~\ref{matador-translation} describes the Spanish-English
word-to-word translation process and the creation of the
Spanish-English translation dictionary. Finally,
Section~\ref{realpoor} addresses the issues involved in creating
analysis and translation resources for very poor language.

%----------------------------------------------------------------------
\section{Matador Input/Output Specification}\label{matador-overall-section}

Matador uses a file exchange format that was designed to maintain
modular independence among its different reusable components (such as
surface sentences, dependency trees and word lattices).  The file
exchange format is also compatible with the style representation of
the Bleu evaluation system, which encodes meta information such as
document ID and system ID.  The exchange format is extensible to
XML features encoding other document structure information such
as \verb|<title>| or \verb|<footnote>|.  Since Matador is an
end-to-end MT system, the input and output are Spanish sentences and
English sentences, respectively.  The Syntactic Dependency (SynDep) is
the main exchange format among the analysis, translation and
generation modules (and within the expansion and linearization submodules
of Exerge).

\paragraph{Syntactic Dependency Specification}\label{input-spec}

The syntax of SynDep is a variation on the Penman Sentence Planning
Language (SPL)\footnote{In Matador, two implementations of this
representation are used depending on the programming language: a
Lisp-friendly implementation and a Perl-friendly implementation.  In
this chapter, only the Lisp-friendly form is used since it is more
human-readable.}:

\begin{sentences}
\item \begin{small}
\begin{verbatim}
<SynDep>   ::= <NODE> || <OR>
<NODE>     ::= (<LABEL> / <INSTANCE> 
                       :pos <PART-OF-SPEECH> 
                       {<RELATION> <SynDep>}* 
                       {<FEATURE> <VALUE>}*)
<OR>       ::= (<LABEL> {:OR <NODE>}+)
<INSTANCE> ::= <STRING> || (*or* <STRING>+)
<PART-OF-SPEECH> ::= N || PN || V || AJ || AV || P || 
                     D || PRO || PX || C || NUM
<RELATION> ::= :subj || :obj || :obj2 || :comp || 
               :pred || :mod || :cc || :main
\end{verbatim}
\end{small}
\end{sentences}	

In any SynDep, the topmost node is a special \verb|*root*| node whose
label is \verb|0|.  For all other nodes, \verb|<LABEL>| is a number
specifying the {\it adjusted source language surface order}. This
order specifies the order of words after (1) adding non-realizable
words such as \verb|*pro*| or \verb|*trace*| (2) deleting some words
such as verbal auxiliaries and (3) reattaching detached clitics such
as Spanish {\it se}.

SynDep nodes are of two types: simple nodes or OR nodes. Simple nodes
indicate a specific instance (lexeme).  While OR nodes indicate an
ambiguity (or multiple instantiations) in the tree.

A small number of parts of speech is allowed (compared to sets used by
the Penn Treebank \cite{Marcus:1994} and others): nouns (\verb|N|),
proper nouns (\verb|PN|), verbs (\verb|V|), adjectives (\verb|AJ|),
adverbs (\verb|AV|), prepositions (\verb|P|), articles (\verb|D|),
pronouns (\verb|PRO|), punctuation (\verb|PX|),
coordinating/subordinating conjunctions (\verb|C|) and numerals
(\verb|NUM|).

The list of relations includes: subject (\verb|:subj|), object
(\verb|:obj|), indirect object (\verb|:obj2|), predicate
(\verb|:pred|), complement \verb|:comp|, coordinating conjunction
(\verb|:cc|) and general modifier (\verb|:mod|).  The relation
\verb|:main| is reserved for verbal nodes attached to \verb|*root*|.

Coordinating conjunctions between two nodes are dependent on the
successor node, which is treated as a feature of the preceding node
(\verb|:mod-cc|).  The coordinating conjunction's relationship is
\verb|:cc| and its POS is \verb|C|.

\spacing{1}
\begin{table*}
\caption{Syntactic Dependency Specification}
\label{Syndep-feats}
\begin{center}
\begin{small}
\begin{tabular}{|p{0.8in}|p{1.7in}|p{1.6in}|p{0.6in}|}
\hline
\textbf{Feature} & \textbf{Definition} &  \textbf{Values} & \textbf{Parts of Speech} \\ \hline
%---------------------------------------------------------------
\verb|:WORD| & Source language word & \verb|<STRING>| & \verb|*| \\ \hline
\verb|:SURFACE| & Target language phrase translation & \verb|<STRING>| & \verb|*| \\ \hline
\verb|:MOD-CC| & Coordinated Modifier & \verb!<NODE>!  & \verb|*| \\ \hline
%---------------------------------------------------------------
\verb|:GEN| & Gender & \verb!(F|M|N)! & \verb|N|, \verb|V|, \verb|PN| \\ \hline
\verb|:NUM| & Number & \verb!(SG|PL)! & \verb|N|, \verb|V| \\ \hline
\verb|:PER| & Person & \verb!(1|2|3)! & \verb|N|, \verb|V| \\ \hline
\verb|:CASE| & Case & \verb!(NOM|ACC|! \verb!GEN)! & \verb|N|, \verb|PN| \\ \hline
\verb|:MOOD| & Mood & \verb!(IND|SUB|! \verb!IMP|CND)! & \verb|V| \\ \hline
\verb|:TENSE| & Tense & \verb!(PRES|PAST|FUT|! \verb!IMPF!$^*$ \verb!|PRET!$^*$\verb|)| & \verb|V| \\ \hline
\verb|:CFORM| & Verb Form & \verb!(INF|FIN)! & \verb|V| \\ \hline
\verb|:ASPECT| & Verb Aspect & \verb!(PROG|PERF)! & \verb|V| \\ \hline
\verb|:SFORM|$^*$ & Source Verb Form & \verb!(INF|PCP! \verb!|GER)! & \verb|V| \\ \hline
\verb|:ABBR| &  Abbreviation & \verb|+| & \verb|N|, \verb|PN| \\ \hline
\verb|:COMPAR| & Comparative & \verb|+| & \verb|AJ|, \verb|AV| \\ \hline
\verb|:SUPER| & Superlative & \verb|+| & \verb|AJ|, \verb|AV| \\ \hline
\verb|:INTERR| & Interrogative & \verb|+| & \verb|PRO|, \verb|D| \\ \hline
\verb|:PERS| & Personal  & \verb|+| & \verb|PRO|, \verb|D| \\ \hline
\verb|:POSS| & Possessive  & \verb|+| & \verb|PRO|, \verb|D| \\ \hline
\verb|:DEM| & Demonstrative  & \verb|+| & \verb|PRO|, \verb|D| \\ \hline
\verb|:REFL| & Reflexive  & \verb|+| & \verb|PRO|, \verb|D| \\ \hline
\verb|:REL| & Relative clause  & \verb|+| & \verb|V| \\ \hline
\verb|:CARD| & Cardinal & \verb|+| & \verb|N|, \verb|NUM| \\ \hline
\verb|:ORD|  & Ordinal & \verb|+| & \verb|N|, \verb|NUM| \\ \hline
\verb|:PLACEMENT| & Punctuation position & \verb!(pre|post! \verb!|rel)!  & \verb|PX| \\ \hline
%---------------------------------------------------------------
\end{tabular}
\end{small}
\end{center}
\end{table*}
\spacing{1.667}

Table~\ref{Syndep-feats} lists the features used in SynDep.  Most of
these features are, in principle, language independent semantic units.
However, some are specifically used in Spanish (indicated in
Table~\ref{Syndep-feats} with $^*$).  These special features are later
translated (see Section~\ref{matador-translation}). The feature value
\verb|+| is used to indicate that a feature is true.

%----------------------------------------------------------------------
%----------------------------------------------------------------------
\section{Analysis}\label{matador-analysis}
Parsing in Matador utilizes the Connexor parsing system
 \cite{Tapanainen:1997}.  Connexor implements a symbolically driven
approach to parsing based on Constraint Grammar \cite{Karlsson:1990}.

\subsection{Connexor Spanish Parsing}
The input to Connexor is a Spanish sentence.  The output is a
functional dependency that is somewhat incompatible with the
input expected by Exerge.  On the one hand, the functional dependencies for
Spanish include thematic relations such as location and instrument.
These relations are specified directly between a verb and its object
regardless of the existence of a preposition.  In this aspect,
Connexor's output is deeper than what Exerge expects.  On the other
hand, some features closer to the surface form are kept such as
complex verb chains signifying passivization.  Other problems with
parsing with Connexor include its ``over-parsing'' of complex
untranslatable alphanumeric sequences, e.g. document references in the
UN corpus such as {\it AA.33.1.C.ii/1991}, its failing to parse
certain words altogether and its lack of handling of empty categories
(i.e. \verb|*trace*| and \verb|*pro*|).  Moreover, the set of parts of
speech and relationship names are not consistent with what Exerge
expects (including the handling of punctuation and conjunctions).  For
these reasons, the output of Connexor is further processed to create a
more Exerge-compatible output.

\subsubsection{Connexor Pre-processing}

Pre-Connexor processing includes mainly the special tokenization of
complex untranslatable alphanumeric sequences so that they are not
parsed.

\subsubsection{Connexor Post-processing}
The following is the list of processes applied to Connexor's output.
The order of application is important because of interactions among
these processes.  Only some are specific to Spanish. For example, pro-drop restoration is language specific but conjunction handling is
not.

\begin{enumerate}
\item Auxiliary verb chains\\ Auxiliary verb chains are replaced with
the features they specify such as perfect/progressive aspect or
passive voice.
\begin{itemize} 
\item Replace the auxiliary {\it haber} and the verb past-participle feature
with the feature \verb|(:perf +)|
\item Replace the auxiliary {\it estar} and the verb past-participle feature
with the feature\verb|(:voice passive)|
\item Replace the auxiliary {\it estar} and the verb present-participle
feature with the feature \verb|(:prog +)|
\end{itemize}

Children and features of auxiliaries are passed to the
parent. Relevant features are mainly mood, tense, number, person and
gender. A \verb|*pro*| is added in subject position if there is
pro-dropping.  Although the auxiliary verb chain looks like English,
it needs to be processed since the chain makes using subcategorization
frames impossible. Moreover, this is problematic to structural
expansion.

\item Reflexive Clitic ``se''\\ The Spanish reflexive clitic ``se'' is
used to indicate a variety of phenomena such as passivization, or
emphasis \cite{Garcia:1975,Maldonado:1988}. In some cases, the meaning is changed in a pragmatic
non-compositional manner.  For example {\it acordar} is {\it to agree}
but {\it acordarse} is {\it to remember}.  Another example is acusar
{\it to accuse} and acusarse {\it to confess}. Cases where meaning
change occurs are indicated in the translation lexicon as separate
entries from basic verbs.  In Connexor's output, the reflexive clitic
``se'' is always separated from its verb and in some cases, it is not
attached at all. If the reflexive ``se'' is not attached to anything,
it is forced to be attached to the closest Verb.  Currently, the
reflexive clitic is treated in one of two ways depending on whether
the infinitive form of the verb appears in the translation dictionary
with the reflexive clitic attached.

\begin{itemize}
\item If the reflexive form of the verb appears, the clitic is
attached to the verb.
\item Otherwise, the clitic is deleted and the feature
\verb|(:voice passive)| is assigned to the main verb.
\end{itemize}

\item Prepositional Phrases (PP)\\ The government relation between
prepositions and their objects is reestablished: prepositions are to
dominate their objects rather than modifying them. All PPs are related
to their parents as \verb|:mod|.

\item Coordinating Conjunctions \\ Coordinating conjunction structures
are modified to be consistent with the general input specification
described in Section~\ref{input-spec}.

\item Punctuation Handling \\ Structures involving punctuation are
modified to be consistent with the general input specification
described in Section~\ref{input-spec}. This includes reattaching PX
nodes under the first ancestral node where the PX is the child with
the largest or smallest source-order position. The \verb|:placement|
feature is set to \verb|pre| or \verb|post| accordingly.  An exception
is PX nodes placed under \verb|*root*| where they get the feature
\verb|(:placement rel)|.

\item Pro Drop\\ In pro-drop languages such as Spanish, the subject
pronoun of a verb can be dropped but is indicated in the morphology of
the verb.  Every verb must have a subject. So, a place holder for the
pro-dropped subject is added.

\begin{itemize}
\item If the verb has no subject and it is a child of \verb|*root*|, then
a subject \verb|*pro*| is added with the verb's features for number and person.

\item If the verb has no subject and it is {\it not} a child of \verb|*root*|, then
a subject \verb|*trace*| is added with the verb's features for number and person.
\end{itemize}

\item Depassivization\\ 
Fully depassivize verbs as follows:
\begin{itemize}
\item If a verb has the feature \verb|(:voice passive)|, then 
  \begin{enumerate}
  \item Change \verb|:subj| to \verb|:obj| and \verb|:obj| to \verb|:obj2|
  \item Add \verb|*pro*| as \verb|:subj|
\end{enumerate}
\end{itemize}
\item Subordinate clauses and relative clauses\\
\begin{itemize}
\item Remove the word ``que'' introducing subordinate clause where it
has no role as in {\it pienso \textbf{que} Maria est\'{a} aqu\'{i}}
({\it I think \textbf{that} Mary is here})
%\item But keep it in relative clauses where it has a syntactic role
%as in {\it el hombre \textbf{que} ama a Maria está aquí} ({\it the
%man \textbf{who} loves Mary is here}).
\end{itemize}

\item Relations, Parts of Speech (POS) and Features\\ Connexor relations, POS
and features are converted to be compatible with the set defined in
the general input specification described in Section~\ref{input-spec}.
Connexor has more relations, POS and features than Exerge expects as
input. In some cases, information is distributed over features and POS
as in proper nouns being indicated with the feature \verb|<proper>|
and the POS \verb|N|. Spanish specific features such as
\verb|(:tense pret)| or \verb|(:mood cnd)| are kept as is.  They are
translated in the next step to possible English features.

\item Psych Verb Handling\\
The class of psych verbs includes exceptions to the universal thematic
hierarchy such as {\it gustar (enjoy/fancy/like/relish)}, {\it escocer
(hurt/itch)} and {\it envidiar (envy)}.  Since the violation of thematic
hierarchy is a source language word feature, it is handled here.  If
these verbs are detected with a \verb|:subj| and \verb|:obj| (or
\verb|:obj2|), then the \verb|:subj| and the object are inverted.
\end{enumerate}

\begin{figure*}[htb]
\centerline{\includegraphics[width=4.5in]{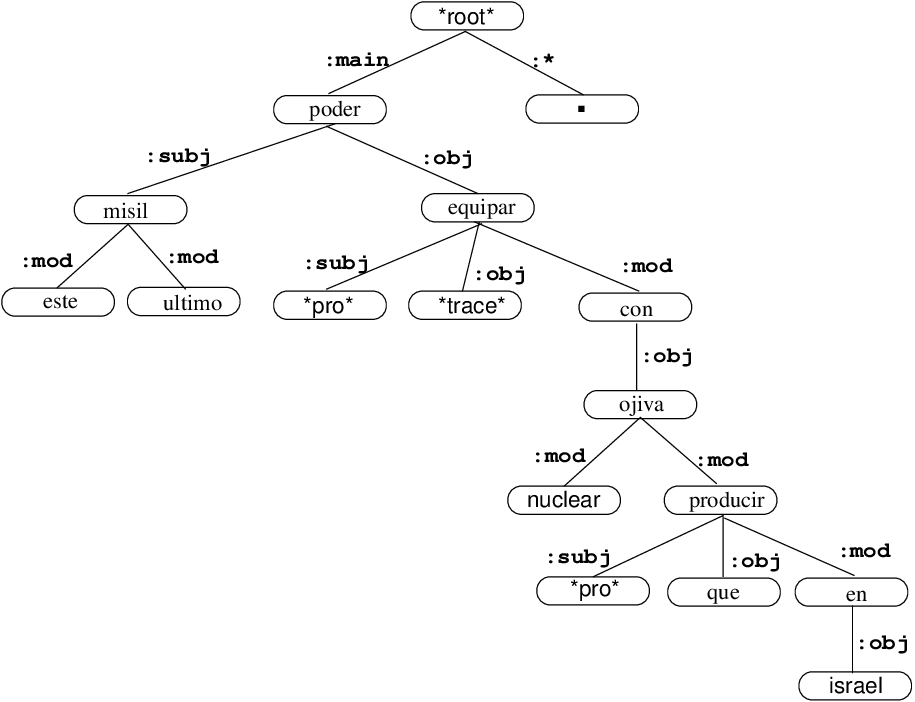}}
\caption{Output of Analysis}
\label{sen23-spanish}
\end{figure*}

Figure~\ref{sen23-spanish} displays the output of parsing for the
Spanish sentence {\it Este \'{u}ltimo misil puede equiparse con ojivas
nucleares que se est\'{a}n produciendo en Israel.} ``This last missile
can be equipped with nuclear warheads which are currently produced in
Israel.''

%%%-------------------------------------------------------------------

\section{Translation}\label{matador-translation}

\subsection{Building the Translation Dictionary}\label{dict-build}

The Spanish-English dictionary was constructed from multiple resources: 
\begin{enumerate}
\item The lexicon of a Spanish Kimmo-based morphological analyzer that
contained English glosses \cite{Dorr:1993}.
\item Spanish-English word-lists from {\it freedict.com}, {\it
spanish.about.com} and the web site of the freely available
multilingual dictionary Ergane \cite{Wilgen:0000}.
\item A Spanish-English word list of abbreviations extracted from a
part of the UN parallel corpus (none of the testing set used later was
included).
%\item A small special dictionary manually created for translating
%features and for cognate-aware back off.
\end{enumerate}

The dictionary had several problems including, but not limited to, (1)
incorrect or nonexistent parts of speech, (2) incorrect lexeme forms
such as including the infinitive particle {\it to} as part of English
verb glosses or the {\it se} clitic where it only acts as a
passivization marker, and (3) spelling errors.  Large parts of the
translation dictionary were manually fixed by one human in 40 hours.
The dictionary also contains some phrasal translations (1-to-n, n-to-m
and n-to-1).

\spacing{1}
\begin{figure}[htb]
\begin{tabular}{p{4.5in}}
\begin{scriptsize} 
\begin{alltt}
abandonar V abandon/desert/forsake/leave/quit
abandonar V cede/give_in/give_up/give_way/relinquish/yield
cesión N abandonment
abandonado AJ forlorn/abandoned/abandonee
abandonamiento N indulgence
abandonar V forsake/abandon/desert/leave/quit/relinquish/shutter/surrender/give_way/succumb
abandono N renunciation/dereliction/failing
abdicación N abandonment/job/task
aflojar V abandon_ship/abate/loose/loosen/relax/slack/slacken/unfasten/untie/sink
deber V owe/should::AUX/must::AUX/have_to::AUX
dejar V lets/assume/bequeath/desert/lay/abandon/allow/let/permit/leave
desamparado AJ abandonee/helpless
desamparar V forsake/abandon
descontar V depreciate/reduce/abandon_ship/cash_up/derate
descuidar V abandon/neglect
expedir V abandon/abdicate/draw_up/give_up/send/dispatch/leave/register
renunciar V abjure/give_up/abandon/renounce/waive/relinquish/surrender

tense:impf FEAT tense:past
tense:pret FEAT tense:past

ico\$ OTHERWISE ic/ical
ente\$ OTHERWISE ent
ción\$ OTHERWISE tion
amente\$ OTHERWISE ly/ely/ally
ía\$ OTHERWISE y 

á OTHERWISE a
é OTHERWISE e
í OTHERWISE i
ó OTHERWISE o
ú OTHERWISE u
ñ OTHERWISE gn/n
\end{alltt}
\end{scriptsize}
\end{tabular}
\caption{A Sample from the Matador Spanish-English Dictionary}
\label{matador-dict}
\end{figure}
\spacing{1.667}

The structure of the translation dictionary is a three-column file
pairing a single Spanish lexeme with a POS and a forward-slash
separated list of English lexemes.  Components of multi-word words are
separated with an underscore (see Figure~\ref{matador-dict}).  By
default, a translation is assumed to affect the lexical choice but not
the POS of the translated word.  If this is not the case, the new POS
is indicated by following the gloss with the marker ``\verb|::|''
(e.g. {\it deber} in Figure~\ref{matador-dict}).

Additionally two other types of entries are included. First, features
are specified as \verb|<feature>:<value>| pairs with the special
``POS'' \verb|FEAT|.  In Figure~\ref{matador-dict}, the Spanish
imperfect tense feature is translated as the English past tense.  The
second type of special entries specify back-off regular expression
operations to be applied to untranslated words.  These are marked with
the special ``POS'' \verb|OTHERWISE|. The first entry of this type in
Figure~\ref{matador-dict} translates words ending with {\it -ico} by
replacing the ending with {\it -ic} and {\it -ical}.  For example,
{\it fantastico} would be translated as {\it fantastic} and {\it
fantastical}, if no entry for it appears in the dictionary. The
overgeneration of these rules is a reasonable back-off that will give
some words a chance of being translated but may not always be correct.

Overall, the dictionary contains 24,278 words and 14 POS with a ratio
of 1.12 POS per word.  The number of word-POS-gloss triples is 50,606
(about 1.86 gloss per word-POS pair and 2.08 gloss per word).  Almost
half of the entries are nouns (48\%). Adjectives make up 20\% and
verbs 18\%.  Proper nouns and Adverbs are 4\% and 3\% respectively.
There are over 900 words (3\%) with unknown part of speech.

\subsection{The Translation Algorithm}\label{trans-algo}
Translation is accomplished through a simple word replacement algorithm.
Matching is done using the word and POS. If the word-POS pair is not
available, the translation algorithm attempts to back off to a union
of all the translations of the word for all available parts of speech.
If this fails, the final back-off is invoked and the unmatched word is
subjected to regular expressions modifications specified in the
\verb|OTHERWISE| rules in the order they appear in.

There are two additional features of the translation algorithm that
are implemented but are not currently used in Matador.  First, the
translation algorithm allows the specification of translation weights
indicating the probability of a word being translated into a
gloss.\footnote{Translation weights can be extracted as part of the
process of building bilingual dictionaries from aligned bilingual
corpora.}  Weights must be normalized.  The weight is specified with
the \verb|%| 
marker attached to the gloss
(e.g. \verb|deber V owe%0.1/| 
\verb|should::AUX%0.9|).
  This information is
passed on through the system and is used in the last statistical
ranking step in Halogen.

The second additional feature is the Translation Memory mode allowing
phrase-to-phrase translation.  This mode detects the appearance of the
\verb|:surface| feature.  If a translation is matched, the feature is
translated; otherwise it is deleted.

\begin{figure*}[htb]
\centerline{\includegraphics[width=4.0in]{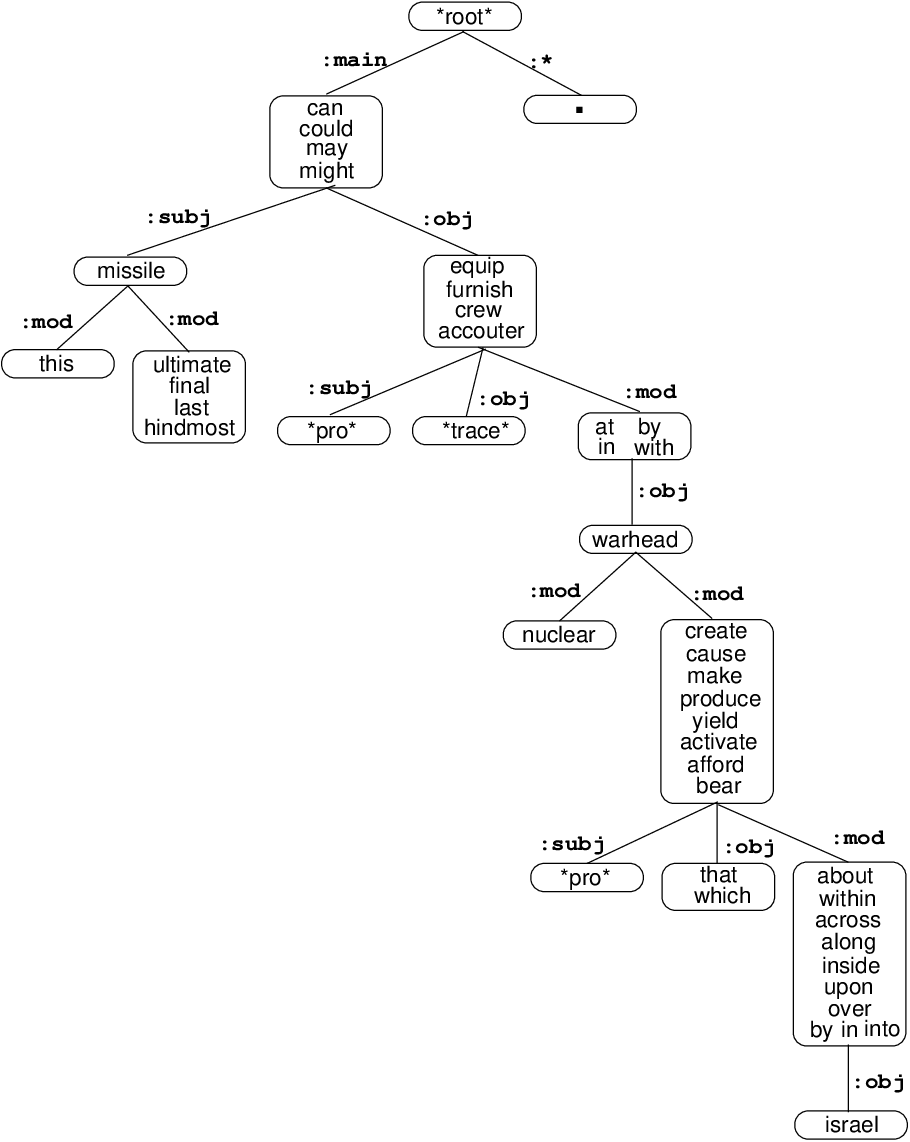}}
\caption{Output of Translation}
\label{sen23-input}
\end{figure*}

Figure~\ref{sen23-input} displays the output of the translation step once
applied to the parse tree in Figure~\ref{sen23-spanish}. 

\section{Resources for {\it Really} Poor Languages}\label{realpoor}
The approach used here to create the translation dictionary and the
choice to use an of-the-shelf parser might not be possible for other
languages.  Efforts like the Translingual Information Detection,
Extraction and Summarization program (TIDES) are focused on fast
acquisition of bilingual resources, even in low-density languages such
as Cebuano.  See, for example, the large (80K line) bilingual
resource extracted in 4 days by the University of Maryland
\cite{Resnik:1999c} at the TIDES surprise language
web site.\footnote{http://www.ldc.upenn.edu/tiki/tiki-index.php}

%Resnik:1999 bibleasparallel
As explained in Section~\ref{system-translation}, the only
restrictions on the translation dictionary are (1) the form of the SL
words needs to be consistent between the parser and the translation
dictionary and (2) the TL words need to be lexemes.  Therefore, it is
possible to use an aligned bilingual parallel corpus together with a
TL morphological analyzer to create a dictionary that maps SL words in
surface form to TL lexemes.  The same bilingual parallel corpus can be
used to infer the structure of parses on the TL side to build a
treebank of the SL side.  This treebank can be used to train
statistical parsers for the SL language
\cite{Charniak:1997a,Collins:1997a}. Statistical parses typically do
not provide lexemes.  This is not a problem for this approach since
the SL word form in the parse and the dictionary would still be
consistent.

The technique described above for creating parsers and translation
dictionaries using parallel corpora is an excellent example of the
{\it generation-heavy approach} applied to the problem of resource
creation as it pertains to GHMT.

\paragraph{}
This chapter described the Spanish-only and Spanish-English components
of the GHMT system Matador.  The availability of a SL parser and a
SL-TL dictionary is the only SL-related requirement for GHMT. The next
chapter describes the implementation details of Exerge, the
English-specific generation component of Matador.
 
%---------------------------------------------------------------------

\chapter{Exerge: Expansive Rich Generation for English}\label{exerge-chapter}

\begin{figure*}[htb]
\centerline{\includegraphics[width=4.5in]{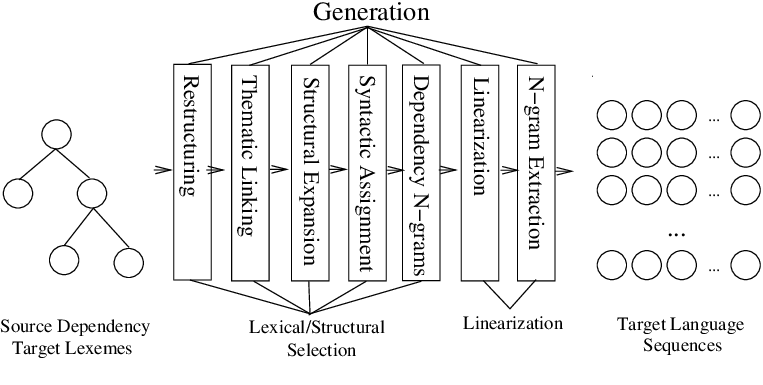}}
\caption{Exerge: Expansive Rich Generation for English}
\label{newsys2}
\end{figure*}

This chapter discusses the implementational details of Exerge, a reusable GHMT
 generation component for translating from other languages
into English.  Exerge --- the core generation engine of Matador ---
consists of seven steps (Figure~\ref{newsys2}).  The first five are
responsible for lexical and structural selection and the last two are
responsible for linearization.  Initially, the source language
syntactic dependency, now with target lexemes, is normalized and
restructured into a syntactico-thematic dependency format.  The
thematic roles are then determined in the thematic linking step.  This
is followed by structural expansion which explores conflated and
inflated variations of the thematic dependency.  The fourth step maps
the thematic dependency to a target syntactic dependency. The first
four steps are all symbolically driven. The fifth step is the first
statistical component where dependency bigram statistics are used for
lexical selection.  After the linearization generates a lattice of
word sequences, they are ranked using a statistical surface n-gram
model.

The next section discusses Exerge's resources.
Section~\ref{input-exerge} discusses the Exerge preprocessing step of
input normalization.  The following seven sections discuss each of
Exerge's steps in the order described above.  Finally,
Section~\ref{exerge-params} presents the many parameters associated
with running Exerge.

\section{Generation Resources}
The generation component utilizes four major resources: a word-class
lexicon, a syntactic-thematic linking map, a categorial-variations
database, and statistical dependency bigram model.

\subsection{Word-Class Lexicon}\label{class}

The word-class lexicon currently contains only verbs and
prepositions. Each of these categories is grouped into classes derived
from the LCS Verb Database (LVD) \cite{Dorr:2001a} and the LCS
Preposition Database (LPD) \cite{Dorr:2001b}.  In the case of verbs,
there are 511 verb classes for 3,131 verbs, totaling 8,650 entries.
An example is shown here:

\begin{sentences}
\item
\begin{footnotesize}
\begin{verbatim}
(DEFINE-WCLASS
 :NUMBER "V.13.1.a.ii"
 :NAME "Give - No Exchange"
 :SENTENCES ("He !!+ed the car to John" "He !!+ed John the car")
 :POS V
 :THETA_ROLES 
       (((ag obl) (th obl) (goal obl to)) ((ag obl) (goal obl) (th obl)))
 :LCS_PRIMS (cause go possessional)
 :SPEC ((ag (animate +)))
 :WORDS (feed give pass pay peddle refund render repay serve))
\end{verbatim}\end{footnotesize}
\end{sentences}

In the case of prepositions, there are 43 preposition
classes, for 125 prepositions, totaling 444 entries.  An
example is shown here:
\begin{sentences}
\item
\begin{footnotesize}
\begin{verbatim}
(DEFINE-WCLASS
 :NUMBER "P.8"
 :NAME "Preposition Class P.8"
 :POS P
 :THETA_ROLES (time)
 :LCS_PRIMS (path temporal)
 :SPEC nil
 :WORDS (until to till from before back_to at after))
\end{verbatim}\end{footnotesize}
\end{sentences}

Note that these entries are only available in the system for
English since it is the target language.  There are no 
equivalent entries for the source language.

%---------------------------------------------------------
\subsection{Syntactic-Thematic Linking Map}

\spacing{1}
\begin{figure}[htb]
\caption{Inventory of Thematic Roles in Exerge}
\label{theta-list-exerge}
\begin{center}
\begin{tabular}{|p{1.3in}|p{2.4in}|}
\hline
\textbf{Thematic Role} &  \textbf{Definition} \\
\hline
AG & agent \\
INSTR & instrument \\
TH & theme \\
EXP & experiencer \\
INFO & information\\
PRED & predicate \\
BEN & beneficiary \\
GOAL & goal\\
SRC & source \\
LOC & location\\
PERC & perceived\\
POSS & possession\\
PROP & event or state \\
PURP & purpose \\
TIME & time \\
MOD-PRED & property modifier \\
MOD-LOC& location modifier \\
MOD-POSS & possessed item modifier \\
MOD-PROP & event or state modifier\\
MOD-TIME & time modifier \\
\hline
\end{tabular}
\end{center}
\end{figure}
\spacing{1.667}

This is a large matrix that was extracted from the LCS Verb Database
(LVD) \cite{Dorr:2001a} and the LCS Preposition Database (LPD)
\cite{Dorr:2001b}.  It relates syntactic ``cases'' to thematic roles.
Syntactic cases include 125 prepositions in addition to {\it :subj,
:obj,\/} and {\it :obj2\/}.  These are mapped to 20 different thematic
roles used in this system (see Figure~\ref{theta-list-exerge}).  These
thematic roles are a subset of the LCS thematic roles shown in
Figure~\ref{theta-list}.  The main difference between the two sets is
that the LCS list contains special thematic roles for prepositions (or
particles) whereas the Exerge thematic roles are never assigned to
prepositions. The total number of links is 341 pairs.  An excerpt of
this resource is shown below.

\begin{sentences}
\item
\begin{footnotesize}
\begin{verbatim}
 (:subj ag instr th exp loc src goal perc mod-poss poss)
 (:obj2 goal src th perc ben)
 (aboard loc goal)
 (about info mod-perc perc poss time purp loc goal pred)
 (according_to purp)
 (across goal loc)
 (in_spite_of purp)
 (in loc mod-poss perc goal poss prop)
\end{verbatim}\end{footnotesize}
\label{linker}
\end{sentences}

%---------------------------------------------------------
\subsection{Dependency Bigram Statistics}\label{depgram-stats}

The dependency bigram language model was created using 127,000 parsed
sentences from the English UN corpus covering over 3 million words.
The parsing was done using Connexor's English parser.  The parses were
post-processed in a similar manner to the post-processing of the
Spanish input described earlier to create a compatible SynDep
treebank. The treebank was traversed and parent-child instance
(lexeme) pairs were counted. The language model is basically the
collection of these counts (totaling 504,039 bigrams for 40,879
lexemes).  Figure~\ref{dep-bigram} shows an excerpt from the language
model.  Each triple denotes a parent node, a child node and the number
of their occurrence.

\spacing{1}
\begin{figure}[htb]
\begin{center}
\begin{small}
\begin{verbatim}
("abandon" "to" 18)           ("afghanistan" "abandon" 1)
("abandon" "*trace*" 11)      ("abandoning" "of" 1)
("abandon" "and" 7)           ("abandon" "respond" 1)
("abandon" "be" 7)            ("abandon" "now" 1)
("abandon" "in" 6)            ("explosive" "abandon" 1)
("abandon" "policy" 6)        ("block" "abandon" 1)
("abandon" "have" 4)          ("in" "abandon" 1)
("abandonment" "of" 4)        ("neglect" "abandon" 1)
("child" "abandoned" 4)       ("invocation" "abandonment" 1)
("abandon" "<comma>" 3)       ("abandon" "cause" 1)
("abandon" "not" 3)           ("abandonment" "human" 1)
("abandon" "they" 3)          ("reduce" "abandon" 1)
("be" "abandon" 3)            ("pledge" "abandon" 1)
("abandon" "for" 2)           ("abandon" "angola" 1)
("abandon" "as" 2)            ("abandon" "facilitate" 1)
("abandon" "effort" 2)        ("israel" "abandon" 1)
("abandon" "by" 2)            ("abandon" "hear" 1)
("abandon" "position" 2)      ("attempt" "abandon" 1)
("abandon" "it" 2)            ("abandon" "area" 1)
("cause" "abandonment" 2)     ("abandonment" "child" 1)
("have" "abandon" 2)          ("as" "abandoning" 1)
("force" "abandon" 2)         ("abandon" "function" 1)
("abandon" "*pro*" 2)         ("nations" "abandon" 1)
("abandon" "child" 2)         ("wound" "abandon" 1)
("abandon" "programme" 2)     ("abandon" "territory" 1)
("urge" "abandon" 2)          
\end{verbatim}
\end{small}
\end{center}
\caption{Sample Dependency Bigram Counts}
\label{dep-bigram}
\end{figure}
\spacing{1.667}

%NYH:is this kosher?  I did not calculate percentages of occurrence here... no probability work
%-----------------------------------------------------------
\subsection{Categorial-Variation Database} \label{catvar}

The Categorial-Variation Database (CatVar) is a database of words and
their categorial variants.  An investigation into the existence of such a
resource found that, so far, none is available.\footnote{The WordNet
project is currently adding such links but only for Nouns and Verbs
(Christiane Fellbaum, pc.).}  This research has involved the creation
of such a resource for English.  The structure of this database is rather
simple: it is flat with an indexing file that is accessible through a
hash table.  For a given word and its optional parts of speech, the
lookup mechanism returns an adjacency list of categorial variants of the
word (including the word itself).  An excerpt is shown here:
\begin{sentences}
\item 
\begin{footnotesize}\begin{verbatim}
 (:V (hunger) :N (hunger) :AJ (hungry))
 (:V (validate) :N (validation validity) :AJ (valid))
 (:V (cross) :N (crossing cross) :P (across))
\end{verbatim}\end{footnotesize}
\label{catvarexample}
\end{sentences}

Given the lack of large-scale resources containing categorial
variations, researchers frequently develop and use alternative
algorithmic approximations of such a resource.  These approximations can
be divided into Reductionist (Analytical) or Expansionist (Generative)
approximations. The former focuses on the conversion of several
surface forms into a common root.  Stemmers such as the Porter Stemmer
 \cite{Porter:1980} are a typical example.  The latter, or expansionist
approaches, overgenerate possibilities and rely on a statistical
language model to rank/select among them. The morphological generator
in Nitrogen is an example of such an approximation
 \cite{Langkilde:1998b}.

There are two types of problems with these approximations: (1)
They are uni-directional and thus limited in usability---A stemmer
cannot be used for generation and a morphological overgenerator cannot
be used for stemming; (2) The crude approximating nature of such
systems cause many problems in quality and efficiency from
over-stemming/under-stemming or over-generation/under-generation.

Consider, for example, the Porter Stemmer, which stems {\it
communication}$_N$, {\it commune}$_N$ and {\it communism}$_N$ to
$commun$.  However, it does not produce this same stem for {\it
communist}$_N$ or {\it communicable}$_{AJ}$ (stemmed to $communist$
and $communic$ respectively).\footnote{For a deeper discussion and
classification of Porter Stemmer's errors, see \cite{Krovetz:1993}.}
Another example is the expansionist Nitrogen morphological generator,
where the morphological feature $+nominalize-verb$ applied to
$develop$ returns eleven variations including $*developage$,
$*developication$ and $*developy$.  Only two are correct
($development$ and $developing$). Such overgeneration multiplied out
at different points in a sentence expands the search space
exponentially, and given various cut-offs in the search algorithm,
might even appear in some of the top ranked choices.

Given these issues, the goal is to build a database of categorial
variations that can be used with both expansionist and reductionist
approaches without the cost of over/under-stemming/generation. 

The CatVar database was developed using a combination of resources and
algorithms including the LCS Verb and Preposition Databases
 \cite{Dorr:2001a}, the Brown Corpus section of the Penn Treebank
 \cite{Marcus:1994}, an English morphological analysis lexicon
developed for PC-Kimmo (Englex) \cite{Antworth:1990}, NOMLEX
 \cite{Macleod:1998}, Longman Dictionary of Contemporary English
(LDOCE)\footnote{An English Verb-Noun list extracted from LDOCE was
provided by Rebecca Green.} \cite{Procter:1978}, WordNet 1.6
 \cite{Fellbaum:1998}, and the Porter stemmer \cite{Porter:1980}.
\hide{The contribution of each of these sources is clearly labeled in the
CatVar database, thus enabling the use of different cross-sections of
the resource for different applications.\footnote{For example, in a
headline generation system (HeadGen), higher Bleu scores were obtained
when using the portions of the CatVar database that are most relevant
to nominalized events (e.g., NOMLEX).}}

Some of these resources were used to extract {\it seed\/} links
between different words (Englex lexicon, NOMLEX and LDOCE). Others
were used to provide a large-scale coverage of lexemes. In the case of
the Brown Corpus, which does not provide lexemes for its words, the
Englex morphological analyzer was used together with the part of
speech specified in the Penn Tree Bank to extract the lexeme form.
The Porter stemmer was later used as part of a clustering step to
expand the seed links to create clusters of words that are categorial
variants of each other, e.g., {\it hunger}$_N$, {\it hungry}$_{AJ}$,
{\it hunger}$_V$, {\it hungriness}$_N$.

The current version of the CatVar (version 2.0) includes 62,232
clusters covering 96,368 unique lexemes.  The lexemes belong to one of
four parts-of-speech (Noun 62\%, Adjective 24\%, Verb 10\% and Adverb
4\%).  Almost half of the clusters currently include one word only.
Three-quarters of these single-word clusters are nouns and one-fifth
are adjectives.  The other one-half are distributed in a Zipf
fashion over clusters from size 2 to 27.

A smaller supplementary database devoted to verb-preposition
variations was constructed solely from the LCS verb and preposition
lexicon using shared LCS primitives to cluster. The database was
inspired by pairs such as {\it cross}$_V$ and {\it across}$_P$ which are used in
Generation-Heavy MT.  But since verb-preposition clusters are not
typically morphologically related, they are kept separate from the
rest of the CatVar database and they were not included in the
evaluation presented in this paper.\footnote{This supplementary database
includes 242 clusters for more than 230 verbs and 29 prepositions.
Other examples of verb-preposition clusters include: {\it avoid}$_V$ and
{\it away\_from}$_P$; {\it enter}$_V$ and {\it into}$_P$; and
{\it border}$_V$ and {\it beside}$_P$ (or {\it next\_to}$_P$).}

The CatVar is web-browseable at {\it http://clipdemos.umiacs.umd.edu/catvar/}.

%[[a comment about why two separate dbases for preps and everything else]]
%----------------------------------------------------------------

\section{Input Normalization}\label{input-exerge}

The input syntactic dependency is converted into a {\it normalized}
dependency representation  in which prepositions are treated as
edge labels (henceforth {\it syntactic relations}) rather than node
labels. Edges in this normalized dependency have two simultaneous
specifications (1) syntactic relation (as in \verb|:subj|, \verb|:obj|, or a
preposition or list of prepositions) and (2) thematic role (as in
$ag$, $th$, or $goal$).

\begin{figure*}[htb]
\centerline{\includegraphics[width=4.5in]{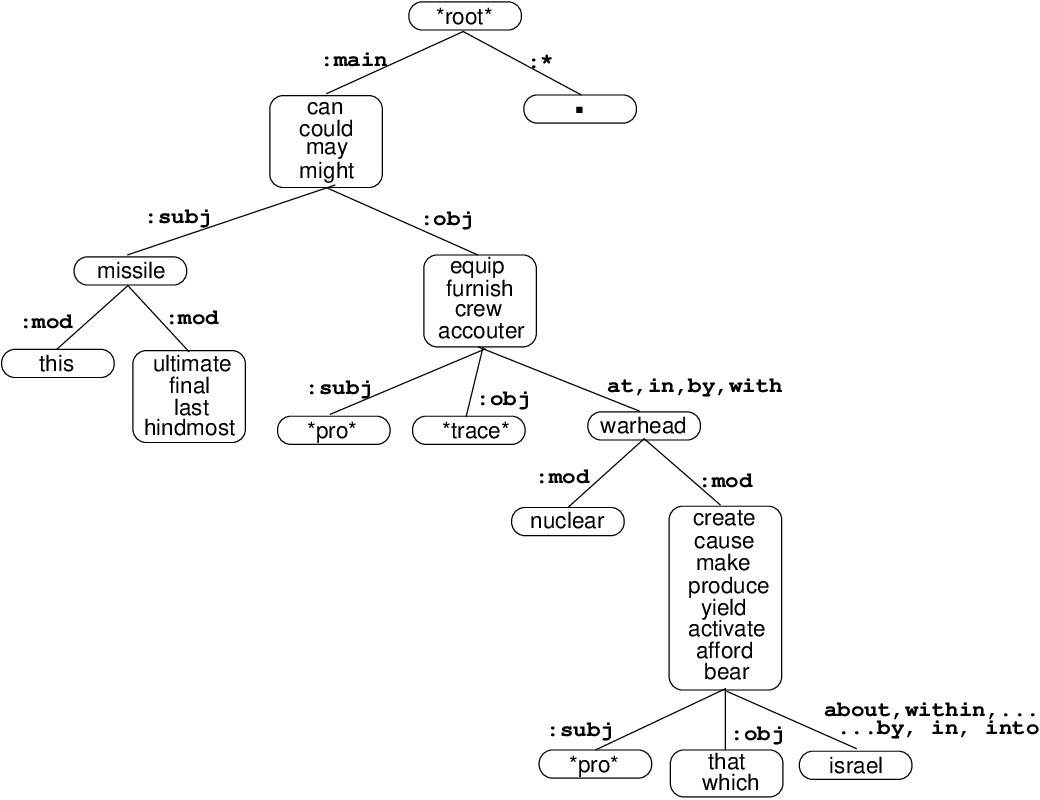}}
\caption{Output of Normalization}
\label{sen23-norm}
\end{figure*}

While the syntactic relations are specified by the input, the thematic
roles are generated in the thematic linking step
(Section~\ref{thetalink-algo}).  The target-correct syntactic
relations are reassigned in the later step of syntactic assignment
(Section~\ref{synass-algo}).  This representation allows access to
both levels simultaneously yet hides them if necessary. 
Figure~\ref{sen23-norm}
exemplifies what the normalized dependency looks like for the input
sentence in Figure~\ref{sen23-input}.

\section{Restructuring}
\spacing{1}
\begin{figure}[htb]
\begin{tabular}{p{4.5in}}
\begin{alltt}
RESTRUCTURE (parent) returns list-of-dependencies

  initialize restructuredParents = NIL

  #Check for restructuring conditions
  if (parent.POS equal V
      and parent.words.POS includes AUX or AV
      and exist (child in parent.children) such-that
          child.synRel equal :obj
          and child.POS equal V)

  #Restructure
  then foreach pos in parent.words.POS
         newParent = replicate parent
         newParent.words = subset of parent.words with pos 
         newParent.POS = pos
         if pos equal V
         then add newParent to restructuredParents
         else swap newParent and newParent.child
              combine arguments of newParent and newParent.child
              add newParent.child to restructuredParents

  return restructuredParents
\end{alltt}
\end{tabular}
\caption{Pseudocode for Restructuring}
\label{pseudo-restruct}
\end{figure}
\spacing{1.667}

The normalized input is then inspected for any violations of
wellformedness. If a source-language verb $parent$ is translated into
an auxiliary (AUX) or an adverb (AV) *AND* $parent$ dominates another
verb, say $child$ in an \verb|:obj| relation, a restructuring of the
dependency is invoked:

\begin{itemize}
\item The translations of $parent$ are separated into different
subsets depending on their POS.

\item For each subset, a restructured dependency is created. Of course,
in cases where a translation does not include a shift in POS, no
restructuring is needed.  The different resulting dependencies are
joined as ambiguous alternatives.

\item If restructuring is necessary, then $parent$ is demoted and
$child$ is promoted to dominate $parent$. Additionally, the
arguments of $parent$ are combined with the arguments of $child$:
if $child$ has a \verb|*trace*| in any syntactic position, this
\verb|*trace*| is what the argument in the subject position of
$parent$ is combined with.
\end{itemize}

\begin{figure*}[htb]
\centerline{\includegraphics[width=4.5in]{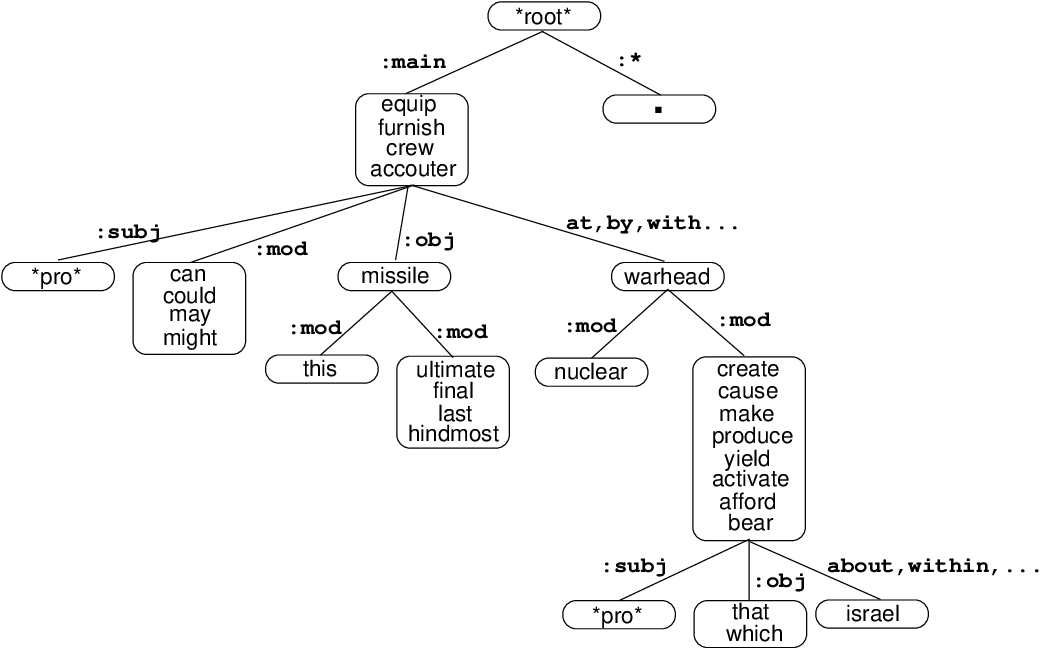}}
\caption{Output of Restructuring}
\label{sen23-rest}
\end{figure*}

Figure~\ref{pseudo-restruct} illustrates the pseudocode for
restructuring.  Figure~\ref{sen23-rest} displays the effect of
restructuring on the sentence in Figure~\ref{sen23-norm}.  The
translations of the Spanish verb {\it poder}, the auxiliaries {\it
can, could,} etc., are demoted to be modifiers of one of their
children, the verb {\it equipar}, which is translated into {\it equip,
furnish,} etc. The subject of the {\it poder} node is merged with the
\verb|*trace*| in the object position of the {\it equipar} node.

%-------------------------------------------------------------------
\section{Thematic Linking}\label{thetalink-algo}

\spacing{1}
\begin{figure}[htb]
\begin{tabular}{p{4.5in}}
\begin{alltt}
THEMATIC-LINKING (synRels, thetaGrid) returns linkSets

  #Create link netweork
  initialize linkNetwork = NIL
  foreach synRel in synRels
    foreach thetaRole in thetaGrid
      if (synRel-thetaRole in  SynThematicLinkingMap)
      then add synRel-thetaRole to linkNetwork

  #Extract all possible link sets
  linkSets = enumerate link sets in linkNetwork
  foreach linkSet in linkSets
    linkSet.score = score(linkSet)
    if (linkSet contains unpaired obligatory thetaRoles)
    then remove linkSet from linkSets
 
  #Rank and Select 
  return all linkSets with maximum score 
\end{alltt}
\end{tabular}
\caption{Pseudocode for Thematic Linking}
\label{pseudo-theta}
\end{figure}
\spacing{1.667}

The syntactic thematic linking is done using a separate module that
takes syntactic relations appearing in the input dependency and the
thematic grid associated with the head verb.  There are usually many
thematic grids resulting from (1) different target translations, (2)
different possible senses associated with a single target translation
(in terms of its verb class memberships) and (3) multiple grids
associated with a single verb class.

For each single thematic grid, a flow network is constructed by
linking it to the syntactic relations appearing with the input.
Entries in the syntactic-thematic linking map constrain the number of
links.  For example, whereas \verb|:subj| links to almost every
possible thematic role, the preposition \verb|aboard| links only to
\verb|(loc goal)|. When a source preposition is ambiguously translated
as a set of target prepositions, all and only the allowed thematic
roles are linked.  All syntactic relations are also linked to the default
thematic modifier role, \verb|mod|.

\begin{figure*}[htb]
\centerline{\includegraphics[]{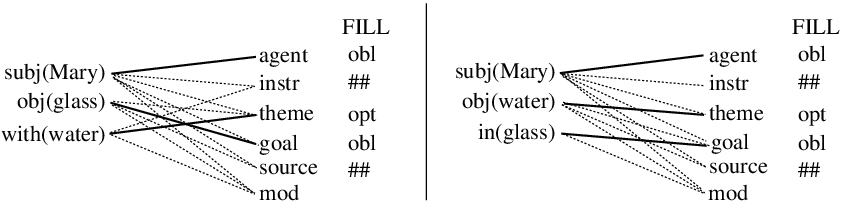}}
\caption{Syntactic-Thematic Linking Example}
\label{fill-link}
\end{figure*}

Figure~\ref{fill-link} illustrates how the correct mapping from syntax
to thematic roles would be done for the two sentences {\it Mary filled the
glass with water\/} and {\it Mary filled water in the glass\/}.
Although the second example is not correct English (albeit good
Korean), the correct roles are assigned mainly because of the
limitations imposed by allowable thematic assignments for the
prepositions.

All of the possible linking sets (sets of paired syntactic-thematic
combinations) are enumerated and scored. Scoring incorporates the
following heuristics:

\begin{itemize}
\item higher weight to pairing obligatory theta roles over optional
theta roles and those over thematic modifier.
\item higher weight to obligatory/optional roles linked to syntactic
arguments over syntactic modifiers.
\item for ambiguous syntactic roles (preposition lists), higher
preference is given to thematic roles linkable by more prepositions.
\end{itemize}

Next, linking sets with at least one obligatory role missing are
eliminated. If no linking sets are left, all are included as a
back-off solution.  Figure~\ref{pseudo-theta} illustrates the
pseudocode for thematic linking for every thematic grid ($thetaGrid$)
associated with a specific verb sense. A later step (not included in 
Figure~\ref{pseudo-theta}) selects amongst the returned link sets:
for each verb (not verb sense), the linking sets with the highest
score over maximum-possible-score ratio are returned.

Although this is a brute force approach and there are excellent
algorithms for finding maximum links \cite{Ford:1962}, there are many
constraints and back-offs that are hard to implement using such
algorithms.  Moreover, the size of the linking networks is relatively
small (the number of syntactic relations or thematic roles is almost
never larger than 10). So, there is no real toll on time/space use.
This step is only applied to verbs and their dependents.

In the rest of this chapter, the relation between arguments and verbs
is represented as a concatenation of the syntactic relation and the
thematic role when it is relevant to display both, e.g.,
\verb|:subj:ag|. For example, {\it Mary filled the glass with water}
is represented as follows:

\begin{sentences}
\item
\begin{small}\begin{verbatim}
(2 \ "fill"  
   :subj:ag (1 \ "mary")
   :obj:goal (4 \ "glass")
   :(with):th (5 \ "water"))
\end{verbatim}\end{small}
\label{ex-fill-glass}
\end{sentences}

Note that a syntactic relation may be a grammatical function such as
\verb|:subj| or it may be a list of one or more prepositions such as
\verb|(with)|.

%4 -------------------------------------------------------------------

%-------------------------------------------------------------------
\section{Structural Expansion}
This step is for exploring alternative structural configurations of
the input.  Several operations are applied here: 
\begin{itemize}
\item First Degree Structural Expansion
  \begin{itemize}
  \item Node-Node Conflation
  \item Node-Relation Conflation
  \item Node-Node Inflation
  \item Node-Relation Inflation
  \end{itemize}
\item Second Degree Structural Expansion
    \begin{itemize}
    \item Node-Node Inflation followed by Node-Relation Conflation
    \item Node-Relation Inflation followed by Node-Node Conflation
    \end{itemize}
\end{itemize}

In all these cases, new created alternatives are OR'ed to the original
structure.

\subsection{Node-Node Conflation}\label{n-conf}

\spacing{1}
\begin{figure}[htb]
\begin{tabular}{p{4.5in}}
\begin{alltt}
NN-CONFLATION-VARIANTS (parent) returns list-of-dependencies

  initialize conflatedParents = NIL

  foreach child in parent.children 
    # Check for conflatability
    if (exist(child' = VerbCatVar(child.word))
        and child'.lcsPrim equal parent.word.lcsPrim
        and child'.thetaGrid 
            equal (parent.word.thetaGrid - child.thetaRole))
    # Apply conflation
    then newParent = replicate(parent)
        remove child from newParent.children
        newParent.word = child'
        add newParent to conflatedParents

  return conflatedParents
\end{alltt}
\end{tabular}
\caption{Pseudocode for Node-Node Conflation Variant Creation}
\label{pseudo-conf}
\end{figure}
\spacing{1.667}

For each one of the arguments of a given verb in the tree, the verb
($parent$) and argument ($child$) pair are checked for conflatability.
A pair is conflatable if (1) there exists a verb $child'$ that is a
categorial variation of the argument, (2) $child'$ and $parent$ both
share the same main lexical conceptual primitive \hide{(not including
\verb|CAUSE|)}, and (3) $child'$ can assign the same thematic roles
that are assigned by $parent$ except for the one that is assigned to
$child$. Figure~\ref{pseudo-conf} illustrates the pseudo-code for this
process.

\hide{
For each one of the arguments of a given verb in the tree, the verb
($V_{head}$) and argument ($Arg$) pair are checked for conflatability.
A pair is conflatable if (1) there exists a verb $V_{conf}$ that is a
categorial variation of the argument, (2) $V_{conf}$ and $V_{head}$
both share the same main lexical conceptual primitive \hide{(not
including \verb|CAUSE|)}, and (3) $V_{conf}$ can assign the same
thematic roles that are assigned by $V_{head}$ except for the one that
is assigned to $Arg$.}

Take the following example for the Spanish {\it yo le di pu\~{n}aladas a
Juan (I gave stabs to Juan)\/} which results in the following thematic
dependency tree after linking is done:

\begin{sentences}
\item
\begin{small}\begin{verbatim}
(3 \ "give"  
   :ag (1 \ "I")
   :th (4 \ "stab")
   :goal (6 \ "juan"))
\end{verbatim}\end{small}
\label{ex-stab-1}
\end{sentences}

The theme \verb$stab$ has a verb categorial variation \verb$stab$
which belongs to two different verb classes, the Poison Verbs (as in
{\it crucify, electrocute\/}, etc.) and the Swat verbs (as in {\it bite,
claw\/}, etc.).  Only the first class shares the same lexical conceptual
primitive as the verb \verb$give$ \verb$(CAUSE GO)$.  Moreover, the
verb \verb$stab$ requires an agent and a goal for the stabbing.
Therefore, a conflated instance is created in this case:

\begin{sentences}
\item
\begin{small}\begin{verbatim}
(3 \ "stab"  
   :ag (1 \ "I")
   :goal (6 \ "juan"))
\end{verbatim}\end{small}
\label{ex-stab-2}
\end{sentences}

If the sentence were, say, {\it I gave the stab a name\/}, the
categorial variation for \verb$stab$ would have failed because it
stood in a goal relationship with its parent.  However, in this
example, the verb categorial variation of $name$, $name_V$, would have
succeeded in conflating with \verb$give$. The result would be {\it I named
the stab\/}.

To deal with the possibility of node loss in cases where nodes
modifying conflated nodes are ignored, a metric called {\it Semantic
Mass} is used to disqualify lossy conflations.  Semantic Mass is
defined as the number of nodes with open class categories (V, N, PN,
AJ and AV).  The default minimum mass loss allowed is 1 (when a
conflation happens).

\subsection{Node-Relation Conflation}\label{r-conf}
\spacing{1}
\begin{figure}[htb]
\begin{tabular}{p{4.5in}}
\begin{alltt}
NR-CONFLATION-VARIANTS (parent) returns list-of-dependencies

  initialize conflatedParents = NIL

  foreach child in parent.children 
    # Check for conflatability
    if (exist(confVerb = VerbCatVar(child.synRel))
        and confVerb.lcsPrim equal parent.word.lcsPrim
        and parent.word.thetaGrid is-subset-of confVerb.thetaGrid 
        and confVerb.thetaGrid.obligatory 
            is-subset-of  parent.word.thetaGrid

    # Apply conflation
    then newParent = replicate(parent)
        newParent.children.child.synRel = *
        newParent.word = confVerb
        add newParent to conflatedParents

  return conflatedParents
\end{alltt}
\end{tabular}
\caption{Pseudocode for Node-Relation Conflation Variant Creation}
\label{pseudo-conf2}
\end{figure}
\spacing{1.667}

For each of the arguments of a given verb in the tree, the verb
($parent$) and argument syntactic relation ($child.synRel$) pair are
checked for conflatability.  The syntactic relation accessed here is
the translated SL syntactic relation (i.e. preposition).  A pair is
conflatable if (1) there exists a verb $confVerb$ that is a categorial
variation of $child.synRel$ (2) $confVerb$ and $parent$ both share the
same main lexical conceptual primitive and (3) $confVerb$ covers all
the thematic roles in $parent$ and $parent$ covers all the
obligatory thematic roles in the theta grid of
$confVerb$.  Note here that the only difference between node-node conflatability
and node-relation conflatability is in condition (3), which is the result
of no node disappearance and thus no thematic role reassignment.
Figure~\ref{pseudo-conf2} illustrates the pseudo-code for this
process.

\hide{
For each of the arguments of a given verb in the tree, the verb
($V_{head}$) and argument syntactic relation ($Arg.synrel$) pair are
checked for conflatability.  The syntactic relation accessed here is
the translated SL syntactic relation (i.e. preposition).  A pair is
conflatable if (1) there exists a verb $V_{conf}$ that is a categorial
variation of $Arg.synrel$ (2) $V_{conf}$ and $V_{head}$ both share the
same main lexical conceptual primitive and (3) $V_{conf}$ covers all
the thematic roles in $V_{head}$ and $V_{head}$ covers all the
obligatory thematic roles in the subcategorization frame of
$V_{conf}$.  Note here that the only difference between conflatability
and relation conflatability is in condition (3), which is the result
of no node disappearance and thus no thematic relation reassignment.}

Consider the following example: {\it John went across the river} which results in
the following thematic dependency tree after linking is done:
%NYH [Spanish??]

\begin{sentences}
\item
\begin{small}\begin{verbatim}
(3 \ "go"  
   :subj:th (1 \ "juan")
   :(across):goal (4 \ "river")
\end{verbatim}\end{small}
\label{ex-cross-river-1}
\end{sentences}

%NYH:[[use paired representation all over?]]

The syntactic relation \verb$across$ has a verb categorial variation
\verb$cross$ which belongs to two different verb classes, the Verbs of
Contiguous Location (as in {\it bridge, intersect}, etc.) and the
Verbs of Inherently Directed Motion (as in {\it enter, invade\/},
etc.).  Only the second class shares the same lexical conceptual
primitive as the verb \verb$go$ (\verb$GO$---the other one is
\verb$BE$).  Moreover, the subcategorization frame of the verb
\verb$cross$ covers all the thematic roles in this example's verb
\verb$go$. The verb \verb$go$ covers all the obligatory roles required
for \verb$cross$.  Therefore, a conflated instance is created in this
case, which includes deleting the syntactic relation \verb$across$ and
substituting it with a syntactic-relation place-holder, \verb|*|:

\begin{sentences}
\item
\begin{small}\begin{verbatim}
(3 \ "cross"  
   :subj:th (1 \ "juan")
   :*:goal (6 \ "river"))
\end{verbatim}\end{small}
\label{ex-stab-3}
\end{sentences}

The syntactic relation to replace \verb|*| (in this case \verb|:obj|)
is assigned in the later step of Syntactic Assignment
(Section~\ref{synass-algo}).

%-------------------------------------------------------------------

\subsection{Node-Node Inflation}
\spacing{1}
\begin{figure}[htb]
\begin{tabular}{p{4.5in}}
\begin{alltt}
NN-INFLATION-VARIANTS (parent) returns list-of-dependencies

  initialize inflatedParents = NIL

  # Check for inflatability
  if (exist(parent' = NounCatVar(parent.word) 
                      or AdjectiveCatVar(parent.word))
      and exist(infVerb) such-that
               infVerb is-member-of LightVerbs
               and infVerb.lcsPrim equal parent.word.lcsPrim
      and exist(thetarole) such-that
               infVerb.thetaGrid 
                   equal (parent.word.thetaGrid + thetaRole))

  # Apply inflation
  then newParent = replicate(parent)
      newParent.word = infVerb
      add parent' to newParent.children in relation thetarole        
      add newParent to inflatedParents

  return inflatedParents
\end{alltt}
\end{tabular}
\caption{Pseudocode for Node-Node Inflation Variant Creation}
\label{pseudo-inf-1}
\end{figure}
\spacing{1.667}
Inflation can be viewed as anti-conflation or the reverse algorithm of
conflation, where the existing head verb ($parent$) is inflated to a
new verb $infVerb$ and an argument categorially related to $parent$ 
($parent'$).

Inflation differs from conflation in that its search space is
enormous: (1) the verb $infVerb$ is unconstrained by categorial
variation and in principle could be any verb of the English language,
(2) $parent'$ can categorially relate in many ways to $parent$
(not just as a verb categorial variation); and (3)
the specific thematic role relating $infVerb$ and $parent'$ is
unspecified.  The biggest problem is the first of the three.  To
address it, the search space of $infVerb$ is restricted to the
following light verbs: {\it do, give, keep, have, make, cause, take,
get, put, be, become, send, move, let, go,} and {\it stay}.

To consider a verb, $parent$ for inflation, (1) there should exist
$parent'$, a noun or adjective categorial variation of $parent$; (2)
there should exist a verb, $infVerb$, selected from the list of light
verbs such that (2.a) $infVerb$ and $parent$ both share the same main
lexical conceptual primitive and (2.b) $infVerb$ can assign the same
thematic roles that are assigned by $parent$ to all of its arguments;
(3) there should exist a thematic roles for $parent'$ that can be
assigned by $infVerb$. 

Conditions (1), (2.a) and (2.b) are parallels of the conditions for
conflatability discussed earlier.  As for condition (3), a preference
is associated with obligatory roles over optional roles and
modifiers. If there are multiple obligatory roles, then inflation
fails. \hide{Inflation also fail for multiple optionals} Once
$infVerb$, $parent'$, and the thematic role are established, inflated
variants of the input are constructed.  Figure~\ref{pseudo-inf-1}
illustrates the pseudo-code for this process.

To demonstrate this process, the example discussed in
Section~\ref{n-conf} is reversed, starting with the following:

\begin{sentences}
\item
\begin{small}\begin{verbatim}
(2 \ "stab"  
   :ag (1 \ "I")
   :goal (3 \ "juan"))
\end{verbatim}\end{small}
\label{ex-stab-4}
\end{sentences}

The head verb \verb$stab$ ($parent$) has the nominal categorial
variations \verb$stab$ and \verb$stabbing$ ($parent'$).\footnote{The
adjective $stabbed$ is also generated but it is eliminated later.}
Out of 88 light verb thematic grids corresponding to various light
verb class memberships, 7 grids corresponding to {\it give}(3), {\it
take}(3) and {\it put}(1) are used in creating multiple inflations.
For the rest of this example, only the first generated inflation will
be used. In this inflation, the light verb {\it give}, covers the
thematic roles \verb|:ag| and \verb|:goal| of the already existing
arguments.  The leftover role, \verb|:th| is assigned to $parent'$:

\begin{sentences}
\item
\begin{small}\begin{verbatim}
(2 \ "give"  
   :ag (1 \ "I")
   :th (INF \ (*or* "stab" "stabbing"))
   :goal (3 \ "juan"))
\end{verbatim}\end{small}
\label{ex-stab-5}
\end{sentences}

%------------------------------------------------------------------
\subsection{Node-Relation Inflation}
\spacing{1}
\begin{figure}[htb]
\begin{tabular}{p{4.5in}}
\begin{alltt}
NR-INFLATION-VARIANTS (parent) returns list-of-dependencies

  initialize inflatedParents = NIL

  # Check for inflatability
  if (exist(newSynRel = PrepositionCatVar(parent.word)) 
      and exist(infVerb) such-that
               infVerb is-member-of LightVerbs
               and infVerb.lcsPrim equal parent.word.lcsPrim
      and infVerb.thetaGrid equal parent.word.thetaGrid)

  # Apply inflation
  then foreach child in parent.children 
      if child.thetaRole is-member-of newSynRel.thetaRoles
      then newParent = replicate(parent)
           newParent.word = infVerb
           newParent.children.child.synRel = newSynRel
           add newParent to inflatedParents

  return inflatedParents
\end{alltt}
\end{tabular}
\caption{Pseudocode for Node-Relation Inflation Variant Creation}
\label{pseudo-inf-2}
\end{figure}
\spacing{1.667}
Node-Relation Inflation has a comparable level of complexity to node-node
Inflation.  The existing verb ($parent$) is inflated to a new verb
$infVerb$ and an argument syntactic relation ($newSynRel$).

The conditions for inflatability of $parent$ are: (1) there should
exist $newSyn$ $Rel$, a prepositional categorial variation of
$parent$ (there can be more than one); (2) there should exist a
verb, $infVerb$, selected from the list of light verbs such that (2.a)
$infverb$ and $parent$ both share the same main lexical conceptual
primitive and (2.b) $infVerb$ can assign the same thematic roles that
are assigned by $parent$ to all of its arguments.

For each argument with a theta role assignable by the preposition(s),
$newSyn$ $Rel.thetaRoles$, a separate inflated instance is created where the
preposition $newSynRel$ is the syntactic relation of that
argument. Figure~\ref{pseudo-inf-2} illustrates the pseudo-code for this
process.

Take the following example (reversing the one presented earlier): {\it John
crossed the river} which results in the following thematic dependency
tree after linking is done: %NYH [Spanish??]

\hide{Node-Relation Inflation has a comparable level of complexity to node-node
Inflation.  Here, the existing ($V_{head}$) is inflated to a new verb
$V_{inf}$ and an argument syntactic relation ($Arg.synrel_{head}$).

The conditions of inflatability of $V_{head}$ are: (1) there should
exist $Arg.synrel_{head}$, a prepositional categorial variation of
$V_{head}$ (there can be more than one); (2) there should exist a
verb, $V_{inf}$, selected from the list of light verbs such that (2.a)
$V_{inf}$ and $V_{head}$ both share the same main lexical conceptual
primitive and (2.b) $V_{inf}$ can assign the same thematic roles that
are assigned by $V_{head}$ to all of its arguments.

For each argument with a theta role assignable by the preposition(s)
$Arg.synrel_{head}$, create a separate inflated instance where the
preposition $Arg.synrel_{head}$ is the syntactic relation of that
argument.

Take the following example (reversing the one presented earlier): {\it John
crossed the river} which results in the following thematic dependency
tree after linking is done: %NYH [Spanish??]
}

\begin{sentences}
\item
\begin{small}\begin{verbatim}
(2 \ "cross"  
   :subj:th (1 \ "juan")
   :obj:goal (3 \ "river"))
\end{verbatim}\end{small}
\label{ex-stab-6}
\end{sentences}

The preposition \verb$across$ is a categorial variant of the verb
\verb$cross$.  Of the light verbs, {\it go} and {\it move} are
candidates for inflations satisfying condition (2) above.  Since
\verb$across$ can only assign the theta roles \verb|goal| and
\verb|loc|, it can only be assigned as syntactic relation for the
\verb|goal|:

\begin{sentences}
\item
\begin{small}\begin{verbatim}
(2 \ "go"  
   :subj:th (1 \ "juan")
   :(across):goal (3 \ "river")
\end{verbatim}\end{small}
\label{ex-cross-river-2}
\end{sentences}
%------------------------------------------------------------------------

\subsection{Second Degree Structural Expansion}

In second degree structural expansion, any of the operations described
earlier can be composed with another operation or with itself. This is done
by applying the second operation to the output of the first
operation.  In this current implementation, only two
combinations are allowed: Node-Node Conflation after Node-Relation Inflation and
Node-Relation Conflation after Node-Node Inflation.  The combination
of two operations is intended to handle cases of complex head swapping
such as {\it Juan atraves\'{o\/} el r\'{\i}o nadando (John crossed the
river swimming)} to {\it John swam across the river}.  This example is
handled by the first of the two second degree operations implemented:

Starting with the following thematic dependency tree after linking:
\begin{sentences}
\item
\begin{small}\begin{verbatim}
(3 \ "cross"  
   :subj:th (1 \ "juan")
   :obj:goal (4 \ "river")
   :mod:mod (6 \ "swim"))
\end{verbatim}\end{small}
\label{ex-cross-river-3}
\end{sentences}
The first operation of node-relation inflation inflates the verb {\it cross} to
{\it go across} resulting in the following:
\begin{sentences}
\item
\begin{small}\begin{verbatim}
(3 \ "go"  
   :subj:th (1 \ "juan")
   :(across):goal (4 \ "river")
   :mod:mod (6 \ "swim"))
\end{verbatim}\end{small}
\label{ex-cross-river-4}
\end{sentences}
The second operation of node-node conflation conflates verbs {\it swim} and
{\it go} resulting in the following:
\begin{sentences}
\item
\begin{small}\begin{verbatim}
(3 \ "swim"  
   :subj:th (1 \ "juan")
   :(across):goal (4 \ "river"))
\end{verbatim}\end{small}
\label{ex-cross-river-5}
\end{sentences}

%---------------------------------------------------------------------
\section{Syntactic Assignment}\label{synass-algo}

In this step, the thematic dependency is turned into a full target
syntactic dependency. This step includes three types of operations.
First, the syntactic relations originating from translating the source
language syntactic relations or resulting from various structural
expansion operations are overwritten or adjusted by target-correct
syntactic relations specified by the verb class subcategorization
frames. In cases where multiple alternations are specified for a verb
class, all are generated. For example, the following thematic
dependency is associated with the verb {\it give}, which has two
alternations:
\begin{sentences}
\item
\begin{small}\begin{verbatim}
(3 \ "give"  
   :ag (1 \ "I")
   :th (4 \ (*or* "stab" "stabbing"))
   :goal (6 \ "juan"))
\end{verbatim}\end{small}
\label{ex-stab-7}
\end{sentences}
The resulting syntactic structures associated with the two
alternations are as follows:
\begin{sentences}
\item \begin{small}\begin{verbatim}
(3 \ "give"  
   :subj (1 \ "I")
   :obj (4 \ (*or* "stab" "stabbing"))
   :mod (6 \ (*or* "to" "toward" "towards" "into" "at" "by") 
           :obj (6 \ "juan")))
\end{verbatim}\end{small}
\label{ex-stab-8a}
\end{sentences}
\begin{sentences}
\item \begin{small}\begin{verbatim}
(3 \ "give"  
   :subj (1 \ "I")
   :obj (4 \ (*or* "stab" "stabbing"))
   :obj2 (6 \ "juan"))
\end{verbatim}\end{small}
\label{ex-stab-8b}
\end{sentences}

At this step, all the prepositions associated with a specific theta
role are generated: in example \ref{ex-stab-8a}, the prepositions
{\it to, toward, towards, into, at,} and {\it by} are generated as possible
realizations of the theta role {\it goal}.  This overgeneration is
addressed in the statistical dependency pruning step next.

Second, verb class specifications are enforced. These include verb
aspect (\verb"prog" or \verb"perf"), verb form (\verb"fin" or
\verb"inf"), and argument categorial specifications (\verb"N" or
\verb|AJ|). If a specified categorial variation is not available for
the argument instance, it is kept as is as a back-off.
For example, the main verb for the Spanish {\it tengo hambre (I have
hunger)\/} translates into (have, own, possess, and be).  For the last
verb (be), there are different classes that have different
specifications on the verb's second argument: a noun and an adjective.
This of course results in {\it I am hungry\/} and {\it I am hunger\/}
in addition to {\it I (have/possess/own) a hunger}.  That is where
statistical extraction is most valuable; to decide
which sequence is more likely.

Third, to combat the explosive overgeneration, a rudimentary
statistically driven clean up of syntactic relations for verbs and
nouns is done here.  This step uses dependency bigram and unigram counts of
words to eliminate prepositions in word-preposition pairs unseen in
the corpus.  The goal of this step is not to rank per se but rather to
reduce the search space. 

This process is done as follows. For every $parent$ node, if $POS$ is
\verb|V| then for each $parent$-$child$ node pair, drop every
preposition $P$ in the syntactic relation that does not have a
($parent.instance$ $P$) or ($P$ $child.instance$) dependency
bigram. $parent.instance$ refers to any of the ambiguous instances of
the node $parent$. If the resulting list of prepositions is empty, restore
it to include all prepositions as a back-off.

If $POS$ is \verb|N|, then check if the $parent$-$child$ pair has
a preference for a \verb|:mod| relation (that is no prepositions). If
not, a similar trimming of prepositions as done for verbs is applied
here. A $parent$-$child$ pair is determined to prefer a \verb|:mod|
relation if its highest ($parent.instance$ $child.instance$)
dependency bigram count is larger than the highest projected
dependency trigram of ($parent.instance$ $P$ $child.instance$). A
projected dependency trigram is calculated as the lowest of the
bigrams ($parent.instance$ $P$) and ($P$ $child.instance$).

For example, the preposition in the Spanish {\it el mundo \textbf{en}
desarrollo (the world \textbf{in} development/developing)} is replaced by a
\verb|:mod| relationship since the dependency bigram ({\it developing
world}) is more common than ({\it world in}) and ({\it in
development/developing}) in English.

%------------------------------------------------------------------------

\section{Statistical Dependency Pruning}

\spacing{1}
\begin{figure}[p]
\begin{tabular}{p{4.5in}}
\begin{alltt}
DEPENDENCY-PRUNE (parent) returns pruned-parent
  SCORE-WORDS parent
  prunedParent = SELECT-WORDS parent
  return prunedParent

SCORE-WORDS (parent) calculates scores and preferences
  foreach child in parent.children
     SCORE-WORDS child
  foreach parentWord in parent.words
    parentWord.score = 0
    foreach child in parent.children
      select childWord from child.words 
             such-that maximize (bigram(parentWord,childWord) 
                                 + childWord.score)
      set prefer(parentWord,childWord) = true
      parentWord.score = parentWord.score 
                         + bigram(parentWord,childWord) 
                         + childWord.score

SELECT-WORDS (parent) returns pruned-parent
  select parentWord from parent.words
         such-that maximize parent.word.score
  parent.words = parentWord
  foreach child in parent.children
    SELECT-CHILD-WORDS (parentWord,child)  
  return parent

SELECT-CHILD-WORDS (parentWord,child) returns pruned-child
  select childWord from child.words
         such-that prefer(parentWord,childWord)
  foreach grandChild in child.children
    SELECT-CHILD-WORDS (childWord,grandChild)  
  return child
\end{alltt}
\end{tabular}
\caption{Pseudocode for Dependency Pruning}
\label{pseudo-prune}
\end{figure}
\spacing{1.667}

This step prunes the ambiguous nodes in the SynDep forest produced by
the last step using dependency bigram statistics (see
Section~\ref{depgram-stats}).

For each SynDep tree in the SynDep forest, a dynamic programming
algorithm with memoization is used to calculate the dependency bigram
scores over all the nodes of that tree.  The tree is traversed bottom
up and each subtree's score is maximized: for each parent instance a
child instance is picked that maximizes the sum of the parent-child
dependency bigram and dependency bigram score of the child.  The child
unigram count is used as a tie breaker.  Once the scoring is
completed, selecting the best unambiguous tree is straightforward.
The scores are not length normalized since all the possible ambiguous
trees have the same structure and number of nodes.
Figure~\ref{pseudo-prune} illustrates the pseudocode for this step.

Associated with dependency pruning is the parameter \verb|*depgram|
\verb|Max| \verb|Tree| \verb|Selection*| which specifies the number of
trees in the input SynDep forest to pass on to the next step.  The
default is to select everything.  If a specific number is used, the
same dependency bigram scores described earlier are used to rank the
trees.  This might cause some problems since these scores are not
length normalized---an issue relevant if deciding among trees (not
within forms of a specific tree).

\section{Linearization}

Linearization is implemented using the Oxygen Linearization
engine~ \cite{Habash:2000} with a slightly modified Oxyl English
grammar written originally for English generation in an LCS-based MT
system~ \cite{Habash:2003b}.

\spacing{1}
\begin{figure}[htb]
\begin{tabular}{p{4.5in}}
\begin{alltt}
:Rule \%XP
  (== ((\$pos @pos))
   ?? (&eq \$pos V) -> (do \%VP (&VP-mod-order @this))
   ?? (&eq \$pos C) -> (do \%CP (&CP-mod-order @this))
   ?? (&eq \$pos AX) -> (@inst)
   ?? (&eq \$pos PRO) -> (do \%ProP (&PRO-mod-order @this))
   ?? (&eq \$pos Num) -> (do \%NumP (&Num-mod-order @this))
   ?? (&eq \$pos D) -> (do \%DP (&DP-mod-order @this))
   ?? (&eq \$pos AJ) -> (do \%AP (&AP-mod-order @this))
   ?? (&eq \$pos AV) -> (do \%AP (&AP-mod-order @this))
   ?? (&eq \$pos N) -> (do \%NP (&NP-mod-order @this))
   ?? (&eq \$pos PN) -> (do \%PNP (&PNP-mod-order @this))
   ?? (&eq \$pos P) -> (do \%PP (&PP-mod-order @this))
   ?? (&eq \$pos PX) -> ((&PX-realize @inst) @mod)
   ?? (&eq \$pos *) -> (do \%*P (&order-unknown @this))
   ?? (&eq \$pos X) -> () 
   -> ((&score @this) @inst))
\end{alltt}
\end{tabular}
\caption{Oxyl Main Rule for Accessing Different Phrase Rules}
\label{lin-example-0}
\end{figure}
\spacing{1.667}

\spacing{1}
\begin{figure}[htb]
\begin{tabular}{p{4.5in}}
\begin{alltt}
:Recast &NP-MOD-order
  (@this 
      <? (:mod-post    / (:MOD)) (&eq @that.pos P)
      <? (:mod-num     / (:MOD)) (&eq @that.pos NUM) 
      <? (:mod-predet  / (:MOD)) (&is-predet @that.inst)
      <? (:mod-postdet / (:MOD)) (&is-postdet @that.inst)	
      <? (:mod-det     / (:MOD)) (&is-det @that.inst) 	 
      <? (:mod-int     / (:MOD)) (&is-int @that.inst)
      <? (:mod-age     / (:MOD)) (&in @that.inst \$mod-age)
      <? (:mod-color-2 / (:MOD)) (&in @that.inst \$color-2)
      <? (:mod-color-1 / (:MOD)) (&in @that.inst \$color-1)
      <? (:mod-color   / (:MOD)) (&in @that.inst \$color)
      <? (:mod-part    / (:MOD)) (&in @that.inst \$mod-part)
      <? (:mod-prov    / (:MOD)) (&in @that.inst \$mod-prov )
      <? (:mod-denom   / (:MOD)) (&in @that.inst \$mod-denom)
      <? (:mod-adj     / (:MOD)) (&eq @that.pos AJ)
      <? (:mod-sub     / (:MOD)) (&eq @that.pos V)
      << (:mod-noun    / (:MOD)))
\end{alltt}
\end{tabular}
\caption{Oxyl Recast for Noun Phrase Modifiers}
\label{lin-example-1}
\end{figure}
\spacing{1.667}

\spacing{1}
\begin{figure}[p]
\begin{tabular}{p{4.5in}}
\begin{alltt}
:Rule \%NP
  (== ((\$mod-det @mod-det)
       (\$num @num)
       (\$inst (&surface-format @inst))
       (\$morpheme (&if (&and (&eq @num PL) 
                              (&neq \$inst "*empty*"))
                        +plural 
                        +singular)))
   -> (@mod-px-1
       @CC
       @mod-predet
       ((x / (&article \$num)) ++ (:cond (&null \$mod-det)))  
       (\$mod-det ++ (:cond (&neq \$num PL)))
       @mod-postdet
       (@mod-num ++ ^NOdet) 
       (@mod-int ++ ^NOdet)
       (@mod-adj ++ ^NOdet)
       (@mod-age ++ ^NOdet)
       (@mod-color-1 ++ ^NOdet)
       (@mod-color-2 ++ ^NOdet)
       (@mod-color ++ ^NOdet)
       (@mod-part ++ ^NOdet)
       (@mod-prov ++ ^NOdet)
       (@mod-noun ++ ^NOdet)
       (@mod-denom ++ ^NOdet)
       (&if (&eq @ABBR +) 
            (&ABBR-realize \$inst) 
            (& \$inst +- \$morpheme))   
       @mod-post
       @mod-sub
       @mod-CC
       @mod-px+1))
\end{alltt}
\end{tabular}
\caption{Oxyl Rule for Noun Phrases}
\label{lin-example-2}
\end{figure}
\spacing{1.667}

The Oxyl English grammar contains 12 phrase-structure rules and 11
recast rules. Figure~\ref{lin-example-0} shows the actual Oxyl rule
used to decide which phrase structure rule to fire.\footnote{For a
detailed presentation of Oxyl's syntax, see \cite{Habash:2001b}.}  In
almost all cases, the subtree to linearize, \verb|@this|, is recasted
first to create some internal ordering amongst its modifiers. For
example, the recasting function \verb|&VP-mod-order| is applied before
firing the verb phrase rule \verb|%VP|.

Morphological generation is implemented using Nitrogen's morphological
generator.  The small size of the grammar does not limit its
overgenerative power since a single phrase structure rule generates
multiple ambiguous linearizations.  For example, the VP rule has three
modes for active voice, passive voice with direct object and passive
voice with indirect object.  Another example is the generation of all
permutations for noun-phrase (NP) modifiers when no specific order
could be decided.

This overgeneration, is constrained by access to other databases and
algorithms.  For example, additional Lisp code supplements the grammar
to handle small VP structure (e.g. generating sequences such as {\it
will not have been being beaten}) and to correct translations of
closed classes by limiting/expanding a seed word and set of
features. The latter is necessary to generate a surface form for
restored pro drops (\verb|*pro*|) from a set of features.  The grammar
is also supplemented with a database of word subclasses such as
temporal modifiers, adjective classes (color, age, number, etc.) and
determiner classes (pre-determiner, determiner, post-determiner). This
knowledge allows the grammar to constrain the relative order of NP
modifiers and sentential complementizers \cite{Quirk:1985}.  Similar
solutions have been used before by other generation systems of all
types (symbolic, statistical and hybrid)
\cite{Mann:1985,Shaw:1999,Malouf:2000,Habash:2003b}. Figures~\ref{lin-example-1}
and \ref{lin-example-2} show the actual Oxyl recast and rule used to
implement the generation of NPs. First, the NP modifiers are recasted
into finer modifier subclasses using semantic word-class membership,
e.g. color or age, or part of speech.  Then, the following order of
modifiers is applied in \verb|%NP|:

\spacing{1}
\begin{itemize}
\item pre-phrasal punctuation
\item coordinating conjunctions ({\it and, or,} etc.)
\item pre-determiners ({\it all, most\_of,} etc.)
\item an optional article if no determiners are specified
\item determiners ({\it a, the,} etc.)
\item post-determiners ({\it first, last,} etc.)
\item numbers
\item intensifiers ({\it very, so,} etc.)
\item ``most-adjectival'' adjectives ({\it important, basic,} etc.)
\item age modifiers ({\it old, young,} etc.)
\item color intensifiers ({\it pale, dark,} etc.)
\item color modifiers ({\it reddish, greenish} etc.)
\item colors ({\it blue, orange, } etc.)
\item participle adjectives ({\it distinguishing, adjusted,} etc.)
\item provenance adjectives ({\it foreign, chinese,} etc.)
\item nouns
\item denominal adjectives ({\it social, original} etc.)
\item the head noun (with morphological generation and special rules for abbreviation realization)
\item prepositional phrases
\item subordinate clauses
\item coordinated phrases
\item post-phrasal punctuation
\end{itemize}
\spacing{1.667}

If multiple modifying children are recasted into one modifier
class (color, age, etc.), all permutations of these children are
linearized automatically.

\paragraph{}
Constituents under the \verb|*root*| node are generated in their
surface language order.  If the \verb|:surface| feature is detected,
both its value and a linearization of the node it appears in are
output.\footnote{The :surface feature specifies a phrase-based
translation of the node and all its children (see
Section~\ref{trans-algo}).}

\section{Statistical Ranking}
The statistical ranking is implemented using the Forest Ranker
component of the generation system Halogen
 \cite{Langkilde:2000,Langkilde:2002}.  While in the Halogen system,
the Halogen grammar generates forests directly, Oxygen linearization
does not. This is because it was originally designed for Nitrogen.
To address this problem, the lattices are converted into forests.

The format of a forest input file is a sequence of lines specifying
forest nodes' label, weight and expansion:

\verb!    <LABEL> [<WEIGHT>]? -> <LABEL>+ | "<WORD>"!

\verb!<LABEL>! conforms to the regular expression
\verb|[a-zA-Z0-9]+.[0-9]+| (the label is solely defined by the numeric
sequence to the right of the period; the alphanumeric characters to
the left of the period are used only for readability).  The
right-hand-side of \verb|->| represents the sequence into which the forest node
\verb|<LABEL>| is expanded.  An ambiguous expansion is represented by
multiple lines with the same left-hand-side. The only exception to
this is the first line specifying the \verb|top| or root of the
forest.  The weight value is optional.

Before converting a lattice, it is modified slightly to remove empty
words and break up multi-word words, both of which are not allowed in
forests.  The lattice is then traversed and repeating sequences are
assigned the same forest subtree label.  A lattice \verb|SEQ| or
\verb|WRD| generates a single forest node specification and a lattice
\verb|OR| generates multiple node specifications.

The following lattice is converted into the forest
in Figure~\ref{ex-forest-halogen}.
 
\begin{sentences}
\item
\begin{small}\begin{verbatim}
(SEQ
 (OR (SEQ (WRD "the" 1) (WRD "bread" 1) (WRD "should" 1) 
     (WRD "be" 1) (OR (WRD "buttered" 1) (WRD "butterred" 1)))
     (SEQ (OR (WRD " " 1) (WRD "the" 1) (WRD "a" 1) (WRD "an" 1))
          (WRD "bilberry" 1) (WRD "should" 1) (WRD "be" 1)
          (WRD "breaded" 1))))
\end{verbatim}\end{small}
\label{ex-lattice-halogen}
\end{sentences}

\spacing{1}
\begin{figure}[p]
\begin{tabular}{p{4.5in}}
\begin{alltt}
top      -> OR.18
OR.18  1 -> OR.17
OR.18  1 -> SEQ.9
OR.17  1 -> SEQ.16
OR.17  1 -> SEQ.15
SEQ.16 1 -> WRD.13 WRD.4 WRD.5 WRD.14
SEQ.15 1 -> OR.12 WRD.13 WRD.4 WRD.5 WRD.14
WRD.14 1 -> "breaded"
WRD.13 1 -> "bilberry"
OR.12  1 -> WRD.11
OR.12  1 -> WRD.10
OR.12  1 -> WRD.2
WRD.11 1 -> "an"
WRD.10 1 -> "a"
SEQ.9  1 -> WRD.2 WRD.3 WRD.4 WRD.5 OR.8
OR.8   1 -> WRD.7
OR.8   1 -> WRD.6
WRD.7  1 -> "butterred"
WRD.6  1 -> "buttered"
WRD.5  1 -> "be"
WRD.4  1 -> "should"
WRD.3  1 -> "bread"
WRD.2  1 -> "the"
\end{alltt}
\end{tabular}
\caption{A Halogen Forest}
\label{ex-forest-halogen}
\end{figure}
\spacing{1.667}

The forest specification is passed on to Halogen's Forest Ranker which
outputs the top one sequence given a specified language model and
modeling scheme.  The language model is created using the CMU
Statistical Toolkit \cite{Clarkson:1997}.  In the following chapter,
more detail is presented on which language models are used.  Halogen
allows six modeling schemes: unigram, bigram, trigram, and all three
with length normalization.

The rest of this section examines the output of statistical ranking
for two Spanish sentences.  To focus on issues related to surface
n-gram models, these examples are run without structural n-gram
pruning.

\begin{sentences}
\item Yo le di pu\~{n}aladas a Juan.
\label{stab}
\begin{footnotesize}\begin{alltt}
I stabbed John .   [ LENGTH 4, SCORE 0.670270 ]
I gave a stab at John .   [ LENGTH 7, SCORE -2.175831 ]
I gave the stab at John .   [ LENGTH 7, SCORE -3.969686 ]
I gave an stab at John .   [ LENGTH 7, SCORE -4.489933 ]
I gave a stab by John .   [ LENGTH 7, SCORE -4.803054 ]
I gave a stab to John .   [ LENGTH 7, SCORE -5.045810 ]
I gave a stab into John .   [ LENGTH 7, SCORE -5.810673 ]
I gave a stab through John .   [ LENGTH 7, SCORE -5.836419 ]
I gave a knife wound by John .   [ LENGTH 8, SCORE -6.041891 ]
I gave John a knife wound .   [ LENGTH 7, SCORE -6.212851 ]
\end{alltt}\end{footnotesize}

\item Juan tiene hambre.
\label{hunger}
\begin{footnotesize}\begin{alltt}
John is hungry .   [ LENGTH 4, SCORE -7.111878 ]
John hunger .   [ LENGTH 3, SCORE -7.474780 ]
John is starved .   [ LENGTH 4, SCORE -7.786015 ]
John is a hunger .   [ LENGTH 5, SCORE -8.173432 ]
John has a hunger .   [ LENGTH 5, SCORE -8.613148 ]
John is a famine .   [ LENGTH 5, SCORE -8.666368 ]
John is the hunger .   [ LENGTH 5, SCORE -8.829170 ]
John is the famine .   [ LENGTH 5, SCORE -8.871368 ]
John be hungry .   [ LENGTH 4, SCORE -9.038840 ]
John is a starvation .   [ LENGTH 5, SCORE -9.105497 ]
\end{alltt}\end{footnotesize}
\end{sentences}

In both examples, the system generates several valid English
translations expressing a wide range of linguistic phenomena such as
conflation and the dative alternation.  This is accomplished purely by
GHMT's TL resources without any specification of, or linking to, the SL
structures {\it dar pu\~{n}aladas} or {\it tener hambre}.  The most
correct form of the output is ranked highest in both cases: {\it I
stabbed John} and {\it John is hungry}.  However, the ranking of the
other choices does not reflect fluency or accuracy well.  For example,
{\it I gave John a knife wound\/} ranks much lower than {\it I gave an
stab at John\/} although the former is more fluent.  The
generation of {\it John is a hunger\/} as a variant of {\it John is
hungry\/} is an inaccurate translation.  These issues can be traced
back to either the symbolic component's overgeneration or the
statistical component's {\it under-\/}extraction.

One case highlighting the issue of fluency is the generation of the
sequence {\it John hunger\/} in example~\ref{hunger}.  Here, the
symbolic rules are not enforcing any subject-verb agreement, which
results in allowing the sequence {\it John hunger\/} in the generated
word lattice together with {\it John hungers\/}.  However, the
statistical model fails to rank {\it John hunger\/} lower than {\it
John hungers\/}, which does not even make it to the top-ten sequences.
This failure is likely due to the smoothing model used for handling
unseen bigrams, which depends on the word unigrams instead.  {\it
Hunger\/} is a more common unigram than {\it hungers\/}.

Another case relevant to the fluency issue is the under-specification
of preposition selection for the verb {\it give\/} in
example~\ref{stab}.  The current constraint is that the selected
preposition {\it could\/} assign the thematic role, in this case, {\it
goal\/}.  Thus, the preposition {\it by\/} selected for {\it I gave a
stab by John\/} has the locational not the agentive sense.  The
statistical model failure here is likely due to uni/bigrams enforcing
fluency locally on a very small window.  

%A possible solution on the
%statistical side is to use structural n-gram language models (similar
%to \cite{Bangalore:2000a}) to capture long-distance dependencies
%between the verb and its modifiers.

The case of generating {\it John is a hunger\/} in
example~\ref{hunger} reflects the dependency of GHMT on TL statistical
knowledge as opposed to translingual knowledge of translation
divergences.  The argument here is that generating the metaphoric {\it
John is a hunger\/} is a ``compromise'' of accuracy worth taking when
generated with more likely sequences such as {\it John is hungry\/}.
If the SL input was a metaphoric {\it John BE hunger}, then other
verbs besides {\it be} would not be generated to start with and the
smaller search space will allow the less likely metaphoric expression
to be selected.  This argument is, of course, hard to evaluate---and
in example~\ref{hunger}, {\it John is a hunger\/} ranks higher than
the poetic {\it John has a hunger\/}.  This ordering is a result of
the statistical extraction use of bigrams in the current system which
picks {\it John is a X} over {\it John has a X} regardless of $X$.

%----------------------------------------------------------------
\section{Exerge Parameters}\label{exerge-params}
\spacing{1}
\begin{table*}
\caption{Exerge System Parameters}
\label{exerge-params-table}
\begin{center}
\begin{small}
\begin{tabular}{|p{1.2in}|p{0.5in}|p{3.0in}|}
\hline
\textbf{Parameter} & \textbf{Values} &  \textbf{Definition} \\ 
\hline
%\verb!dictionary! & \verb!<FILE>! & Translation dictionary \\ \hline
\verb!thematicLinking! & \verb!T|nil! & Apply Thematic Linking; if set to \verb|nil|, no expansion takes place \\ \hline

\verb!retainOriginal! & \verb!T|nil! & Retain a copy of the original SynDep in the output of structural expansion\\ \hline
\verb!restructuring! & \verb!T|nil! & Run the restructuring module \\ \hline

\verb!structural-! \verb!  Expansion! & \verb!T|nil! & Run structural expansion; \\ \hline
\verb!maxConflation! & \verb!<NUM>! & Maximum number of structural expansions generated by conflation. \\ \hline
\verb!maxRConflation! & \verb!<NUM>! & Maximum number of structural expansions generated by relation conflation. \\ \hline
\verb!maxInflation! & \verb!<NUM>! & Maximum number of structural expansions generated by inflation. \\ \hline
\verb!maxRInflation! & \verb!<NUM>! & Maximum number of structural expansions generated by relation inflation. \\ \hline
\verb!secondDegree-! \verb!  StrExpansion! & \verb!T|nil! & Run second degree structural expansion\\ \hline
\verb!maxMassLoss! & \verb!<NUM>! & The maximum number of semantically relevant nodes allowed to be lost in structural expansion \\ \hline

\verb!depgramPruning! & \verb!T|nil! & Run dependency-based statistical pruning \\ \hline
\verb!depgramMaxTree! & \verb!<NUM>|! \verb!  nil! & The number of top SynDep trees to be selected, if set to \verb|nil|, all trees are selected \\ \hline

\verb!halogenModel! & \verb|1-6| & Halogen modeling scheme: 1,2,3 specifies unigram, bigram and trigram respectively; 4,5,6 specifies length normalized unigram, bigram and trigram respectively \\ \hline
\verb!halogenLMfile! & \verb!<FILE>! & Halogen language model file (CMU statistical toolkit binary language model format) \\ \hline

\end{tabular}
\end{small}
\end{center}
\end{table*}
\spacing{1.667}

Currently, Exerge has 14 parameters specifying different run modes or
features.  Table~\ref{exerge-params-table} defines these parameters
and there possible values.  The \verb!thematic! \verb!Linking! parameter
controls whether thematic linking takes place or not.  If thematic
linking is not invoked, structural expansion and syntactic assignment
are not invoked.  The \verb!retainOriginal! parameters instructs
Exerge to keep a copy of the input and pass it to the output together
with other overgenerated variants.  

The \verb!restructuring! and \verb!structuralExpansion! parameters
controls whether the restructuring and structural expansion modules
are invoked or not. The parameters \verb!maxConflation!,
\verb!maxRConflation!, \verb!maxInflation!, \verb!maxRInflation! set
the maximum number allowed for variations created by the various
structural expansion submodules.  In the current implementation of
Matador all of these four values are set to 10.
The parameter \verb!secondDegreeStrExpansion! controls whether second
degree expansion is invoked or not.  This parameter is ignored
(considered \verb|nil|) if the parameter \verb!structuralExpansion! is
set to \verb|nil|.  The parameter \verb!maxMassLoss! sets the maximum
number of semantically relevant nodes allowed to be lost in
structural expansion.  This parameter was discussed in
Section~\ref{n-conf}.  

The parameter \verb!depgramPruning! controls whether the structural
n-gram pruning module is invoked or not.  The \verb!depgramMaxTree!
parameter defines the maximum number of trees to select after
structural n-gram pruning is invoked.  The value of
\verb!depgramMaxTree! is set to \verb|nil| in Matador, which causes
all pruned trees to be selected.  

The \verb!halogenModel! and \verb!halogenLMfile! parameters specify
the mode in which the Halogen forest ranker is run (e.g. unigrams,
bigrams, etc.) and the language model file it uses, respectively.

\paragraph{}
This chapter described the implementational details of Exerge.  In the
next chapter, Matador and all of its components, including Exerge, are
evaluated.

%%%---------------------------------------------------------------%%%
%%%---------------------------------------------------------------%%%
%%%---------------------------------------------------------------%%%
%%%---------------------------------------------------------------%%%
\chapter{Evaluation}\label{eval-chapter}

This chapter presents an evaluation of the performance of Matador
as a complete system followed by several intrinsic evaluations of
specific Matador components.  Discussions of results and issues
observed in the test data are included in each section where they
are relevant.

\section{Overall Evaluation}
The goal of this evaluation is to determine the output quality and
robustness of Matador. For purposes of comparison, four systems are
evaluated using test sets from three corpora with different genre.

\subsection{Evaluation Metrics}
Three automatic metrics are used:

\begin{itemize}
\item BLEU\\ As described in Chapter~\ref{related-work-chapter}, Bleu
(BiLingual Evaluation Understudy) is a method of automatic translation
evaluation that is quick, inexpensive and language independent and
most importantly correlates highly with human evaluation
\cite{Papineni:2001}.  Bleu is used as a metric for translation
quality.\footnote{Throughout this chapter, Bleu scores are presented
multiplied by 100.} Bleu is used with all 1 to 4-grams with no context
sensitivity.  Confidence intervals are computed with a sample size of
50 for 95\% confidence level. The Bleu specific version number is
v1.03.

\item Reference Modified Precision (RMP)\\ This is the modified
precision value of the gold standard (or ``reference'' in Bleu's
lingo) given one or more test sets.  Bleu allows multiple references
to evaluate a single test.  RMP is the exact opposite: the precision
of a single reference, or gold standard, is determined given one or
more test sets.  The goal is to understand how much of the reference is
covered by which test sets and to measure the overlap among these test
sets.  Bleu is used to calculate RMP with the gold standard used
as test and different combinations of the evaluated systems' output as
references.  The RMP value is not a Bleu score, since no length
penalty is applied.  The value is simply the geometric mean of the 1
to 4-gram precision (which itself is a sub-score in Bleu).  The specific
variation of this metric used in this chapter is RMP-1,
which is the reference {\it unigram} precision.

\item Runtime (in seconds) is used as a measure of system efficiency.
\end{itemize}

\subsection{Evaluated Systems}
The following are the evaluated systems:

\begin{itemize}
\item Gisting (GIST): This simple approach is basically
word-to-word translation \cite{Resnik:1997b}.  A dictionary of 391,026
Spanish surface to English lexeme pairs is used with a unigram
language model to resolve any ambiguity.  This system is considered
the baseline.

\item Systran (SYST): This is a commercial Transfer-based (purely
symbolic) MT system.  The version used is Systran Spanish-English
Professional edition with four translation glossaries (Political
Science, Military Science, Legal and
Business/Economics).\footnote{http://www.systransoft.com/} Systran's
Spanish-English has been developed over several hundred person-years
and is considered here the industry standard of Spanish-English MT.

\item IBM Model 4 (IBM4): This is a primarily statistical MT system
 \cite{Brown:1990,Brown:1993,Berger:1994}.  The translation model was
trained using Giza \cite{Al-Onaizan:1999} on 50,000
Spanish-English sentence pairs from the UN Spanish-English corpus
 \cite{Graff:1994}.  Simple tokenization was used and consisted of
down-casing all words and separating all punctuation marks.  The language
model is built from the English side of the training data in addition
to 450 thousand sentences from the English side of the Arabic-English
UN corpus \cite{Xu:2002}. Decoding is done using ISI ReWrite Decoder
 \cite{Germann:2000}.\footnote{Using fast greedy decoding, model 4
translation and bigrams language modeling.}

\item Matador (MTDR): All of the system's modules described in
Chapters~\ref{matador-chapter} and~\ref{exerge-chapter} are used.
SN-gram pruning is used only for lexical selection within
dependencies. The parameters for conflation and inflation are set to
allow a maximum of 10 variants per dependency.  The surface N-gram
language model used in ranking is the same as that used in the IBM4
system described above.
%comment on why using same LM  and not changing with genre!
 The Halogen ranking scheme used is bigrams with length normalization.
\end{itemize}

The four letter system identifiers, e.g., SYST or MTDR, are used in
the rest of this chapter to mark examples produced by their
corresponding systems.  The two letter identifiers, EN and SP, are
used to mark examples from the human gold standard English reference
and the Spanish input, respectively.

\subsection{Test Sets}

\begin{table*}
\begin{center}
\caption{Evaluation Test Sets}
\label{testsets}
\begin{small}
\begin{tabular}{|p{2.0in}|p{0.7in}|p{0.7in}|p{0.7in}|}
\hline
  & \textbf{UN} & \textbf{FBIS} & \textbf{Bible} \\ \hline
{\it Size (sentences)} & 2,000 & 2,000 & 1,000 \\ \hline
{\it Sentence Length (words)} & 15.39 & 19.27 &  16.38 \\ \hline
\end{tabular}
\end{small}
\end {center}
\end{table*}

Three blind test sets are evaluated: (1) 2,000 sentences from the UN
Spanish-English corpus \cite{Graff:1994}; (2) 2,000 sentences from the
FBIS Spanish-English corpus;\footnote{The U.S.Foreign Broadcast
Information Service (FBIS).} and (3) 1,000 sentences from the
Bible. See Table~\ref{testsets}.  Each Spanish sentence had one
English (gold standard) translation that was used as the Bleu
reference.  The one-reference behavior of Bleu is not optimal, but,
unfortunately, there are no Spanish-to-multiple-English parallel
corpora similar to those available for Chinese and Arabic
\cite{multichin,multiarab}.

\begin{figure*}[htb]
\centerline{\includegraphics[width=4.5in]{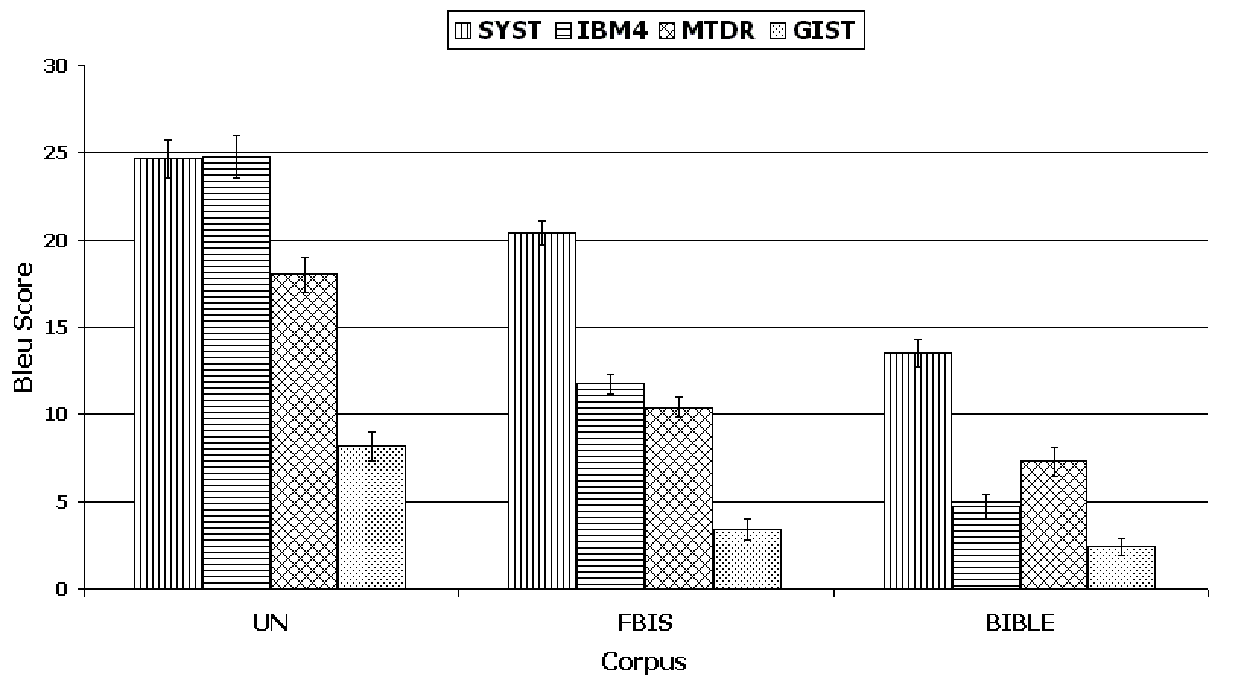}}
\caption{Overall Evaluation Results}
\label{diss-bleu}
\end{figure*}

The three corpora, UN, FBIS and Bible, were selected to cover a wide
range of genre to examine the behavior of the evaluated systems under
different conditions.  This is important since poverty of resources
forces systems to be trained or built using whatever resources are
available, which may not necessarily be the same as what needs to be
translated.  The results of the overall evaluation are shown in
Table~\ref{bleu-table} and Figure~\ref{diss-bleu}.

\spacing{1}
\begin{table*}
\begin{center}
\caption{Overall Evaluation Results}
\label{bleu-table}
\begin{small}
\begin{tabular}{|p{1.0in}|p{1.0in}|p{1.0in}|p{1.0in}|}
\hline
  & \textbf{UN} & \textbf{FBIS} & \textbf{Bible} \\ \hline
SYST & 24.66 +/- 1.09 &   20.39 +/- 0.70  &  13.52 +/- 0.8 \\ \hline
IBM4 & 24.78 +/- 1.21 &   11.76 +/- 0.55  &   4.71 +/- 0.66 \\ \hline
MTDR & 18.01 +/- 1.00 &   10.42 +/- 0.57  &  7.29 +/- 0.83 \\ \hline
GIST & 8.17 +/- 0.81 &    3.40 +/- 0.61  &   2.40 +/- 0.50 \\ 
\hline
\end{tabular}
\end{small}
\end {center}
\end{table*}
\spacing{1.667}

The evaluation shows that although MTDR scores lower than IBM4 on the
corpus where all language models were trained (UN), MTDR has a higher
degree of robustness and scores higher when tested on text with new
genre (Bible).  SYST and GIST are the best and worst respectively for
all corpora.\footnote{All Systems ran successfully on all sentences
except that MTDR failed on a total of 21 sentences out of all 5,0000
(0.42\%). Failure happened exclusively at the last stage, in Halogen
statistical ranking.}

As for runtime, GIST was the fastest, finishing all 2,000 sentences
in the UN corpus in less than 16 seconds.  This is followed by SYST
(90 seconds\footnote{This run was done on a Pentium 4 PC with 1.7Ghz and 512MB
of memory}), IBM4 (8,495 seconds) and finally MTDR
(14,155 seconds).\footnote{Except for SYST, all other systems ran on
a SparcIII, with 750Mhz and 1GB of memory.}

The rest of this section compares the behavior of MTDR against IBM4
since SYST and GIST's performance make them excellent upper and lower
bounds---score-wise and also by the amount of work needed to create
them.

\subsection{Bleu Bias}
One problem with Bleu scoring when only a single reference is used is
that lexical choice becomes more important for evaluation than other
criteria that are dependent on the correct lexical choice.  For
example, the generation of a synonym of a word in the correct relative
word order to another word scores less than the correct lexical choice
for the two words in the wrong relative order.  Moreover, all
``incorrect'' lexical choices are treated equally regardless of how
close or different the chosen words are to the correct words.  Even
morphologically related words are not considered correct. There were
many instances of such cases in the UN test data:
 
\begin{sentences}
\item \label{bb1}SP: Los programas de ajuste estructural  se \textbf{han} {\it \textbf{aplicado rigurosamente}}.\\
EN: Structural adjustment programmes \textbf{had} been {\it \textbf{rigorously implemented}}.\\
IBM4: structural adjustment programmes \textbf{have} been {\it \textbf{applied strictly}}.\\
MTDR: Programmes of structural adjustment \textbf{have} been {\it \textbf{added rigurosament}}.

\end{sentences}

In example \ref{bb1}, the correct auxiliary ({\it have}) was selected
by both IBM4 and MTDR but the gold standard used it in a different
tense ({\it had}).  The difference, a single tense feature with
minimal effect on understanding is penalized as much as failing to
translate a word or mistranslating a word.  In the same example, no
distinction in penalty is made between {\it strictly} and the
incorrect back-off translation {\it rigurosament}.  In the following
examples, the lexical choice is dangerously misleading yet the
same basic word-mismatch penalty is applied:

\begin{sentences}
\item \label{bb2}SP: Se \textbf{instalaron} tres nuevos mercados rurales.\\
EN: Three new rural markets were \textbf{established}.\\
IBM4: \textbf{minefields} three new rural markets .\\
MTDR: Three new rural markets were \textbf{installed}.\\
\item \label{bb3}SP: El Programa Mundial de la OMS sobre el SIDA \textbf{coordina} la vigilancia de esta enfermedad en el mundo.\\
EN: World-wide AIDS surveillance is \textbf{co-ordinated} by the WHO Global Programme on AIDS.\\
IBM4: the global programme of force on aids \textbf{bangladesh} the monitoring of this disease in the world .\\
MTDR: The worldwide programme of the WHO on the aids \textbf{coordinates} the maintenance of this disease in the world.
\end{sentences}

In example~\ref{bb2}, the verb {\it instalar (establish)} was
translated in IBM4 as {\it minefield} and in example~\ref{bb3}, the
verb {\it coordinar (coordinate)} is translated as {\it bangladesh}!
In both examples, the MTDR translations, {\it install} and {\it
coordiantes} receive a penalty equal to that received by {\it
minefield} and {\it bangladesh}.

These last two examples, both from the UN test data where IBM4 scored
high, exemplify an interesting pattern of behavior in statistical MT
systems.  IBM4 did either extremely well or extremely poorly at
lexical choice. In some cases in the UN corpus, it generated almost
perfect phrases, which it most likely saw in the training data.  Its
performance deteriorated significantly in the other two test sets.
When IBM4 had to deal with a previously unseen word or sequence, it
randomly picked words for translation and, in some cases, dangerously
changed the meaning.

In the next two sections, two evaluations of the effect of these biases
are presented.

\subsubsection{Inflectional Variation Bias}\label{ivbias}
To determine the effect of inflectional variations on Bleu scores,
samples of 100 sentences from each the three tested corpora and their
English gold standard references were morphologically analyzed using
Connexor's English parser.  The effect of morphological analysis is to
normalize all the compared words (in test and reference sets) over
inflectional variations.

\begin{sentences}
\item \label{ivb}Raw: The \textbf{activities supported} through IFPPE during 1989 \textbf{are described} below.\\
Morph: the \textbf{activity support} through ifppe during 1989 \textbf{describe} below.  
\end{sentences}

Example~\ref{ivb} highlights some of the changes to the form of
sentences. The first sentence (marked as Raw) is what a sentence looks
like before inflectional normalization.  The second sentence (marked
as Morph) is the result of inflectional normalization.  The features
of number and aspect/tense are removed (as in {\it activities
$\rightarrow$ activity} and {\it supported $\rightarrow$ support}.
Since morphological analysis was done as part of parsing, features
such as voice are also normalized.  This can be seen in {\it be
described $\rightarrow$ describe}.

Since this set is only a sample of the original test set, Bleu scoring
was done for the raw sentences and the morphologically analyzed
sentences.  The results are presented in Table~\ref{morph-bleu}.

\spacing{1}
\begin{table*}
\begin{center}
\caption{Bleu Scores after Inflectional Normalization}
\label{morph-bleu}
\begin{small}
\begin{tabular}{|p{1.0in}|p{1.0in}|p{1.0in}|}
\hline
 &\textbf{UN} & \textbf{UN-Morph} \\ \hline
SYST & 	21.53 &	43.35	 \\ \hline	
IBM4 &	25.05 &	45.66	 \\ \hline	
MTDR &	14.91 &	37.20	 \\ \hline	
GIST &   5.66 &	29.29	 \\ \hline\hline	
 & \textbf{FBIS}  &	\textbf{FBIS-Morph} \\ \hline
SYST &	22.82 &  47.25	   \\ \hline
IBM4 &	12.85 &	 31.24	   \\ \hline
MTDR &  11.82 &  32.50     \\ \hline
GIST & 	 6.30 &  23.26     \\ \hline\hline
 & \textbf{Bible}  & \textbf{Bible-Morph} \\ \hline
SYST &	12.19 &  32.58     \\ \hline
IBM4 &	 3.65 &	 27.85     \\ \hline
MTDR &   5.59 &  29.00     \\ \hline
GIST &   2.85 &  26.34     \\ \hline
\end{tabular}
\end{small}
\end {center}
\end{table*}
\spacing{1.667}

The first column in Table~\ref{morph-bleu} shows the Bleu score before
 inflectional normalization and the second column shows the score
 after inflectional normalization.  The Bleu scores of all evaluated
 systems rose significantly.  GIST experienced the greatest increase
 since the gisting output was not morphologically inflected to start
 with.  SYST and GIST held their positions as upper and lower limits
 even though in this sample, IBM4 scored higher than SYST on the UN
 corpus before normalization. The distance between IBM4 and MTDR in
 the UN test set reduced by 17\%.  MTDR and IBM4 switch ranks on the
 FBIS test set which provides further support for the argument
 presented earlier regarding MTDR's robustness when used on new genre.

\subsubsection{Synonym Bias}

To evaluate the effect of Bleu's bias regarding synonyms of words, the
words in the same samples (but not the gold standard reference) described in the previous section were
extended to all their possible synonyms using WordNet.  Consider the following example:

\begin{sentences}
\item \label{sb}Raw: The activities supported through IFPPE during 1989 are described below.\\
Morph: the activity support through ifppe during 1989 describe below.\\
WordNet Expansion: {\small the measure measurement step quantity amount quantum bill measuring
mensuration standard criterion touchstone meter beat cadence measure
bar measure measuring rod measuring stick activity action activeness
bodily process body process bodily function natural process natural
action radioactivity activeness support back up see through back
endorse plump for plunk for hold sustain hold up confirm corroborate
substantiate affirm subscribe stand underpin bear out defend fend for
bolster bolster up patronize patronage support through support back up
support back endorse plump for plunk for hold support hold up underpin
bear out ifppe during 1989 describe depict draw report trace draw line
outline delineate identify discover key key out distinguish name below
at a lower place to a lower place beneath below infra downstairs down
the stairs on a lower floor below under below below at a lower place
to a lower place beneath below downstairs down the stairs on a lower
floor below under below .}
\end{sentences}

In this example, the words in the inflectionally normalized (Morph)
version of the sentences are expanded to their WordNet synonyms. For
example, the word {\it support} is expanded to {\it endorse,
support\_through, defend} and many others.  No effort was taken to do
word sense disambiguation. The point is to measure the effect on word
matching by allowing matches of semantically related words while
prohibiting matches of unrelated words such as {\it minefield} and
{\it establish}.  Since Bleu precision cannot be applied directly to a
large bag of words, the RMP metric is used. This is done by
calculating the precision score of the inflectionally normalized
reference against the wordNet-expanded test sentences.  Only unigrams
are considered since bigrams cannot be measured in a bag of words.
The results are presented in Table~\ref{wn-rmp}.

\spacing{1}
\begin{table*}
\begin{center}
\caption{The Effect of Synonym Expansion}
\label{wn-rmp}
\begin{small}
\begin{tabular}{|p{1.0in}|p{1.0in}|p{1.0in}|p{1.0in}|}
\hline
     &	\textbf{UN}    & \textbf{UN-Morph}       & \textbf{UN-Morph-WordNet}\\ \hline
IBM4 &  60.83 & 76.56  & 79.36 (+2.80) \\ \hline
SYST &  61.98 &	77.83  & 81.44 (+3.61) \\ \hline
GIST &  33.16 & 61.41  & 67.41 (+6.00) \\ \hline
MTDR &  54.60 & 72.70  & 76.05 (+3.35) \\ \hline\hline

     &	\textbf{FBIS}  &	\textbf{FBIS-Morph}     & \textbf{FBIS-Morph-WordNet} \\ \hline
IBM4 &  45.66 & 62.26  & 66.62 (+4.36) \\ \hline
SYST &  57.75 & 71.33  & 75.78 (+4.45) \\ \hline
GIST &  31.77 & 55.11  & 61.16 (+6.49) \\ \hline
MTDR &	47.78 & 64.07  & 68.69 (+4.62) \\ \hline\hline

     &  \textbf{Bible} & \textbf{Bible-Morph}    & \textbf{Bible-Morph-WordNet} \\ \hline
IBM4 &	29.87 &	54.63  & 58.40 (+3.77) \\ \hline
SYST &	49.51 & 66.80  & 71.53 (+4.73) \\ \hline
GIST &	27.92 &	55.90  & 61.08 (+5.99) \\ \hline
MTDR &	40.34 &	61.58  & 66.80 (+5.22) \\ \hline\hline
\end{tabular}
\end{small}
\end {center}
\end{table*}
\spacing{1.667}

The first and second column of Table~\ref{wn-rmp}, present the RMP-1
scores for raw text and inflectionally normalized text. The last
column presents the RMP-1 score after synonym expansion.  This is followed by the amount
of increase relative to the second column.  In
all three test sets, the increase in the RMP-1 score for MTDR is
higher than that for IBM4.  The average increase for MTDR is 21\% over
IBM4. This proves that IBM4's mismatches are less semantically related
than MTDR's.

\subsection{Information Loss}

IBM4 consistently lost parts of the sentences which effectively
boosted its modified precision value.  It was the only system that
received a brevity penalty by Bleu. The following is an example of
typical loss of information in IBM4:

\begin{sentences}
\item SP: El da\~{n}o \textbf{causado} al pueblo de Sud\textbf{\'{a}frica} jam\'{a}s debe subestimarse.\\
EN: The damage \textbf{caused} to the people of his country should never be underestimated.\\
IBM4: the damage the people of south must never underestimated .\\
MTDR: Never the \textbf{causado} damage to the people of South \textbf{Africa} should be underestimated.
\end{sentences}

In this example, the words {\it causado (caused)} and {\it \'{a}frica
(Africa)} do not appear at all in IBM4.  MTDR fails to translate {\it
causado} correctly, but it is generated nonetheless. The missing
preposition {\it to} in IBM4 ({\it the damage the people}) can cause
an erroneous reading where {\it the people} are the agent of {\it
underestimating}.

Other systems experienced occasional loss of information, too, but not
at the same rate. IBM4's sentence length is on average 6\% shorter
than the gold standard.  SYST and GIST both are 9\% longer and MTDR
is only 4\% longer. 

\hide{
\begin{table*}
\begin{center}
\caption{Modified Reference Precision}
\label{bleu-revr}
\begin{small}
\begin{tabular}{|p{1.0in}|p{1.0in}|p{1.0in}|p{1.0in}|}
\hline
  & \textbf{UN} & \textbf{FBIS} & \textbf{Bible} \\ \hline
GIST &   9.03  &   3.50  &   2.50 \\ \hline
SYST &  27.07  &  21.41  &  15.21 \\ \hline
IBM4 &  24.63  &  11.72  &   4.69 \\ \hline
MTDR &  18.91  &  10.41  &   7.51 \\ \hline\hline
GIST+MTDR  &  20.88  &  11.74  &   8.43 \\ \hline
SYST+MTDR  &  30.77  &  23.97  &  17.38 \\ \hline
IBM4+MTDR  &  32.17  &  16.86  &   9.70 \\ \hline
GIST+SYST  &  27.90  &  22.09  &  15.80  \\ \hline
GIST+IBM4  &  28.80  &  13.26  &   6.19   \\ \hline
SYST+IBM4  &  36.58  &  24.59  &  16.25  \\ \hline
\end{tabular}
\end{small}
\end {center}
\end{table*}

The increase in IBM4+MTDR (70\% of MTDR alone and 30\% of IBM4 alone)
suggests that the translation choices the two systems make are
complementary to a great extent.
}

\subsection{Linguistic Features}

MTDR handles the translation of linguistic features such as tense and
pro-drop restoration much better than IBM4 does.  The following
examples were observed in the UN test set:
\begin{sentences}
\item \label{lf1}SP: Adem\'{a}s \textbf{dijo} que solamente una inyecci\'{o}n masiva de capital extranjero ...\\
EN: Further, \textbf{he said} that only a massive injection of foreign capital ...\\
IBM4: further \textbf{stated} that only a massive inyecci\'{o}n of capital abroad ...\\
MTDR: Also \textbf{he spoke} only a massive injection of foreign capital ...
\item \label{lf2}SP: Sin embargo, no \textbf{se suministr\'{o}} informaci\'{o}n concreta respecto de esos casos.\\
EN: However, no specific information on the cases involved \textbf{was provided}.\\
IBM4: however , not \textbf{be provided} specific information on those cases .\\
MTDR: Specific information \textbf{was} not \textbf{provided} without embargo proportion of cases.
\end{sentences}
In example~\ref{lf1}, the subject of the Spanish {\it dijo (he said)},
was restored in MTDR {\it he spoke} but not in IBM4 {\it stated}.  In
example~\ref{lf2}, two mistakes appear in IBM4: the auxiliary {\it be}
is not conjugated correctly and the subject {\it specific information}
appears after the verb rather than before it.  In this case IBM4 was
able to translate the complex Spanish passive verb into a passive
English verb, but not moving the subject can cause a
misunderstanding of whether {\it specific information} is really the
subject or object of the verb {\it provide}.

What these example show is that although IBM4 produced better lexical
choice, the ungrammatical nature of the sentence structure can mislead
readers about its content.  Bleu cannot capture syntactic long range
phenomena that are spread over more than a 4-gram. To measure the
``grammaticality'' of the different systems' output, samples of 100
sentences from each the three tested corpora and their gold standard
references were parsed using Connexor's English parser. The Spanish
input was also parsed using Connexor's Spanish parser.  The goal of
this experiment is to determine the correctness of the tested systems
output using parser decisions.  Different outputs can be correct
translations yet have radically different parses. Therefore, two
specific phenomena that can be easily evaluated and that reflect
complex long range parsing choices are focused on: verb determination
and pro-drop restoration.

\subsubsection{Verb Determination}
Determining that a word is a verb depends on satisfying
subcategorization specifications of that verb, such as presence of
subject and object, etc.  The number of words per sentence observed as
verbs is calculated for all sample sentences (see
Table~\ref{verb-num}).

\spacing{1}
\begin{table*}
\begin{center}
\caption{Verbs Per Sentence}
\label{verb-num}
\begin{small}
\begin{tabular}{|p{1.0in}|p{1.0in}|p{1.0in}|p{1.0in}|}
\hline
  & \textbf{UN} & \textbf{FBIS} & \textbf{Bible} \\ \hline
English & 1.37 & 2.07 & 2.14 \\ \hline
Spanish	& 1.49 & 2.10 & 2.23 \\ \hline\hline
GIST    & 1.72 & 2.25 & 2.08 \\ \hline
IBM4    & 1.31 & 1.61 & 1.33 \\ \hline
SYST    & 1.56 & 2.11 & 2.05 \\ \hline
MTDR    & 1.46 & 2.08 & 2.20 \\ \hline
\end{tabular}
\end{small}
\end {center}
\end{table*}
\spacing{1.667}

Table~\ref{verb-num} highlights the fact that the number of words
observed as verbs in IBM4 sentences is radically smaller than all other
systems and the gold standard. It is an outlier for all three test
sets being on average 77\% smaller than the next closest value.  This
is even true for the UN and FBIS test sets where IBM4 scored a higher
Bleu score than MTDR. This implies that sentence structure in IBM4 is
consistently ungrammatical to a high degree.

\subsubsection{Pro-drop Restoration}
The restoration of dropped subjects when translating from languages
like Spanish to English is very important to translation
correctness. To determine how well the different systems accomplished
this task, the ratio of ``realized subjects'' to all verbs is
calculated.  A subject is an argument with a \verb|:subj| relation to
a parent with a part of speech \verb|V|.  A realized subject is a
subject that is not \verb|*pro*| or \verb|*trace*|.  The results are
shown in Table~\ref{realsubj}.

\spacing{1}
\begin{table*}
\begin{center}
\caption{Percentage of Realized Subjects}
\label{realsubj}
\begin{small}
\begin{tabular}{|p{1.0in}|p{1.0in}|p{1.0in}|p{1.0in}|}
\hline
  & \textbf{UN} & \textbf{FBIS} & \textbf{Bible} \\ \hline
English & 70.07\% & 75.84\% & 76.17\% \\ \hline
Spanish	& 31.54\% & 36.19\% & 34.53\% \\ \hline\hline
GIST    & 41.86\% & 41.78\% & 41.83\% \\ \hline
IBM4    & 55.73\% & 54.04\% & 51.13\% \\ \hline
SYST    & 69.23\% & 72.99\% & 77.56\% \\ \hline
MTDR    & 70.55\% & 68.75\% & 75.91\% \\ \hline
\end{tabular}
\end{small}
\end {center}
\end{table*}
\spacing{1.667}

The first two rows in Table~\ref{realsubj} highlight the stark
difference in subject realization between English and Spanish.  Less
than 30\% of English subjects are not realized.  Examples include
subjects appearing in subordinate clauses as \verb|*trace*|: {\it I
want to (*trace*=I graduate)}.  The percentage of realized subjects in
the output of MTDR and SYST is consistent with that of the English
gold standard. GIST's output, as would be expected, is closer to
Spanish.  The output of IBM4 is quite in the middle of the spectrum,
which suggests some amount of pro-drop restoration but only 50\% of
the time.  This last percentage is calculated as the ratio of the
difference between IBM4 and Spanish over the difference between the
English gold standard and Spanish.

%It is important to point out that this is only an estimation due to a
%variety of problems such as errors in parsing and translation
%divergences.  The very wide gap between the Systran and Matador on the
%one hand and IBM4 and GIST on the other is clearly a significant
%indicator of the amount of loss.

%----------------------------------------------------------------------
%----------------------------------------------------------------------
%----------------------------------------------------------------------
%----------------------------------------------------------------------
%----------------------------------------------------------------------
%----------------------------------------------------------------------

\section{Component Evaluation}

This section discusses and evaluates the behavior of specific components in Matador
and its effect on Matador's output.

\subsection{Parsing Evaluation}
Since there is no treebank of any of the evaluated corpora, a test
sample of 100 sentences from the UN test set was parsed and corrected
manually. Of all parent-child links, 17.42\% had at least one error
(wrong parent, wrong relation, wrong POS, wrong lexeme, etc.).  24\%
of all sentences were perfectly parsed.

Examples of bad parses include assigning the lexeme {\it estudiar}, a
verb, to the noun {\it estudiantes} and leaving some prepositions
unattached which leads to very bad translations as in the following example:

\begin{sentences}
\item \label{pe}SP: la necesidad \textbf{de} elaborar m\'{e}todos \textbf{para} determinar la capacidad institucional total\\
  EN: the need \textbf{to} develop methods \textbf{of} identifying aggregate institutional capacity \\
  MTDR: the need created approaches told the total institutional capacity is landed \textbf{in to}
\end{sentences}
 
In example~\ref{pe}, the prepositions {\it de} and {\it para} were
attached directly to the root of the parse tree with no links to
anything else. This is why MTDR generated them in a sequence at the
end of the sentence.

%-----------------------------------------------------------------

\subsection{Translation Dictionary Adequacy Evaluation}
To evaluate the performance of Matador's dictionary, it was extended
with a dictionary extracted from Systran's large Spanish-English
dictionary.  Matador's original dictionary described in
Section~\ref{matador-translation} contains 50,606 word-pos-gloss
triples corresponding to 24,278 Spanish words.  The extended
dictionary contains 119,584 word-pos-gloss triples corresponding to
66,534 Spanish words.  The Bleu scores of the output did not change
significantly or in a consistent manner: the UN test set moved from
18.01 +/- 1.00 up to 18.59 +/- 0.91; but the Bible score moved down
from 7.29 +/- 0.83 to 6.85 +/- 0.72.  These results suggest that the
size of the translation dictionary is adequate.

Examples of the type of errors in the translation lexicon that
appeared in the test set include incorrect spelling, e.g.  {\it (papel N
paper/\textbf{roll})} instead of {\it (papel N paper/\textbf{role})}.
Another example is the use of wrong morphological form for the
lexemes, e.g.  {\it (reglamento N \textbf{regulations/rules})} instead
of {\it (reglamento N \textbf{regulation/rule})}.

%--------------------------------------------------------------
\subsection{Categorial Variation Database Evaluation}

\begin{figure}[ht]
\centerline{\includegraphics[width=3.0in]{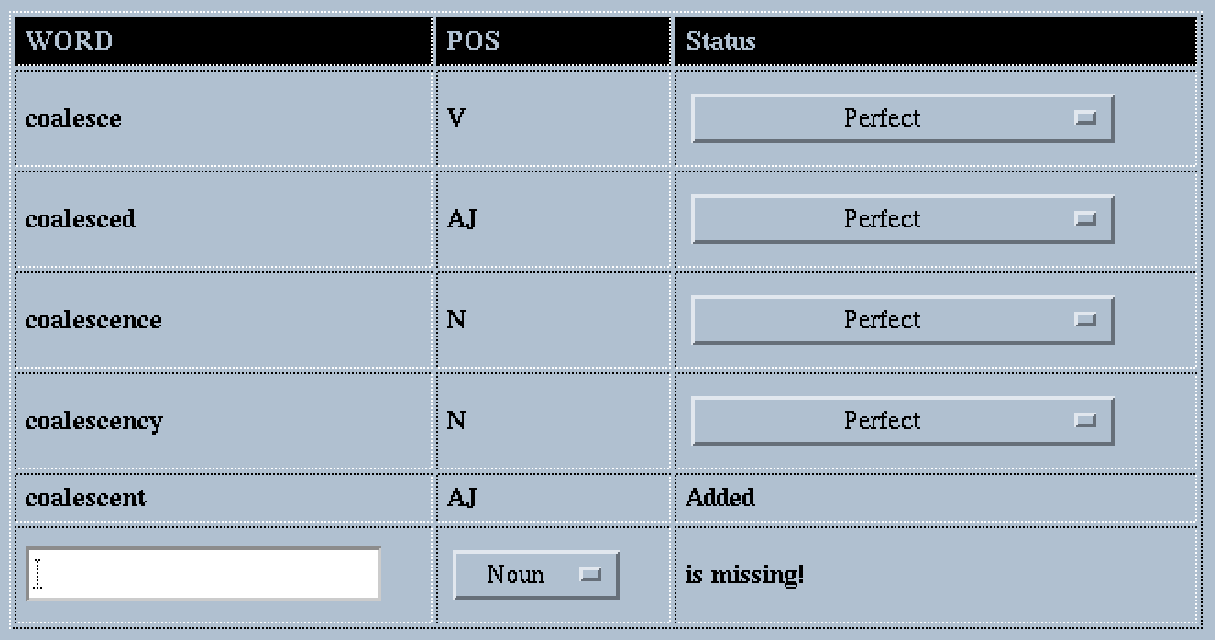}}
%\centerline{\includegraphics[bb = 18 469 601 776 ,width=3.0in]{graphics/cluster-eval-new.ps}}
%\centerline{\includegraphics[bb = 203 545 350 699, angle=-90,origin=c,width=3.0in]{graphics/cluster-eval-new.ps}}
\caption{Evaluation}
\label{evaluation}
\end{figure}

To determine the recall and precision of the CatVar given the lack of a
gold standard, 8 native speakers were asked to evaluate 400
randomly-selected clusters.  Each annotator was given a set of 100
clusters (with two annotators per set).  Figure~\ref{evaluation} shows
a segment of the evaluation interface which was web-browseable.

The annotators were given detailed instructions and many examples to
help them with the task.  They were asked to classify each word in
every cluster as belonging to one of the following categories:
\begin{itemize}
\item Perfect: This word definitely belongs in this cluster.
\item Perfect (except for part of speech problem).
\item Perfect (except for spelling problem).
\item Not Sure: It is not clear whether a word that is derivationally
  correct belongs in a set or not.
\item Does not Belong: This word does not belong in this cluster.
\item May not be a Real Word: This word is not known and could not be 
  found it in a dictionary. 
\end{itemize}

The interface also provided an input text box to add missing words to
a cluster.  In calculating the inter-annotator agreement, mismatches
in word additions were not considered as disagreement since some
annotators could not think up as many possible variations as
others. After all, this was not an evaluation of their ability to
think up variations, but rather of the coverage of the CatVar
database.  Even though there were six fine-grained classifications,
the average inter-annotator agreement was high (80.75\%).  Many of the
disagreements, however, resulted from the fine-grainness of the
options available to the annotators.

In a second calculation of inter-annotator agreement,
the annotators' choices were simplified by placing them into three groups
corresponding to Perfect (Perfect and Perfect-{\it but\/}), Not-sure
(Not-sure and May-not-be-a-real-word) and Wrong (Does-not-belong).
This annotation-grouping approach is comparable to the clustering
techniques used by \cite{Veronis:1998} to ``super-tag'' fine grained
annotations.  After grouping the annotations, average inter-annotator
agreement rose up to 98.35\%.

The cluster modifications produced by each pair of annotators assigned
to the same cluster were then combined automatically in an
approximation to post-annotation inter-annotator discussion, which
traditionally results in agreement: (1) If both annotators agreed on a
category, then it stands; (2) One annotator overrides another in cases
where one is more sure than the other (i.e., Perfect overrides
Perfect-but-with-error/Not-sure and Wrong overrides Not-sure); (3) In
cases where one annotator considered a word Perfect while the other
annotator considered it Wrong, the two view were compromised as a
Not-sure.  The union of all added words was included in the combined
cluster.

The 400 combined clusters covered 808 words. 68\% of the words were
ranked as Perfect. None had spelling errors and only one word had a
part-of-speech issue.  23 words (less than 3\%) were marked as
Not-sures. Only 6 words (less than 1\%) were marked as Wrong.  There
were 209 added words (about 26\%). However 128 words (or 61\% of
missing words) were not actually missing, but rather not linked into
the set of clusters evaluated by a particular annotator.  Some of
these words were clustered separately in the database.\footnote{The
128 words that were ``not really missing'' were clustered in 89 other
clusters not included in the evaluation sample.}  The rest of the
missing words (81 words or 10\% of all words) were not present in the
database, but 50 of them (or 62\%) were linkable to existing words in
the CatVar using simple stemming (e.g., the Porter stemmer, whose
relevance is discussed next).

The precision was calculated as the ratio of perfect words to all
original (i.e. not added) words: 91.82\%. The recall was calculated as
the ratio of perfect words divided by all perfect plus all added
words: 72.46\%.  However, if  the not-really missing words were excluded,
the adjusted recall value becomes 87.16\%.  The harmonic mean or
F-score\footnote{F-score = $\frac{2 \times Precision \times Recall}
{Precision + Recall}$.} of the precision and recall is 81.00\% (or
89.43\% for adjusted recall).

\subsubsection{Linkability Evaluation: Comparison to Porter Stemmer}

To measure the contribution of Catvar with respect to the ``linking
together'' of related words, it is important to define the concept of
{\it linkability\/} as the percentage of word-to-word links in the
database resulting from a specific source.  For example, {\it Natural
linkability\/} refers to pairs of words whose form does not change
across categories such as {\it zip}$_V$ and {\it zip}$_N$ or {\it
afghan}$_N$ and {\it afghan}$_{AJ}$. {\it Porter linkability\/} refers
to words linkable by reduction to a common Porter stem.  {\it CatVar
linkability\/} is the linkability of two words appearing in the same
CatVar cluster.

\begin{figure}[h]
\centerline{\includegraphics[width=2.5in]{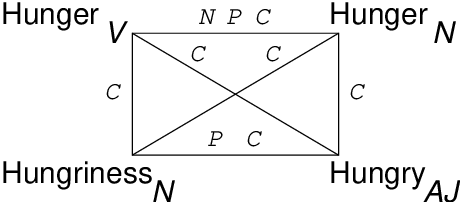}}
\caption{Three Types of Links}
\label{links}
\end{figure}

Figure~\ref{links} shows an example of all three types of links in the
{\it hunger\/} cluster.  Here, {\it hunger}$_N$ and {\it hunger}$_V$
are linked in three ways, Naturally (N), by the Porter stemmer (P),
and in CatVar (C).  Porter links {\it hungry}$_{AJ}$ and {\it
hungriness}$_N$ via the common stem {\it hungri} but Porter does not
link either of these to {\it hunger}$_N$ or {\it hunger}$_V$ (stem
{\it hunger\/}).  The total number of links in this cluster is six,
two of which are Porter-determinable and only one of which is
naturally-determinable.

The calculation of linkability applies only to the portion of the
database containing multi-word clusters (about half of the database)
since single-word clusters have zero links.  The 48,867 linked words
are distributed over 14,731 clusters with 89,638 total number of
links. About 12\% of these links are naturally-determinable and 70\%
are Porter-linkable.  The last 30\% of the links is a significant
contribution of the CatVar database, compared to the Porter Stemmer,
particularly since this stemmer is an industry standard in the
information retrieval community. Porter Stemmer is of course less
relevant to MT because it is only useful for analysis whereas CatVar
can be used for analysis and generation.

\subsection{Structural Expansion Evaluation}
The effect of structural expansion is evaluated by running the test
sets described earlier with a variation of Matador that does not
include thematic linking, structural expansion or syntactic
assignment.  The source-language dependency trees undergo node
translation and go directly to linearization and statistical ranking.
All the resources used in analysis, translation, linearization and
ranking such the dictionary and statistical language model are the
same as Matador's. The results are presented in
Table~\ref{mtdr-depgist}.

\spacing{1}
\begin{table*}
\begin{center}
\caption{Matador versus Dependency Gisting}
\label{mtdr-depgist}
\begin{small}
\begin{tabular}{|p{1.0in}|p{1.0in}|p{1.0in}|p{1.0in}|}
\hline
  & \textbf{UN} & \textbf{FBIS} & \textbf{Bible} \\ \hline
MTDR &  18.01 +/- 1.00  &  10.42 +/- 0.55 &    7.29 +/- 0.83 \\ \hline
DEPG & 17.87 +/- 1.02   & 10.87 +/- 0.54  &   7.33 +/- 0.88 \\ 
\hline
\end{tabular}
\end{small}
\end {center}
\end{table*}
\spacing{1.667}

The results show that the effect of expansion in this implementation
is not significantly helpful.

As the previous sections have shown,
the other modules in the system produced numerous errors whose effect
on thematic linking and structural expansion should not be
underestimated.  Further research into the causes of such errors and
into making structural expansion more robust is needed.  The thematic
level of representation is very important for a range of applications
in NLP such as Natural Language Understanding.

  A sample of size 100 sentences was extracted randomly from the UN
Corpus to determine how different systems handled divergences.  The
sample contained 88\% full sentences, 10\% nominal phrases and 2\%
questions.  One percent were not correctly aligned and another 1\%
were in German not Spanish.  In 15\% of the sentences, a mismatch in
the translation was observed with an average 2.2 words missing in the
Spanish but appearing in the English and 1.1 words missing in the
English but appearing in the Spanish. In 9\% of the sentences,
segments were clearly missing in both languages:

\begin{sentences}
\item SP: Una declaraci\'{o}n del Comit\'{e} Especial contra el\\
  EN: A statement from the Special Committee against
\end{sentences}
 
In this sample set, 40 cases of divergences were identified. Forty
percent of these divergences needed pragmatic knowledge.  For example,
{\it ofercer posibilidades de comercializacion (offer possibilities
for commercialization)} is translated as {\it be marketable}; and {\it
tratar de establecer la paz (try to establish peace)} is translated as
{\it to seek peace}. The rest are in principle resolveable in Matador,
but Bleu gave MTDR credit for four cases and it gave IBM4 credit for 7
cases.  In many cases, Matador was unable to produce the correct output
or a variant of it due to failures in different components.

If problems with syntax or semantically-related lexical choice are
ignored, it is possible to classify the handling of translation
divergences into three broad categories: correct, possible and
incorrect.  Correctly handled cases are those in which a perfect match
or a slightly different match is found.  For example, the generation
of {\it should} where the reference gold standard chooses {\it must}
is essentially correct although no Bleu credit is given.  Possible
cases are those perfectly understood cases in which no divergence
handling takes place or a different kind of divergence handling takes
place.  For example, one system generated {\it to be object of
attention} instead of the reference's {\it to receive attention}.
Finally, incorrect cases are totally wrong cases that are not
understandable.  For example, the output {\it this is not dable}
contains an untranslated Spanish word ({\it dable} meaning {\it
possible}).  The reference translation is {\it this is impossible}.

\spacing{1}
\begin{table}
\begin{center}
\begin{tabular}{|p{1in}|p{1in}|p{1in}|p{1in}|}
\hline 
 & \textbf{Incorrect} & \textbf{Possible} & \textbf{Correct}\\
\hline
\textbf{IBM4} & 13 (32.5\%) & 14 (35\%) & 13 (32.5\%) \\
\hline
\textbf{MTDR} & 11 (27.5\%) & 16 (40\%) & 13 (32.5\%) \\
\hline
\end{tabular}
\end{center}
\caption{Handling of Translation Divergences in MTDR and IBM4}
\label{diver-results}
\end{table}
\spacing{1.667}

Table~\ref{diver-results} displays the number of divergence cases that
fall in each of these three categories for both IBM4 and MTDR. The
classification was done by one speaker of English. The results show
that in terms of divergence handling, MTDR and IBM4 have a comparable
overall performance.

Structural expansion issues fall into two categories: thematic linking
and over-expansion.  Regarding thematic linking, the system has no
information about the statistical distribution of thematic roles and
the prepositions that assign them.  For example, whereas the
preposition {\it with} is mostly associated with {\it instrument} and
{\it companion}, it appears in the lexicon with {\it theme} ({\it fill
the cup with water}) and {\it goal} ({\it mix the sugar with the
eggs}).  This leads to cases of incorrect assignment of theta roles.
A possible solution is a sophisticated language model accounting for
thematic relations.  A corpus that could be used as training data is
currently under development at the University of
Pennsylvania---Proposition Bank (PropBank) \cite{Palmer:2001}.

Regarding Over-expansion, Matador produced odd or incorrect output due
to inappropriate expansions.  In the following two examples, {\it
destroy totally} and {\it publish only/exclusively} are wrongly
conflated into {\it total$_V$} and {\it exclude}, respectively:
\begin{sentences}
\item SP: Adem\'{a}s, \textbf{destruy\'{o} totalmente} sus cultivos de subsistencia ...\\
EN: It had \textbf{totally destroyed} Samoa's staple crops ... \\
IBM4: furthermore , \textbf{destroyed completely} their crops subsistence ... \\
MTDR: Furthermore, it \textbf{totalled} their cultivations of subsistence ...
\item SP: (Dicha adici\'{o}n \textbf{se publica s\'{o}lo} en a\~{n}os impares.)\\
EN: (That addendum is \textbf{issued in odd-numbered} years only.)\\
IBM4: ( such addendum \textbf{issued only} in odd years .\\
MTDR: concerned addendum is \textbf{excluded} in odd years.)
\end{sentences}

Another example of the effect or over-expansion is shown below:

\begin{sentences}
\item SP: El UNICEF \textbf{tiene} un gimnasio subvencionado ...\\
EN: UNICEF \textbf{has} a subsidized fitness centre ... \\
IBM4: unicef \textbf{has} a subsidized fitness ... \\
MTDR:  The UNICEF \textbf{is} a gym it subsidized ...
\end{sentences}

In this example, the verb {\it tener (have)} is translated as {\it
be}.  This expansion is intended to handle cases like {\it tener
hambre (have hunger)} and {\it be hungry}.

%-----------------------------------------------------------------

\subsection{Structural N-gram Evaluation}

\begin{figure*}[htb]
\centerline{\includegraphics[width=4.5in]{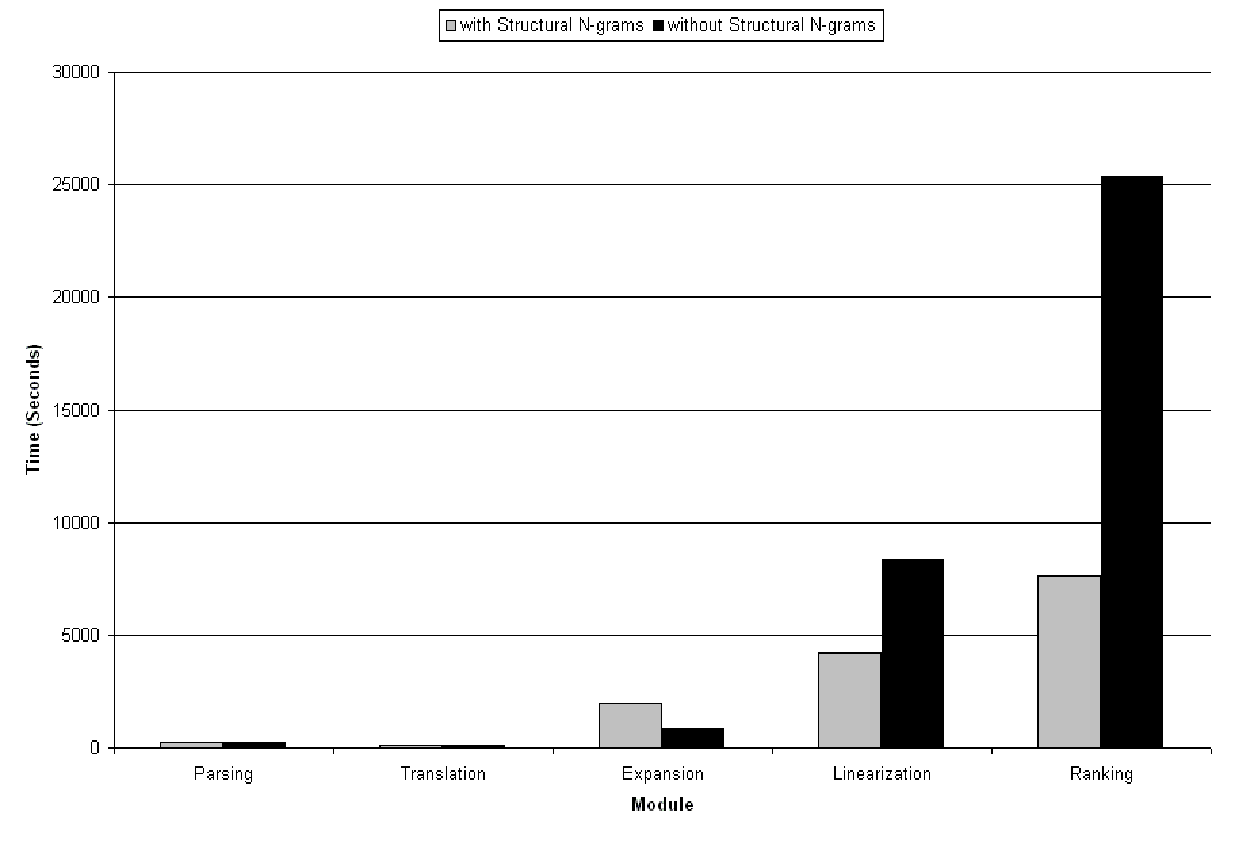}}
\caption{Structural N-gram Evaluation -- Matador's Components}
\label{eval-depgram}
\end{figure*}

The contribution of Structural N-grams (SN-grams) is evaluated by
omitting N-grams in the application of Matador on the UN test set.
The runtime and resulting Bleu scores are presented in
Table~\ref{eval-depgram-table}.  A breakdown of the time over the
different Matador modules is presented in Figure~\ref{eval-depgram}.
\spacing{1}
\begin{table*}
\begin{center}
\caption{Structural N-gram Evaluation}
\label{eval-depgram-table}
\begin{small}
\begin{tabular}{|p{2.0in}|p{1.0in}|p{1.0in}|}
\hline
  & \textbf{Bleu Score} & \textbf{Time (sec)} \\ \hline
with Structural N-grams & 18.01 +/- 1.00   & 14,155 \\ \hline
without Structural N-grams & 17.94 +/- 1.04 & 34,908 \\ 
\hline
\end{tabular}
\end{small}
\end {center}
\end{table*}
\spacing{1.667}

The use of SN-grams decreases runtime by 59.45\% with no negative
effect on text quality.  Running the SN-gram pruning doubles the
runtime of the Expansion module.  However the payoff is a 50\% decrease in
runtime of linearization and ranking, both of which are
significantly costlier time-wise than expansion.

There are two issues with respect to the application of SN-grams: lack
of linear-order information and lack of syntactic knowledge.  First,
SN-gram pruning is unaware of linear order or the relation between
sister words. Linguistic phenomena that depend on such relations are
not captured.  For example, the selection of the form of the
indefinite article in English ($a$ versus $an$) is dependent on the
phonology of its following word. The decision, however, in the SN-gram
model is dependent on the parent-child relationship.  This leads to
errors in Matador's output such as {\it it is an possible exception}.

Regarding the lack of syntactic knowledge, the current implementation
of SN-grams is blind to part-of-speech information, which often leads to
confusing relationships and erroneous selections, especially in a
language like English where Verb/Noun homographs are common.

In addition, the current implementation of SN-grams is only aware of bigram
relations.  This, together with the lack of part-of-speech information
can lead to erroneous selections such as {\it they continued to their
efforts}.

Finally, a possible problem that needs to be investigated is the SN-grams
unawareness of structure in a Probabilistic Context-Free Grammar
(PCFG) sense.  This may lead to erroneous selection of prepositions
which are dependent on the selection of case/prepositions of sister
nodes.

\subsection{Surface N-gram Language Model Evaluation}

The effect of the language model is evaluated using the
RMP-1 metric, which gives a sense of lexical selection independent of
linear order.  The goal is to calculate how well a ``perfect''
language model could have done given the lattice produced by the
previous system modules and to compare that to what the language model
does.  To determine the former, a file is created where, for each
sentence, all the unique words in the lattice are listed. This file
was used as a Bleu reference to determine the 1-gram precision
coverage of the human gold standard.  The result is 68.91. To
determine the latter, the file containing the lattice words is
substituted with Matador's output.  The 1-gram precision for this case
is 55.39.  The ratio of the score of Matador's output relative to a
``possible output'' given the data, in this case 80.38\%, is a lower
limit on how bad the language model is. That is, the language model
could not have possibly reduced the 1-gram scores by more than
19.26\%.  This evaluation was conducted using the UN sample described
earlier.

 One issue relevant to the language model is the use of bigrams
instead of trigrams, which are known to perform better.  This decision
is purely based on technical issues, namely that Halogen's runtime
performance with trigrams is prohibitively long \cite{Daume:2002}.

\paragraph{}
This chapter described a thorough evaluation of Matador and its
various components. The results conclude that Matador is a cross-genre
robust translation system with superior output quality, in terms of
grammaticality and accuracy, relative to a primarily statistical approach.  The
evaluation of Matador's components revealed a higher degree of
linkability in CatVar database relative to Porter stemmer and a
superior efficient performance associated with using structural
n-grams.

%%%%%%%%%%%%%%%%%%%%%%%%%%%%%%%%%%%%%%%%%%%%%%%%%%%%%%%%%%%%%%%%%%%
%%%%%%%%%%%%%%%%%%%%%%%%%%%%%%%%%%%%%%%%%%%%%%%%%%%%%%%%%%%%%%%%%%%
%%%%%%%%%%%%%%%%%%%%%%%%%%%%%%%%%%%%%%%%%%%%%%%%%%%%%%%%%%%%%%%%%%%
%%%%%%%%%%%%%%%%%%%%%%%%%%%%%%%%%%%%%%%%%%%%%%%%%%%%%%%%%%%%%%%%%%%
%%%%%%%%%%%%%%%%%%%%%%%%%%%%%%%%%%%%%%%%%%%%%%%%%%%%%%%%%%%%%%%%%%%
%%%%%%%%%%%%%%%%%%%%%%%%%%%%%%%%%%%%%%%%%%%%%%%%%%%%%%%%%%%%%%%%%%%
%%%%%%%%%%%%%%%%%%%%%%%%%%%%%%%%%%%%%%%%%%%%%%%%%%%%%%%%%%%%%%%%%%%
%%%%%%%%%%%%%%%%%%%%%%%%%%%%%%%%%%%%%%%%%%%%%%%%%%%%%%%%%%%%%%%%%%%
%%%%%%%%%%%%%%%%%%%%%%%%%%%%%%%%%%%%%%%%%%%%%%%%%%%%%%%%%%%%%%%%%%%
%%%%%%%%%%%%%%%%%%%%%%%%%%%%%%%%%%%%%%%%%%%%%%%%%%%%%%%%%%%%%%%%%%%
%%%%%%%%%%%%%%%%%%%%%%%%%%%%%%%%%%%%%%%%%%%%%%%%%%%%%%%%%%%%%%%%%%%
%%%%%%%%%%%%%%%%%%%%%%%%%%%%%%%%%%%%%%%%%%%%%%%%%%%%%%%%%%%%%%%%%%%
%%%%%%%%%%%%%%%%%%%%%%%%%%%%%%%%%%%%%%%%%%%%%%%%%%%%%%%%%%%%%%%%%%%
\chapter{Conclusion}\label{last-chapter}

This thesis has presented a novel machine translation approach,
Generation-Heavy Hybrid Machine Translation, and an implementation of
it for Spanish-English, Matador.  An evaluation of Matador's
translation quality was conducted using the IBM Bleu metric and
comparing against three systems --- simple gisting, primarily
statistical (IBM Model 4) and purely symbolic (Systran) --- over three
corpora (UN, FBIS and Bible).  The evaluation shows that although
Matador scores lower than IBM-model 4 on the corpus where all language
models were trained (UN), Matador has a higher degree of robustness
and scores higher when tested on text with new genre (Bible).
Additionally, Matador's output quality, in terms of grammaticality and
accuracy, is superior to IBM Model 4. Systran and gisting are the best
and worst respectively for all corpora.  A second evaluation verifies
the advantages of using structural n-grams, which increase Matador's
speed by 60\% with no negative effect on text quality.

The rest of this chapter describes the contributions of this
research and its future work.

\section{Contributions}\label{contributions}

The following are the contributions of this research:

\begin{itemize}
\item A new model for machine translation that transcends the need for
large amounts of symmetric knowledge such as transfer rules,
interlingual lexicons, or large annotated bilingual corpora. This
generation-heavy model banks on the availability of target-language
resources to allow for large-scale machine translation from
resource-impoverished source languages (high resourceability).

\item A systematic framework for handling translation divergences that
uniformly accommodates a wide range of seemingly different divergence
types and their interactions. This approach is distinct from all previous
approaches to translation divergences in that it depends fully on
target-language resources (lexical semantics, categorial variations,
statistical language models).

\item A hybrid (symbolic-statistical) generation approach that expands
the concept of symbolic overgeneration to include conflation and
head-swapping of structural variations.

\item A general reusable and extensible model for the evaluation and
development of other researchers' analysis systems and lexicons; the
aim is to allow researchers to test long-standing hypotheses about
translation or generation. Some modules may be swapped out or used
separately by other researchers (i.e. Oxygen and Categorial Variation
Database)

\item A model for language-independent syntactic-to-thematic linking
that obviates the need for monolingual alternation and cross-lingual
argument/oblique realization variations, thus eliminating
hard-coded/language-specific mechanisms for argument linking.
Inverted thematic hierarchies and preposition-to-role mappings are
used to constrain the linking.

\item A structural n-gram model for natural language generation that
improves speed with no loss in quality.

\item An evaluation of the translation quality of four models of
machine translation with varying levels of complexity. 

\end{itemize} 

\section{Direction of Future Work}

This section describes future directions associated with different
parts of the system.

\subsection{Applying Lessons Learned}

The extensive evaluation conducted in Chapter~\ref{eval-chapter} has
brought to light a number of important issues regarding the
implementation of Exerge and Matador.  The first objective of future
work is to apply these lessons learned to the system and reevaluate
it.

\subsection{Categorial Variation Database}

Future work includes (1) improving the word-cluster ratio; (2)
absorbing more of the single-word clusters into existing clusters or
other single-word clusters; (3) enriching the clusters with types of
derivational relations such as ``nominal-event'' or ``doer'' to
complement part-of-speech labels; (4) measuring the applied
contribution of using the CatVar in natural-language applications,
specifically information retrieval; and (5) incorporating CatVar into
new applications such as parallel corpus word alignment.

\subsection{Structural Expansion}

Future work on structural expansion includes extending the work on
generation-heavy verb-divergence to noun-divergences in an effort to
create a general model for divergence handling.

Additionally, the implementation of verb divergence handling in
Matador was limited to main verbs in a dependency tree and only then
to cases where conflated/inflated nodes had no children.  This is a
major limitation that should be removed and a general implementation
of the basic structural expansion operations should be created to
truly be able to test their contribution.  Another extension is to
make the system remember the scores used in earlier operations in
order to make better judgments later. For example, currently, the
scoring to decide which thematic linking sets to use is destroyed once
the step is over.

Finally, structural expansion can be improved with better thematic
linking.  Once PropBank is both available and large-scale enough to
provide appropriate training data, it will be used to collect
statistics on the assignment of thematic roles by prepositions to
substitute the symbolic heuristics currently used with statistical
models.

\subsection{Structural N-grams}
SN-grams were shown to be very helpful in Matador. Therefore, some of
their limitations, which were described earlier, will be addressed to
exploit their usefulness more fully.

Other extensions of the use of SN-grams include combined modeling with
surface n-grams.  Within this approach, a dependency tree is
linearized ``in-tree'' (by rotating the children to the left or right
of parent and relative to each other).  Long-distance dependencies and
short-distance dependencies would be modeled simultaneously.

An evaluation comparing the performance of SN-grams to other
structural models from the parsing community, specifically Tree-grams
\cite{Simaan:2000}, is planned to evaluate the effect of not
accounting for phrasal structure in SN-grams.

\subsection{Source Retargetability Evaluation}\label{resourceability}
\begin{figure}[htb]
\centerline{\includegraphics[width=5.0in]{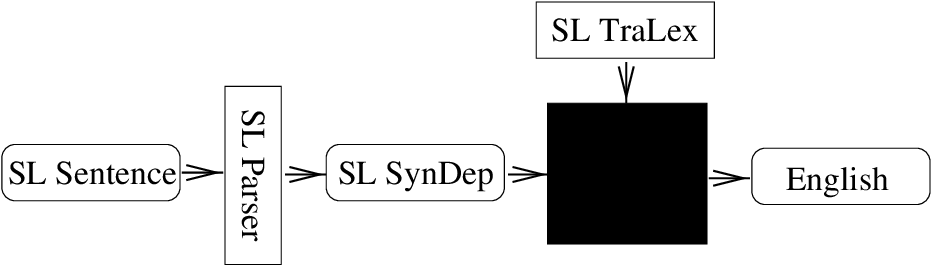}}
\caption{Retargetability to a new source language}
\label{overview}
\end{figure}

To evaluate Exerge's source-language independence, a second MT system
for Chinese-English will be implemented (see Figure~\ref{overview}).
Basically, two components are needed: a SL parser and a SL-TL
translation dictionary.  The black box in Figure~\ref{overview} is
occupied by the Exerge system. A wrapper around the parser might be
needed to ensure its output is in the correct syntactic dependency
(SynDep) format compatible with Exerge.  The new MT system will be
evaluated, just like Matador was, on system performance and text
quality.

\subsection{Human Translation Quality Evaluation}

There are issues with automatic evaluation techniques such as
the Bleu metric.
These issues can be better understood by comparing how humans would
respond to the translation quality of the output of the four systems
tested in Chapter~\ref{eval-chapter}.  To accomplish this, 
 two evaluations will be conducted: one for Accuracy (Fidelity) and one for
Fluency (Intelligibility). Both tests would use a set of 25 sentences
to be randomly selected from the test sets used earlier.  The total
number of subjects will be 50 (25 for Accuracy and 25 for Fluency).
All subjects are to be native speakers of English.  Two evaluations
will be performed online using a web interface:

\begin{itemize}
\item Accuracy Evaluation \\
  This evaluation measures the Accuracy or Fidelity of the translation
  system, i.e., how well a system preserves the meaning of the original
  text whether the target language is fluent or not.  For each of the 25
  sentences evaluated, subjects will be given the gold standard human
  translation and the machine translation output of each of the four
  evaluated systems.  Subjects will be asked to rate the translation
  accuracy of each machine translation sentence on a 5-point scale that
  is loosely based on Nagao's 7-point scale for Fidelity
  \cite{Nagao:1989} (see table \ref{acc-criteria}).  A score of 5 is
  given where the content of the original sentence is fully conveyed
  (might need minor corrections). A score of 1 is given where the
  content of the original sentence is not conveyed at all.
  \spacing{1}
  \begin{table}[ht]
    \begin{center}
      \begin{tabular}{|l|p{5.0in}|}
	\hline
	5&\small{contents of original sentence conveyed (might need minor corrections)}\\ \hline
	4&\small{contents of original sentence conveyed BUT errors in word order}\\ \hline
	3&\small{contents of original sentence generally conveyed BUT errors in relationship between phrases, tense, singular/plural, etc.}\\ \hline
	2&\small{contents of original sentence not adequately conveyed, portions of original sentence incorrectly translated, missing modifiers}\\ \hline
	1&\small{contents of original sentence not conveyed, missing verbs, subjects, objects, phrases or clauses}\\ \hline
      \end{tabular}
    \end{center}
    \caption{Accuracy Criteria}
    \label{acc-criteria}
  \end{table}
  \spacing{1.667}

\item Fluency Evaluation \\
  The purpose of this evaluation is to measure the Fluency (or
  Intelligibility) of the translation systems.  In the fluency
  evaluation, subjects will be given five samples to score for each of
  the 25 sentences.  In addition to the four evaluated systems, the gold
  standard is to be evaluated on its fluency too to give us a
  normalizable reference point.  Subjects will be asked to rate the
  Fluency of the sentences on a 5-point scale that is loosely based on
  Nagao's intelligibility scale metric \cite{Nagao:1989}. The scale
  ranges from 5 (clear meaning, fluent sentence) to 1 (meaning
  absolutely unclear, sentence not fluent).  Table~\ref{flu-criteria}
  details the criteria used in measuring fluency.

  \spacing{1}
  \begin{table}
    \begin{center}
      \begin{tabular}{|l|p{5.0in}|}
	\hline
	5&\small{clear meaning, good grammar, terminology and sentence structure}\\ \hline
	4&\small{clear meaning BUT bad grammar, bad terminology or bad sentence structure}  \\ \hline
	3&\small{meaning graspable BUT ambiguities due to bad grammar, bad terminology or bad sentence structure }\\ \hline
	2&\small{meaning unclear BUT inferable} \\ \hline
	1&\small{meaning absolutely unclear}\\ \hline
      \end{tabular}
    \end{center}
    \caption{Fluency Criteria}
    \label{flu-criteria}
  \end{table}
\end{itemize}
  \spacing{1.667}

\paragraph{}
This chapter concludes this thesis, which has presented
Generation-Heavy Hybrid Machine Translation.  GHMT is an asymmetrical
hybrid approach to MT that has a higher degree of retargetability and
robustness compared to other symmetrical approaches that either have a
much higher cost (symbolic approaches) or a much lower quality on new
input genre (statistical approaches). The implementational details of
Matador, a Spanish-English implementation of GHMT, and an extensive
evaluation have been discussed.

%[[add conclusion: 
%future: use bilingual corpora... to do this may even be better than ISI, still in GHMT spirit]]

%%%%%%%%%%%%%%%%%%%%%%%%%%%%%%%%%%%%%%%%%%%%%%%%%%%%%%%%%%%%%%%%%%%%%%%%
%\chapter*{Appendix A}
%\addcontentsline{toc}{chapter}{Appendix A}

%%%%%%%%%%%%%%%%%%%%%%%%%%%%%%%%%%%%%%%%%%%%%%%%%%%%%%%%%%%%%%%%%%%%%%%%
%\addcontentsline{toc}{chapter}{Bibliography}
% Enter the name of bibtex  bibliography database, if you wish
%\renewcommand \bibname {\centerline{\large \rm BIBLIOGRAPHY}}
 
\newpage
\addcontentsline{toc}{chapter}{Bibliography}
\begin{small}
\bibliography{bibliography,dissertation}
%\bibliographystyle{fullname}
%Style for final thesis should be plain
\bibliographystyle{plain}
\end{small}

\end{document}